\documentclass{declare}

\usepackage{amsmath}
\usepackage{amssymb}
\usepackage{xspace}
\usepackage[shortlabels,inline]{enumitem}
\usepackage{pdfpages}
\setlist[itemize]{leftmargin=*,topsep=2pt,itemsep=2pt}
\setlist[enumerate]{leftmargin=*,topsep=2pt,itemsep=2pt}

\newcommand{\turnfigwidth}{0.62\linewidth}

\usepackage{cuted}
\usepackage{makecell}
\usepackage{multirow}
\usepackage{amsthm}
\usepackage{algorithm}
\usepackage[noend]{algpseudocode}
\usepackage{tikz}
\usetikzlibrary{positioning}
\usepackage[table]{xcolor}
\usepackage{adjustbox}

\newtheorem{proposition}{Proposition}

\newcommand{\dlt}[1]{{\scriptsize\textcolor{gray}{#1}}}
\DeclareRobustCommand{\bline}[1]{\mbox{\frenchspacing\texttt{#1}}}

\providecommand{\turnfigwidth}{\linewidth}

\algrenewcommand\algorithmiccomment[1]{\hfill{\footnotesize(#1)}}

\makeatletter
\@ifclassloaded{declare}{}{\usepackage[T1,OT1]{fontenc}}
\makeatother
\definecolor{methodnavy}{RGB}{22,52,102}
\makeatletter
\newcommand{\method@setseries}{%
  \edef\method@cur{\f@series}\def\method@b{b}\def\method@bx{bx}%
  \ifx\method@cur\method@b \def\method@ser{b}\else
  \ifx\method@cur\method@bx\def\method@ser{b}\else
                           \def\method@ser{m}\fi\fi}
\DeclareRobustCommand{\method@name}{\leavevmode{\method@setseries\color{methodnavy}%
  \usefont{T1}{LinuxLibertineT-TLF}{\method@ser}{n}RLVB}}
\newcommand{\method}{\texorpdfstring{\method@name}{RLVB}\xspace}
\makeatother

\providecommand{\mainref}[1]{\cref{#1}}

\makeatletter
\DeclareRobustCommand\onedot{\futurelet\@let@token\@onedot}
\def\@onedot{\ifx\@let@token.\else.\null\fi\xspace}

\def\etal{\emph{et al}\onedot}
\makeatother

\let\cite\citep

\DeclareTextFontCommand{\tabbf}{\bfseries}
\AtBeginEnvironment{tabular}{\let\textbf\tabbf}

\logo{\includegraphics[height=10.5mm]{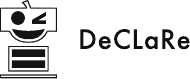}}
\secondarylogos{\includegraphics[height=10.5mm]{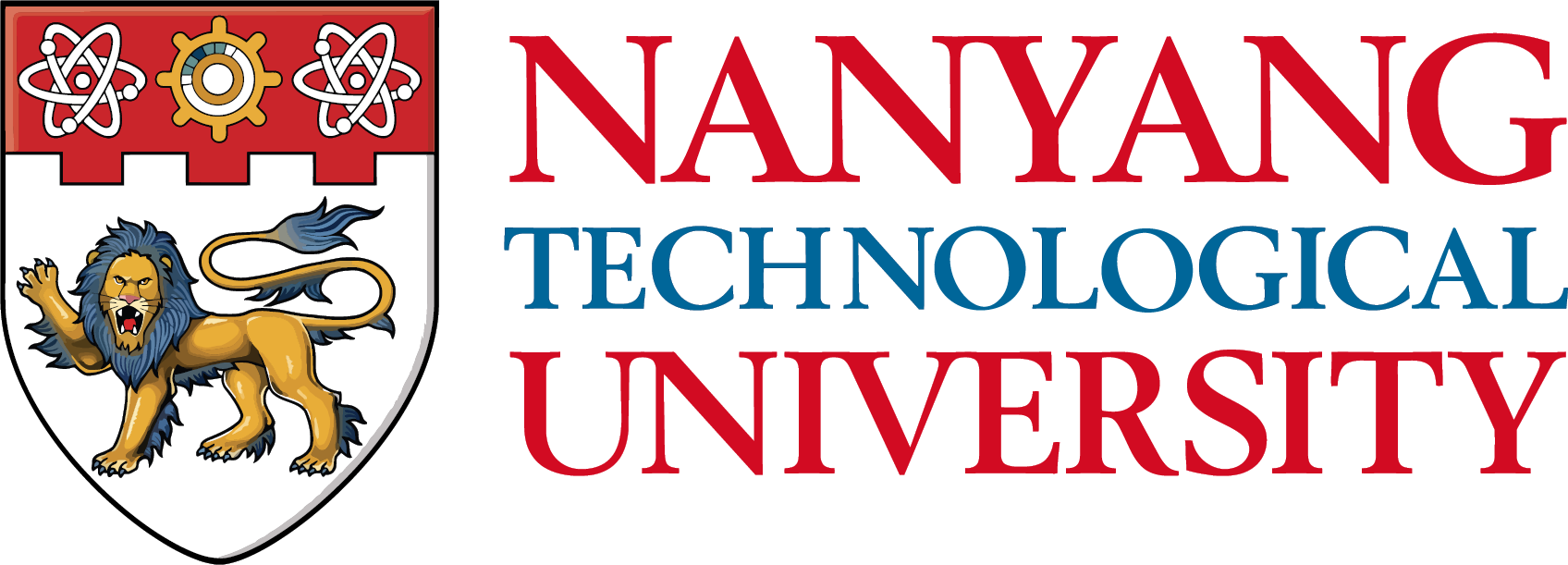}}

\title{Lazy Agents Answer Against Their Own Beliefs: Reinforcement Learning with Verifiable Beliefs Teaches Them to Act on What They Believe}

\author[*]{Richmond Sin Jing Xuan}
\author[*]{Peilin Feng}
\author{Wenbin Wang}
\author{Soujanya Poria}
\affiliation{DeCLaRe Lab, Nanyang Technological University, Singapore}
\contribution[*]{Equal contribution}

\date{\today}
\metadata[\faGithub\ Code]{\url{https://github.com/declare-lab/RLVB}}

\makeatletter
\let\kl@abstracttext\abstract
\renewenvironment{abstract}{}{}
\kl@abstracttext{\begin{abstract}
AI agents working on long tasks often stop too soon and commit to an answer before they have enough evidence, a behavior often called laziness. In our evaluations, they frequently commit even while their stated belief still names several possible answers. At every turn, the agent decides whether to take a further step or to commit to an answer. An outcome reward at the end of the episode scores only the final answer. It does not show which turns helped, and it cannot teach the agent when to commit, because a lucky guess receives the same reward as an answer backed by evidence. During training, however, the simulator knows the rules linking the hidden answer to the evidence, so it can compute the exact posterior over the answer at every step and tell whether the evidence is already enough. We propose \textbf{Reinforcement Learning with Verifiable Beliefs (\method)}, which takes this posterior as the belief supported by the evidence. \method rewards the agent for taking further steps while this belief still allows several answers and for committing once the evidence is sufficient. The reward depends only on the evidence the agent has gathered and needs no judge, critic or reward model. It never reads the belief line the agent writes before each action, although the update trains this line together with the action. We also introduce Pro-Worlds, a benchmark of visual puzzles where the agent views a large world through a small window and must decide when it has enough visual evidence to commit. We evaluate three vision-language models on three benchmarks, two of which are never seen in training. \method improves the base model in all nine settings, by 17.1\% on average and by 10.4\% with expectation maximization (EM) as the learner. Under the outcome reward, GRPO and PPO fall below the base model on average. \method also outperforms PPO with a process reward from an LLM judge by 9.3\% on average while using 1.6 times fewer GPU-hours. To find where this gain comes from, we group the answers by the evidence the agent has when it answers. The trained agent answers with sufficient evidence in 51\%, 12\% and 74\% of the episodes on the three benchmarks, against 8\%, 0.4\% and 36\% for the base model. For every method, 66 to 92\% of the answers given with sufficient evidence are correct. The accuracy gain of the trained agent therefore comes from answering more often at high levels of evidence, while at the same level of evidence its answers are no more accurate than those of the base model. The belief line shows how the trained agent reaches this evidence: it acts on its own belief. It commits at only 0.1 to 2.2\% of the turns at which the evidence is insufficient and its line still names several answers, against 3 to 68\% for the base model with the same line. Over all turns, its line is not consistently more exact than that of the base model, and each of its steps adds about as much evidence as a step of the base model. Because the trained agent acts on its belief, we expect better reading of the evidence to translate into better answers.
\end{abstract}}
\makeatother

\newcommand{\klsupplpage}{1}
\newcommand{\maketitlesupplementary}{%
  \setcounter{page}{\klsupplpage}%
  {\titlefont\sffamily\bfseries\textcolor{klgold}{Supplementary Material}\par}%
  \vspace{0.6em}{\color{klgold}\hrule height 0.8pt}\vspace{1em}}

\begin{document}
\maketitle

\section{Introduction}
\label{sec:intro}

\begin{figure}[!htbp]
  \centering
  \definecolor{fpossible}{HTML}{EB6834}\definecolor{fruled}{HTML}{898781}
  \begin{tikzpicture}[lab/.style={circle, inner sep=0pt, minimum size=8.5pt, font=\sffamily\tiny\bfseries, text=white, draw=white, line width=0.6pt}]
    \node[anchor=south west, inner sep=0] (img) at (0,0) {\includegraphics[width=\turnfigwidth]{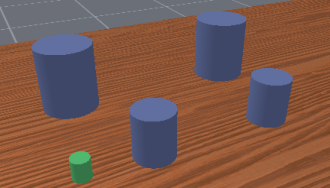}};
    \begin{scope}[x={(img.south east)}, y={(img.north west)}]
      \node[lab, fill=fruled]    at (0.197, 0.585) {1};
      \node[lab, fill=fpossible] at (0.467, 0.277) {2};
      \node[lab, fill=fruled]    at (0.664, 0.750) {3};
      \node[lab, fill=fpossible] at (0.818, 0.479) {4};
    \end{scope}
  \end{tikzpicture}\\[3pt]
  {\footnotesize\raggedright The agent must find the red cylinder. Cup 1 hides a green cylinder and cup 3 is empty. The red cylinder is therefore under \textcolor{fpossible}{cup 2} or \textcolor{fpossible}{cup 4}, each with probability one half.\par}\vspace{4pt}
  {\footnotesize\setlength{\tabcolsep}{2.6pt}
  \begin{tabular}{@{}ll rrr c c@{}}
    \toprule
    \multicolumn{2}{@{}l}{Turn at this history} & \multicolumn{3}{c}{\method (posterior)} & Truth & Outcome \\
    \cmidrule(r){1-2}\cmidrule(lr){3-5}
    \texttt{Possible:} & Action & belief & action & Reward & belief & reward \\
    \midrule
    \texttt{cups 2, 4} & \texttt{LIFT 2}   & $-0.07$ & $0.00$  & $\mathbf{-0.07}$ & $1/2$ & later \\
    \texttt{all}       & \texttt{LIFT 2}   & $-2.64$ & $0.00$  & $-2.64$          & $1/4$ & later \\
    \texttt{cups 2, 4} & \texttt{SUBMIT 4} & $-0.07$ & $-7.42$ & $-7.49$          & $1/2$ & 1 or 0 \\
    \texttt{cup 4}     & \texttt{SUBMIT 4} & $-9.92$ & $-7.42$ & $-17.34$         & \textbf{1 or 0} & 1 or 0 \\
    \bottomrule
  \end{tabular}\par}
  \caption{\textbf{Posterior process rewards on a real scene of the cups task.} The table scores four alternative turns after the agent
  has lifted cups 1 and 3. \method scores them against the exact posterior. \emph{Truth} is the share of true claims in the belief, as scored by a judge that sees the hidden answer. \emph{Outcome} is the final reward. The \method scores in the table include both terms of \cref{eq:score}. The reported \method is trained with the action term.}
  \label{fig:turn}
\end{figure}

Agents that work over long horizons are now used in games, embodied settings, and other interactive environments~\cite{yao2023react,yang2025embodiedbench,wang2025vagen,wang2025ragen}. Such an agent must keep working until it has gathered sufficient evidence to act~\cite{bajcsy1988active,das2018embodied}. However, current agents often stop too early~\cite{laban2025llms,liu2026roboquest,cuadron2025danger} and answer with a guess. This behavior is called laziness~\cite{ma2025quantifying}. In the cups task in \cref{fig:turn}, the agent lifts one cup at a time to find a hidden object. After two lifts, the evidence is not yet sufficient because the object could still be under either of two cups. The next action should therefore lift another cup instead of guessing. In our experiments, the base models often guess even after they write that several cups remain possible.

At every turn, an evidence-seeking agent should keep a belief that matches the evidence. It should take steps aimed at what this belief leaves open. It should commit only when the belief identifies the answer. Knowing how much one knows and acting on it is called metacognition~\cite{flavell1979metacognition}. Metacognition has two parts: monitoring what one knows and controlling what one does with it~\cite{nelson1990metamemory}. Language models can often recognize what they know~\cite{kadavath2022language}. However, the base models in our evaluations commit too early, even though our prompts tell them to commit only once the evidence identifies the answer. When they write a belief line before each action, they commit at 3 to 68\% of the turns at which the evidence is insufficient and this belief line still names several answers. They often state that the answer remains open and commit anyway. These models do not act on what their own belief line states. A check in the harness that rejects the first answer does not solve this problem, because the harness cannot read the observations, as \cref{sec:more_looks} shows. Whether the evidence determines the answer depends on what the agent has observed. Only the agent reads these observations. The agent's own belief can therefore guide each decision once the agent learns to act on it. We call an agent reliable when its answers are both correct and backed by sufficient evidence, because a user who receives an answer usually cannot check how the agent reached it. Training an agent to make this decision reliably over a long horizon is a central problem in agentic reinforcement learning (RL).

RL with verifiable rewards (RLVR) is a common way to train large language models on
math~\cite{shao2024deepseekmath,zeng2025simplerl,liu2025understanding,yu2025dapo} and
code~\cite{le2022coderl,gehring2025rlef,wei2025swerl} tasks. A verifier checks the final answer and rewards the model when the answer
is correct~\cite{lambert2024tulu,guo2025deepseek,team2025kimi}. Recent work trains multi-turn agents with the same
kind of framework~\cite{zhou2024archer,bai2024digirl,jin2025searchr1,wang2025ragen,wang2025vagen,feng2025gigpo}. Over a long horizon,
however, a reward on the final answer cannot tell which steps helped and which steps caused a
failure~\cite{minsky1961steps,sutton2018reinforcement}. A lucky guess is reinforced as fully as an answer supported by the evidence.
A careful search is penalized whenever the episode is lost. Agents also tend to answer before the evidence supports
the final answer~\cite{laban2025llms,liu2026roboquest} or fall back on fixed patterns that ignore the evidence~\cite{wang2025ragen,wang2026ragen2}.
As \cref{tab:process} shows, the base models on HiddenGames give 64 to 97\% of their answers before their evidence is sufficient, whether or not they write a belief line. RL with the outcome reward even lowers the accuracy of these models on average.

To credit single steps, recent work returns to the PPO framework~\cite{schulman2017proximal} and rewards every step with a process
reward~\cite{uesato2022solving,lightman2024lets}. This reward is often given by a process reward model or an LLM
judge~\cite{cui2025prime,khalifa2025thinkprm}. Yet the value of a step is hard to estimate, because the critic of PPO is fitted to
outcomes that depend on the hidden target. As \cref{fig:credit}(b) shows, an LLM judge that sees the target may reward a lucky guess and penalize a correct but still uncertain belief. The critic and the judge also make training more expensive, because the critic is a second model trained next to the policy and the judge is called at every step.

Agentic RL usually trains in a sandbox simulator~\cite{wang2025ragen,wang2025vagen,feng2025gigpo}. This simulator knows much more than whether the final answer is correct. It knows the hidden target and what each action will reveal about it. Prior work passes this privileged information to a teacher or a critic~\cite{pinto2018asymmetric,chen2019learning,choudhury2025better}. In tasks where the agent gathers evidence, this information lets the simulator compute the exact posterior over the hidden answer after every step. This posterior is the belief state~\cite{astrom1965optimal,kaelbling1998planning}. The final answer is then no longer the only thing the simulator can verify. The posterior is the belief supported by the evidence. The simulator can verify it at every step, just as RLVR verifies the answer at the end. This belief shows whether the agent should take further steps or answer. This moves us from reinforcement learning with verifiable rewards to \textbf{Reinforcement Learning with Verifiable Beliefs (\method)}.

At each turn, we give every method its full history of actions and observations. As a result, no method has to remember what it has explored. From the exact posterior, the agent learns when to commit and when to keep working. The reward also scores which step the agent chooses next, but the trained agent does not choose more informative steps than the base model, as \cref{sec:more_looks} shows. Before each action, the agent states the answers it still sees as possible, such as \bline{Possible: cups 2, 4}. This belief line shows whether the agent thinks it has sufficient evidence. The reward never reads this line. Because the update raises the likelihood of the whole turn, it still trains the line along with the action. As \cref{sec:more_looks} shows, the trained agent then learns to act on the line. The turn's reward compares the agent's next action to the action a teacher chooses by looking ahead on the exact posterior~\cite{howard1966information}. The reward can also score the stated belief against the posterior, but scoring the action is enough, as \cref{sec:ablation} shows. The reward never pays for a guess. It depends on the evidence the agent has gathered. It does not depend on which answer happens to be correct in the episode. In \cref{fig:turn}, cups 2 and 4 are still possible. Lifting another cup gets a reward close to zero. Answering cup 4 gets about $-7$, whether or not the object is under cup 4. At each history, the agent samples several of its own turns and makes the higher-reward turns more likely. The learner uses either a policy gradient that compares turns within a group, as in GRPO~\cite{shao2024deepseekmath}, or expectation maximization (EM)~\cite{dempster1977maximum}. \method needs no judge, critic or reward model because the posterior is exact.

Our contributions are:
\begin{itemize}
  \item \textbf{\method.} We use the simulator's privileged information to compute the exact posterior after every step. We turn this posterior into a process reward. This reward scores each action against a teacher acting on the exact posterior. The reward never reads the agent's belief line, but the update still trains the line along with the action. After training, the agent rarely commits while it has insufficient evidence and its belief line still names several answers. The policy learns from this reward using policy gradient or EM. It needs no judge, critic or reward model.
  \item \textbf{Experimental results.} On three vision-language models and three benchmarks, \method improves the base model in all nine settings,
  by 17.1\% on average and by 10.4\% with EM as the learner. It outperforms SFT and RL with the outcome reward in every
  setting. It also outperforms PPO with a process reward from an LLM judge on average and uses 1.6 times fewer GPU-hours, or
  4 times fewer with EM. Along with this gain, the share of answers given too early falls from 76--87\%
  to 44--46\%.
  \item \textbf{Why the training helps.} \method answers with sufficient evidence in 51\%, 12\% and 74\% of episodes on the three benchmarks. The base model does so in 8\%, 0.4\% and 36\% of episodes. Using a belief line and the exact posterior, we find that the trained agent acts on its own belief. It commits at only 0.1 to 2.2\% of the turns with insufficient evidence at which its belief line still names several answers. The base model with the same line commits on 3 to 68\% of those turns. When it commits, the trained agent gives the answer its line names, just like the base model. Across all turns, its belief line is not consistently more exact than the base model's. The single answers it names are often wrong. Its steps add roughly as much evidence as the base model's. In metacognitive terms, training mainly links decision control to the agent's own monitoring. Monitoring improves only at the states that both agents reach, where the line names a single answer too early less often. We expect a better reading of the evidence to lead to better answers.
  \item \textbf{Pro-Worlds Benchmark.} We introduce Pro-Worlds, a new benchmark of visual puzzles. The agent sees a large world only through a small window. It moves this window and must decide when it has seen enough to answer. Pro-Worlds tests this decision in visual worlds never seen in training.
\end{itemize}

\section{Related Work}
\label{sec:related}

\paragraph{RL for language and vision-language agents.}
RL with outcome rewards has improved the reasoning of language models~\cite{guo2025deepseek,team2025kimi,shao2024deepseekmath,yu2025dapo}. Self-training methods follow a similar idea, as they sample the model's own rationales and keep those that reach a correct answer~\cite{zelikman2022star,singh2024beyond}. For multi-turn agents, RAGEN~\cite{wang2025ragen} finds that sparse and delayed rewards fail to reinforce the reasoning of each step. GiGPO~\cite{feng2025gigpo} obtains a finer credit by grouping the steps that share a state. VAGEN~\cite{wang2025vagen} adds a turn-level reward for the agent's estimate of the world state. These rewards still come from the outcome or from a learned estimate, while \method computes its rewards from the exact posterior.

\paragraph{Process rewards.}
Process reward models score each step of a solution instead of only the final answer~\cite{lightman2024lets}. Most are trained on human or Monte Carlo labels, and their quality depends on how these labels are made~\cite{zhang2025lessons}. Others learn from outcomes alone~\cite{cui2025prime}, reason before they judge~\cite{khalifa2025thinkprm}, or derive the step reward from a regularized RL objective~\cite{yao2026prl}. Closer to our setting, PRInTS~\cite{lee2025prints} learns a reward model for long-horizon information seeking. Our process reward is computed exactly by the environment, and it needs no reward model, no judge and no step labels.

\paragraph{Belief states and information gathering.} The belief state is the posterior over the hidden state. It is a sufficient statistic for acting under partial observability~\cite{astrom1965optimal,kaelbling1998planning}. The value of an evidence-gathering action is the information gained from it~\cite{howard1966information,lindley1956measure}. BED-LLM~\cite{choudhury2026bedllm} asks the questions with the highest expected information gain at test time. Bayesian teaching~\cite{qiu2026bayesian} fine-tunes a model to imitate a Bayesian assistant. CURIO~\cite{wan2025curio} and ECHO~\cite{nath2026echo} reward each turn in a multi-turn interaction by how much it improves a belief over a hidden variable. Unlike these methods, \method uses a lookahead on the exact posterior to value the decision to stop. It also compares several turns at the same history. Agent-BRACE~\cite{singh2026agentbrace} is closest to our work. It trains an agent in text environments to state its belief as claims about the environment, each with a certainty word. Its policy acts on this belief instead of the history. An LLM judge grades these claims against the true hidden state rather than the posterior. When two answers are equally likely, the judge rewards a claim that names only the true answer, just as an outcome reward rewards a lucky guess. In \method, the agent is visual and states the set of answers it still sees as possible. The exact posterior scores its action and can also score this belief. As a result, the reward is verifiable and needs no judge.

\section{Method}
\label{sec:method}

\begin{figure*}[!htbp]
  \centering
  \includegraphics[width=\textwidth]{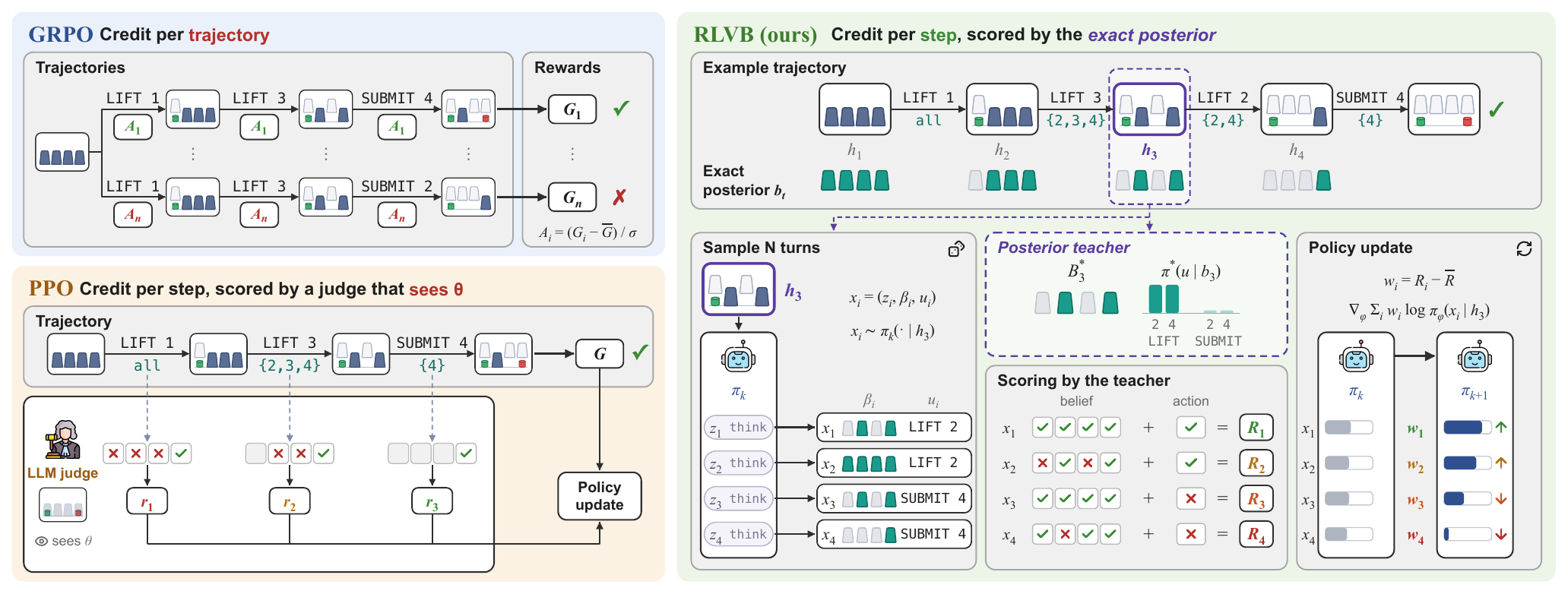}
  \caption{\textbf{Three RL frameworks on the cups task.} The red cylinder $\theta$ is under cup 4.
  (a) GRPO gives every step the advantage of its trajectory, and the two shared lifts therefore get opposite credit that depends only on the final guess.
  (b) In PPO, a judge that sees $\theta$ scores each step. It penalizes the correct belief \texttt{all} and rewards the lucky guess \texttt{\{4\}}.
  (c) \method scores the action of each turn sampled at $h_3$ against the teacher policy $\pi^*$, which acts on the exact posterior, and raises the turns with a reward above the group mean. The figure also shows a variant that scores each belief line against $B^*_3$. We report this variant as an ablation.}
  \label{fig:credit}
\end{figure*}

\Cref{fig:credit} compares the three RL frameworks of \cref{sec:intro} on the cups task.
GRPO~\cite{shao2024deepseekmath} learns from the outcome reward and gives every step the advantage of its trajectory.
PPO~\cite{schulman2017proximal} learns from a process reward given to every step by a judge that sees the target.
\method learns from a process reward given to every step by the exact posterior.
The outcome and the judge both depend on which cup happens to hide the object in this episode, while the posterior depends only on the evidence the agent has gathered.
\Cref{sec:problem} formulates the problem, \cref{sec:teacher,sec:score} define the posterior teacher and its process reward, and \cref{sec:learning} trains the agent on this reward.

\subsection{Problem formulation}
\label{sec:problem}

\paragraph{Exact posterior.}
A simulator knows more than the agent trained in it.
Prior work passes this privileged information to a teacher or a critic as the true state~\cite{pinto2018asymmetric,chen2019learning,choudhury2025better}, while we use it to compute the exact posterior over what the agent cannot see.
Each task has a hidden target $\theta \in \Theta$ drawn from a prior $p_0$, a correct answer $y^*(\theta) \in \mathcal Y$, and evidence actions $a_t \in \mathcal A$ with observations $o_t \sim p(\cdot \mid \theta, a_{1:t})$.
In the cups task of \cref{fig:credit}, $\theta$ is the cup hiding the red cylinder, an evidence action lifts a cup, and an answer names a cup.
The simulators of our tasks in \cref{sec:benchmarks} enumerate $\Theta$, and before every step $t$ they compute the exact posterior
\begin{equation}
  b_t(\theta) \propto p_0(\theta) \textstyle\prod_{s < t} p(o_s \mid \theta, a_{1:s}).
  \label{eq:belief}
\end{equation}
Since all our tasks have deterministic observations, $b_t$ is $p_0$ restricted to the targets consistent with every observation.
We write $b_t(y)$ for the posterior probability that $y$ is correct, $b_t(Y)$ for the mass of a set $Y \subseteq \mathcal Y$, and $B^*_t = \{y \in \mathcal Y : b_t(y) > 0\}$ for the answers still allowed by the evidence. We drop the index $t$ when the history is clear.
In \cref{fig:credit}(c), lifting cups 1 and 3 leaves $B^*_3 = \{2, 4\}$.

\paragraph{Agent.}
The agent is a vision-language model $\pi_\phi$ that reasons before it acts~\cite{wei2022chain,guo2025deepseek}.
At step $t$, it reads the task instruction and its history $h_t$ and writes a turn $x_t = (z_t, \beta_t, u_t) \sim \pi_\phi(\cdot \mid h_t)$.
The history holds the past actions and observations and the last belief line of the agent.
The turn consists of a reasoning $z_t$ in the agent's thinking block, a belief line $\beta_t$ and an action $u_t \in \mathcal A \cup \mathcal Y$.
The belief line names the set $B_t$ of answers still possible in the agent's view.
It is exact if it names exactly the answers allowed by the evidence, too narrow if it rules out one of them, and too wide if it keeps an answer ruled out by the evidence.
At $h_3$ in \cref{fig:credit}, the line \bline{Possible: cups 2, 4} is exact, \bline{Possible: cup 4} is too narrow and the open line \bline{Possible: all} is too wide.
Because the history keeps no earlier reasoning, the belief line carries the agent's belief from one turn to the next.
We call $d_t = (\beta_t, u_t)$ the decision of the turn and treat $z_t$ as a latent variable.

\paragraph{Objective.}
The agent and a task form a partially observable Markov decision process (POMDP)~\cite{kaelbling1998planning} whose hidden variable $\theta$ stays fixed during an episode.
An answer $y$ at step $T_{\text{stop}}$ ends the episode with the outcome reward
\begin{equation}
  G = c(T_{\text{stop}})\,\mathbf 1\big[y = y^*(\theta)\big], \qquad c(t) = e^{-\lambda t / T},
  \label{eq:outcome}
\end{equation}
where $\lambda$ is a time discount and $T$ is the step limit. An episode without an answer earns $0$. An optimal agent decides using only $b_t$ and $t$, because $(b_t, t)$ is a sufficient statistic of $h_t$ for $\mathbb E[G]$~\cite{astrom1965optimal}. Choosing when to answer is an optimal stopping problem~\cite{wald1947sequential,chow1971great}. The exact posterior lets us value every action with a lookahead on $b_t$. It also lets us verify every belief line against $B^*_t$. We want the agent to learn this stopping decision. It should keep going while it has insufficient evidence and answer once it has sufficient evidence. The action term also scores which evidence action the agent picks, but improving that choice is not the goal of this work. \Cref{sec:teacher,sec:score} turn this action value and belief line check into a reward for each step.

\subsection{The posterior teacher}
\label{sec:teacher}

The exact posterior gives two quantities at every history, and both depend only on the evidence.
One is the value of every next action, and these values define the teacher policy $\pi^*$ that scores the action.
The other is the set $B^*_t$ of answers still allowed by the evidence, and a variant also checks the belief line against this set.

The teacher values an answer $y$ by the log of its discounted chance of being right, and an evidence action $a$ by the expected value of the posterior $b^{a,o}$ after its observation $o$~\cite{howard1966information}:
\begin{equation}
\begin{alignedat}{2}
  &Q(b, y) &&= \log\big(c(t)\,\hat\rho(b)\, b(y)/b_{\max}\big),\\
  &Q^{(n)}(b, a) &&= \textstyle\sum_o p(o \mid b, a)\, V^{(n-1)}(b^{a,o}),\\
  &V^{(n)}(b) &&= \max_u Q^{(n)}(b, u),
\end{alignedat}
\label{eq:values}
\end{equation}
where $b_{\max} = \max_y b(y)$, $\hat\rho(b)$ is the measured answer accuracy of the agent, $u$ ranges over the answers and the evidence actions with $Q^{(n)}(b, y) = Q(b, y)$, and $n$ is the depth of the lookahead (supplementary material).
The teacher answers with probability $p_{\text{ans}} = \sigma(\Delta/\tau)$, where $\sigma$ is the logistic function and $\Delta$ the value of the best answer minus that of the best evidence action.
It chooses among the answers or among the evidence actions by a Boltzmann distribution with temperature $\tau$~\cite{haarnoja2017reinforcement}:
\begin{equation}
\begin{aligned}
  \pi^*(y \mid b) &= p_{\text{ans}}\, \operatorname{softmax}_y\big(Q(b, \cdot)/\tau\big),\\
  \pi^*(a \mid b) &= (1 - p_{\text{ans}})\, \operatorname{softmax}_a\big(Q^{(n)}(b, \cdot)/\tau\big).
\end{aligned}
\label{eq:teacher}
\end{equation}

The teacher depends on the agent only through $\hat\rho$, which is measured from the agent's own answers. Where these answers are unreliable, answering has a low value and the teacher asks for more evidence. The stopping rule of the teacher thus fits the agent (supplementary material).

In \cref{fig:credit}(c), $b_3$ splits evenly between cups 2 and 4.
Answering now is right with probability at most $1/2$, while lifting either cup reveals where the cylinder is and lets the agent answer correctly one step later.
The teacher therefore puts almost all of $\pi^*(u \mid b_3)$ on the two lifts.

\subsection{Posterior process rewards}
\label{sec:score}

The teacher gives each turn a reward at the history where the agent wrote it. The reward uses the exact belief $b$, which the simulator can verify. The main term checks the action against $b$. A second term, the belief term, directly checks the belief stated in the agent's line:
\begin{equation}
  R(h, x) = \underbrace{\log \pi^*(u \mid b)}_{\text{action}} + \underbrace{S_{\text{dec}} + \eta\,(S + W)}_{\text{stated belief}},
  \label{eq:score}
\end{equation}
where $\eta$ is a small weight. The action term is the action's log probability under the teacher. The teacher acts on the exact belief $b$. As a result, the action term rewards taking more steps while $b$ still allows several answers. Once $b$ no longer allows several answers, it rewards giving an answer. In this way, the action term is a belief-based reward. It uses the belief that the simulator can verify. It does not use the line written by the agent. The agent commits when the action $u$ is an answer. It keeps working when $u$ is an evidence action. Therefore, the action term rewards both the decision to commit or keep working and the choice of evidence action.

The action term penalizes the two possible mistakes very differently: answering before the evidence is sufficient receives a very low reward ($-7.42$ in \cref{fig:turn}), while taking one additional step after the evidence is sufficient incurs only the small time discount of a single step. The reward therefore primarily teaches the agent not to answer too early.

\method learns from the verifiable belief $b$ through the action term. In every turn, the agent writes its belief line right before its action. Because the update increases the likelihood of the whole turn, it trains the line together with the action. As \cref{sec:more_looks} shows, the agent then learns to commit or keep working as its own belief line says. This is the main idea of \method.

The belief term is a variant that also rewards the line itself. We train \method (EM) with both terms, and we train \method with the action term alone. We report the belief term as an ablation in \cref{sec:ablation}. It does not raise accuracy.
The belief term reads the line as the uniform distribution $b_\beta$ over $B$, with $\log b_\beta(y) = -m$ for an answer outside $B$.
It combines three scores of $\beta$ against $b$:
\begin{equation}
\begin{alignedat}{2}
  &S(\beta; b) &&= \textstyle\sum_y b(y) \log b_\beta(y),\\
  &W(\beta; b) &&= -m\,|B \setminus B^*|/|B|,\\
  &S_{\text{dec}}(\beta; b) &&= \log \pi^*(a_\beta \mid b),
\end{alignedat}
\label{eq:terms}
\end{equation}
where $a_\beta = \arg\max_u \pi^*(u \mid b_\beta)$ is the teacher's most probable action if the line were the posterior.

$S$ is the log score of the line~\cite{good1952rational}, but it punishes a line that is too narrow far more than one that is too wide.
On its own, $S$ would teach the agent to hedge.
We therefore score the line mainly by its implied decision.
$S_{\text{dec}}$ is a proper scoring rule of the teacher's decision problem~\cite{grunwald2004game,gneiting2007strictly}, and $W$ charges a line that is too wide for the answers ruled out by the evidence.
The small weight $\eta$ on $S + W$ breaks ties between lines that imply the same action.
Since $R$ depends only on the posterior and the decision, no later observation enters it.
\Cref{fig:turn} gives both terms for the four turns at $h_3$ in \cref{fig:credit}(c).
The action term gives both lifts $0$ and both answers $-7.42$, whatever their belief line.
The belief term then ranks the exact line above the open line and the guess \bline{Possible: cup 4}.

\subsection{Policy update}
\label{sec:learning}

\begin{algorithm}[t]
\caption{Training with posterior process rewards}
\label{alg:main}
\begin{algorithmic}[1]
\Statex \textbf{Input:} base model $\pi_0$, training tasks, number of phases $k_{\max}$, turns per history $N$, step limit $\varepsilon$.
\Statex \textbf{Output:} trained model $\pi_\phi$.
\For{$k = 0, \ldots, k_{\max} - 1$}
  \State Fix $\pi_k \gets \pi_\phi$ and play episodes with $\pi_k$.
  \For{each history $h$ of these episodes}
    \State Sample $N$ turns $x_i \sim \pi_k(\cdot \mid h)$ and score each by $R$ in \cref{eq:score}.
    \State Weight the turns against each other by \cref{eq:adv} or by EM.
  \EndFor
  \State Take gradient steps on \cref{eq:wll} until the KL divergence from $\pi_k$ exceeds $\varepsilon$.
\EndFor
\end{algorithmic}
\end{algorithm}

\method updates the policy one step at a time (\cref{fig:credit}(c) and \cref{alg:main}), and we train it in phases.
In phase $k$, a fixed copy $\pi_k$ of the model plays episodes, and at every history $h$ of these episodes we sample $N$ turns $x_i \sim \pi_k(\cdot \mid h)$ and score each by \cref{eq:score}.
The teacher's action replaces the agent's in these episodes with a probability that starts at 0.5 and decays from phase to phase, as in DAgger~\cite{ross2011reduction}. On half of the training scenes, one more episode starts with random evidence actions, and the histories then also cover states rarely reached by the agent's own actions.
The update compares these $N$ decisions at the same history, and a decision is thus credited only against alternatives under the same evidence.
Trajectory RL such as GRPO~\cite{shao2024deepseekmath} instead credits every step of an episode with the advantage of the whole episode, and the two gradient estimates differ in exactly this respect:
\begin{equation}
\begin{aligned}
  \text{trajectory:}\quad & \textstyle\sum_e \sum_t A_e\, \nabla_\phi \log \pi_\phi(x_{e,t} \mid h_{e,t}),\\
  \text{step:}\quad & \textstyle\sum_h \sum_i w_i\, \nabla_\phi \log \pi_\phi(x_i \mid h),
\end{aligned}
\label{eq:estimators}
\end{equation}
where $A_e$ is shared by all the steps $t$ of episode $e$, and $w_i$ weighs turn $i$ against the other turns at $h$.

Each phase moves the model toward the improved policy~\cite{peters2010relative,levine2018reinforcement,donsker1975asymptotic}
\begin{equation}
  q_k(x \mid h) \propto \pi_k(x \mid h)\, e^{R(h,x)/\alpha},
  \label{eq:gibbs}
\end{equation}
which maximizes $\mathbb E_\pi R - \alpha\,\mathrm{KL}(\pi \,\|\, \pi_k)$ at every history.
The model moves toward $q_k$ by gradient steps on the weighted log likelihood
\begin{equation}
  \mathcal L(\phi) = -\textstyle\sum_h \sum_{i=1}^{N} w_i \log \pi_\phi(x_i \mid h).
  \label{eq:wll}
\end{equation}
We use two weightings, the first for \method and the second for its variant \method (EM).
\method uses the policy gradient~\cite{williams1992simple}. It is group relative as in GRPO, but its group is the $N$ turns at one history.
Each turn is weighted by its reward minus the mean reward of the other turns~\cite{ahmadian2024back}:
\begin{equation}
  w_i \propto R(h, x_i) - \tfrac{1}{N-1} \textstyle\sum_{j \ne i} R(h, x_j).
  \label{eq:adv}
\end{equation}
This weight is proportional to $R(h, x_i) - \bar R(h)$ in \cref{fig:credit}(c).
\method (EM) instead uses expectation maximization (EM)~\cite{dempster1977maximum,abdolmaleki2018maximum} and weights the turns by $w_i \propto e^{R(h, x_i)/\alpha}$, normalized over the history.
These normalized weights estimate $q_k$ itself.
The two weightings are two limits of one family (supplementary material).
To keep each step small, a phase ends when the KL divergence of $\pi_\phi$ from $\pi_k$ exceeds $\varepsilon$.
Each phase is then one step of policy mirror descent on $R$~\cite{lan2023policy,tomar2022mirror}.
\method gives a turn its own reward only, because the lookahead of the teacher already values what follows.
GRPO with process supervision~\cite{shao2024deepseekmath} instead adds the rewards of all later steps to a step.
With both weightings, a turn gets no positive weight if the teacher rejects its action, that is, if $\pi^*(u \mid b) < \pi^*(u_{\text{best}} \mid b)/20$ for the teacher's most probable action $u_{\text{best}}$.
A small length reward favors the shorter reasoning among correct turns, as in Kimi k1.5~\cite{team2025kimi}.
The supplementary material gives the rules for sampling, acceptance and fitting.

\section{Experiments}
\label{sec:experiments}

\subsection{Benchmarks}
\label{sec:benchmarks}

We evaluate on three benchmarks: HiddenGames, Pro-Worlds and RoboQuest~\cite{liu2026roboquest}. In all benchmarks, the answer is hidden. The agent must take actions to collect evidence and decide when it has seen enough to answer.

\paragraph{HiddenGames.}
It consists of seven tasks from existing agent benchmarks: MineSweeper~\cite{wang2025popgym}, MiniGrid KeyCorridor~\cite{chevalier2023minigrid}, FrozenLake and Sokoban~\cite{wang2025vagen}, and three Tabletop tasks rendered in ManiSkill3~\cite{tao2024maniskill3}. We train on these tasks and test on 756 of their games with seeds, maps and colors never seen in training.

\paragraph{Pro-Worlds.}
To assess transfer to out-of-domain visual tasks, we introduce Pro-Worlds, a benchmark that tests how visual agents gather, retain and combine evidence before they commit to an answer. In each game, the agent sees a larger world only through a small window, moves the window to collect evidence, and must decide when it has seen enough to answer. The window shows one site of the world at a time. The agent can make at most 35 moves before it names one of eight candidates (ten in Jigsaw) as its answer. We compute the exact posterior of every game from its game engine. A teacher acting on this posterior solves every game in 14.7 moves on average. We test on 15 games (Face, Jigsaw, Kaleido, Bird, Fish, Recipe, Turtle, Balloon, Rocket, Bell, Bee, Candle, Sailboat, Barn and Lantern), with 20 plays of each.

\paragraph{RoboQuest.}
To test transfer to tasks without an exact posterior, we use the ten mobile manipulation tasks of RoboQuest~\cite{liu2026roboquest}. They take place in simulated kitchens and require search, inspection and interactive testing, such as finding the odd parcel by weighing. We test on 50 games of each task. The supplementary material gives further details of all three benchmarks.

\subsection{Experiment Setup}
\label{sec:setup}

\paragraph{Models.}
We train three open-source vision-language models: Qwen3-VL-4B-Thinking and Qwen3-VL-8B-Thinking~\cite{bai2025qwen3vl}, and Qwen3.5-4B~\cite{qwen2026qwen35}. 

\paragraph{Baselines.}
\emph{Base} is the untrained model and outputs an action only. \emph{Base (w/ belief)} is the same model with a belief line before the action. We consider two groups of trained baselines. The first group acts without a belief and learns from whole episodes. \emph{SFT} fine-tunes the base model on the teacher's actions in winning games. \emph{GRPO}~\cite{shao2024deepseekmath} and \emph{PPO}~\cite{schulman2017proximal} train on the outcome reward of each episode. \emph{PPO} also learns a critic. The second group learns from a process reward. In \emph{PPO (belief + action)}, the model writes its belief as short claims with certainty words and then an action. PPO trains it with the outcome reward of the episode. A learned value function carries this reward back to every turn. At each turn, an LLM judge (DeepSeek V4.1 Flash~\cite{deepseek2026v41flash}) rewards the belief by checking its claims against the newest observation and the true hidden state. Only the belief gets a per-turn reward. The decision to continue or to answer is rewarded only by the outcome of the episode. Both rewards of this baseline are based on the truth. The reward of \method is instead based on the evidence. The outcome reward pays a correct answer even when it was a guess. The judge rewards a true claim even when the evidence does not support it. PPO (belief + action) also samples whole episodes and learns from one outcome per episode. \method instead samples multiple turns at each history and rewards each of them against the exact posterior. All methods train LoRA adapters~\cite{hu2022lora} for up to 50 updates on the same tasks. At every turn, every method receives the full history of its actions and observations but not its earlier reasoning.

\paragraph{Metrics.}
We test the adapter after the last update of each run. Each test runs three times with different sampling seeds. We report the mean accuracy over the three runs\footnote{The accuracy of \method varies by up to 4\% across training runs. When trained three times on Qwen3.5-4B with small changes to the settings, the three runs reached 48.5 to 50.0\% on HiddenGames, 28.3 to 32.1\% on Pro-Worlds, and 54.0 to 56.9\% on RoboQuest.}. Accuracy is the share of episodes ending with the correct answer. The supplementary material lists the settings of \method.

\subsection{Main results}
\label{sec:main}

\Cref{tab:main} reports the accuracy of every method.

\begin{table*}[!htbp]
  \centering
  \footnotesize
  \setlength{\tabcolsep}{2.4pt}
  \begin{adjustbox}{max width=\textwidth}%
  \begin{tabular}{@{}lc *{9}{r@{\hspace{1.5pt}}l}@{}}
    \toprule
    & & \multicolumn{6}{c}{Qwen3-VL-8B-Thinking} & \multicolumn{6}{c}{Qwen3-VL-4B-Thinking} & \multicolumn{6}{c}{Qwen3.5-4B} \\
    \cmidrule(lr){3-8} \cmidrule(lr){9-14} \cmidrule(l){15-20}
    Method & Belief & \multicolumn{2}{c}{HiddenGames} & \multicolumn{2}{c}{Pro-Worlds} & \multicolumn{2}{c}{RoboQuest} & \multicolumn{2}{c}{HiddenGames} & \multicolumn{2}{c}{Pro-Worlds} & \multicolumn{2}{c}{RoboQuest} & \multicolumn{2}{c}{HiddenGames} & \multicolumn{2}{c}{Pro-Worlds} & \multicolumn{2}{c}{RoboQuest} \\
    \midrule
    \multicolumn{20}{@{}l}{\textit{Reference}} \\
    Posterior teacher & & 97.5 &  & 100.0 &  & -- &  & 97.5 &  & 100.0 &  & -- &  & 97.5 &  & 100.0 &  & -- &  \\
    \midrule
    \multicolumn{20}{@{}l}{\textit{Base}} \\
    Base &  & 22.9 &  & 14.8 &  & 36.2 &  & 24.6 &  & 17.2 &  & 28.4 &  & 26.5 &  & 20.2 &  & 39.6 &  \\
    Base (w/ belief) & \checkmark & 23.9 & \dlt{+1.0} & 20.5 & \dlt{+5.7} & 39.8 & \dlt{+3.6} & 24.5 & \dlt{$-$0.1} & 18.2 & \dlt{+1.0} & 34.0 & \dlt{+5.6} & 38.1 & \dlt{+11.6} & 23.9 & \dlt{+3.7} & 41.3 & \dlt{+1.7} \\
    \midrule
    \multicolumn{20}{@{}l}{\textit{Baselines without a process reward}} \\
    SFT (action-only) &  & 27.0 & \dlt{+4.1} & 19.3 & \dlt{+4.5} & 42.5 & \dlt{+6.3} & 31.2 & \dlt{+6.6} & 18.4 & \dlt{+1.2} & 34.5 & \dlt{+6.1} & 26.9 & \dlt{+0.4} & 13.6 & \dlt{$-$6.6} & 38.3 & \dlt{$-$1.3} \\
    PPO (action-only) &  & 23.7 & \dlt{+0.8} & 12.6 & \dlt{$-$2.2} & 39.2 & \dlt{+3.0} & 26.5 & \dlt{+1.9} & 14.4 & \dlt{$-$2.8} & 20.9 & \dlt{$-$7.5} & 22.9 & \dlt{$-$3.6} & 11.6 & \dlt{$-$8.6} & 12.6 & \dlt{$-$27.0} \\
    GRPO (action-only) &  & 23.1 & \dlt{+0.2} & 14.9 & \dlt{+0.1} & 34.7 & \dlt{$-$1.5} & 24.9 & \dlt{+0.3} & 15.2 & \dlt{$-$2.0} & 24.3 & \dlt{$-$4.1} & 32.5 & \dlt{+6.0} & 15.1 & \dlt{$-$5.1} & 36.9 & \dlt{$-$2.7} \\
    \midrule
    \multicolumn{20}{@{}l}{\textit{Baseline with a belief-based process reward}} \\
    PPO (belief + action) & \checkmark & 26.4 & \dlt{+3.5} & 16.9 & \dlt{+2.1} & 51.4 & \dlt{+15.2} & 25.9 & \dlt{+1.3} & 15.1 & \dlt{$-$2.1} & 38.7 & \dlt{+10.3} & 40.5 & \dlt{+14.0} & 29.1 & \dlt{+8.9} & \textbf{57.1} & \dlt{+17.5} \\
    \midrule
    \multicolumn{20}{@{}l}{\textit{Ours}} \\
    \method (EM) & \checkmark & 29.0 & \dlt{+6.1} & \textbf{32.4} & \dlt{+17.6} & 47.2 & \dlt{+11.0} & 31.5 & \dlt{+6.9} & 23.1 & \dlt{+5.9} & 42.6 & \dlt{+14.2} & 42.9 & \dlt{+16.4} & 28.1 & \dlt{+7.9} & 47.3 & \dlt{+7.7} \\
    \method & \checkmark & \textbf{31.1} & \dlt{+8.2} & 29.9 & \dlt{+15.1} & \textbf{62.7} & \dlt{+26.5} & \textbf{36.2} & \dlt{+11.6} & \textbf{28.7} & \dlt{+11.5} & \textbf{61.2} & \dlt{+32.8} & \textbf{48.5} & \dlt{+22.0} & \textbf{32.1} & \dlt{+11.9} & 54.0 & \dlt{+14.4} \\
    \bottomrule
  \end{tabular}
  \end{adjustbox}
  \caption{\textbf{Accuracy (\%) of three vision-language models on the three benchmarks.} HiddenGames tests unseen games of the training tasks. Pro-Worlds and RoboQuest are never seen in training. The rows group the methods by training signal: none, imitation or the outcome reward, a belief-based process reward from an LLM judge, and the posterior process reward of \method. A check mark under Belief means that the agent states a belief before each action. Numbers are means over three sampling seeds. Grey numbers give the change against Base. Bold marks the best model in each column. The posterior teacher acts on the exact posterior, which is not available in RoboQuest.}
  \label{tab:main}
\end{table*}

\paragraph{Stating a belief.} Stating a belief helps even without training. Base (w/ belief) is better than Base in eight of the nine settings. It improves over Base by 3.8\% on average, and by 11.6\% with Qwen3.5-4B on HiddenGames. However, stating a belief does not stop early answers. Base (w/ belief) still commits on 3 to 68\% of turns where it has insufficient evidence and its own belief line names several answers (\cref{sec:more_looks}).

\paragraph{Comparison with the baselines.} \method has the highest accuracy. \method or \method (EM) is the best method in eight of the nine columns of \cref{tab:main}. Both learners improve Base in every setting, by 17.1\% on average with \method and by 10.4\% with \method (EM). Both also outperform SFT, GRPO and PPO (action-only) in every setting. The gains transfer to the two benchmarks never seen in training, by up to 17.6\% on Pro-Worlds and up to 32.8\% on RoboQuest. In contrast, the outcome reward helps little. On average, GRPO (action-only) loses 1.0\% and PPO (action-only) loses 5.1\%. The gain comes mainly from the posterior reward, because two different learners both give large gains with it. \method and GRPO (action-only) use the same learner, which weights each sample against its group. With this learner, the posterior process reward outperforms the outcome reward in all nine settings, by 18.1\% on average.

\paragraph{Judged process reward.} \method is also better than a judged process reward. PPO (belief + action) gains 7.9\% over Base on average. \method does better than it in eight of the nine settings, by 9.3\% on average. The only exception is RoboQuest with Qwen3.5-4B. To compare their behavior, we replay every test episode and compute the exact posterior at each turn. We report the results in \cref{tab:process}. Across all nine settings, \method produces the fewest premature answers and takes the most steps. On HiddenGames, 34.5--43.1\% of answers from \method are premature, compared to 58.9--94.3\% for PPO (belief + action) and 64.2--95.1\% for Base (w/ belief). However, its belief line is not the most often exact in any of the six settings. \Cref{sec:more_looks} shows that this gain comes from acting on the belief line. The agent rarely commits while the line still names several answers.

\begin{table*}[!htbp]
  \centering
  \footnotesize
  \setlength{\tabcolsep}{2.45pt}
  \resizebox{\textwidth}{!}{
  \begin{tabular}{@{}l rrrrr c rrrrr c rrr@{}}
    \toprule
    & \multicolumn{5}{c}{HiddenGames} & & \multicolumn{5}{c}{Pro-Worlds} & & \multicolumn{3}{c}{RoboQuest} \\
    \cmidrule(lr){2-6} \cmidrule(lr){8-12} \cmidrule(l){14-16}
    & \multicolumn{4}{c}{Behavior} & Belief & & \multicolumn{4}{c}{Behavior} & Belief & & \multicolumn{3}{c}{Behavior} \\
    \cmidrule(lr){2-5} \cmidrule(lr){6-6} \cmidrule(lr){8-11} \cmidrule(lr){12-12} \cmidrule(l){14-16}
    Method & Correct$\uparrow$ & Prem.$\downarrow$ & Steps & Post.$\uparrow$ & Exact$\uparrow$ && Correct$\uparrow$ & Prem.$\downarrow$ & Steps & Post.$\uparrow$ & Exact$\uparrow$ && Correct$\uparrow$ & Prem.$\downarrow$ & Steps \\
    \midrule
    \multicolumn{16}{@{}l}{\textbf{Qwen3.5-4B}} \\[1pt]
    \quad Base & 26.5 & 80.8 & 1.9 & 31.7 & -- && 20.2 & 98.0 & 12.4 & 19.6 & -- && 39.6 & 52.8 & 5.5  \\
    \quad Base (w/ belief) & 38.1 & 64.2 & 3.8 & 41.7 & 47.8 && 23.9 & 95.8 & 18.0 & 25.4 & \textbf{82.4} && 41.3 & 48.5 & 6.0   \\
    \quad PPO (belief + action) & 40.5 & 58.9 & 4.5 & 47.4 & -- && 29.1 & 91.0 & 24.1 & 29.7 & -- && \textbf{57.1} & 29.5 & 7.8   \\
    \quad \method (EM) & 42.9 & 48.7 & 7.2 & 52.8 & \textbf{60.4} && 28.1 & 88.2 & 25.4 & 31.8 & 78.9 && 47.3 & 40.9 & 6.6   \\
    \quad \method & \textbf{48.5} & \textbf{34.7} & 8.2 & \textbf{65.0} & 60.3 && \textbf{32.1} & \textbf{83.7} & 29.3 & \textbf{33.6} & 75.0 && 54.0 & \textbf{28.9} & 8.0   \\
    \quad \method, with the belief term & 46.5 & 43.0 & 6.6 & 54.6 & 53.3 && 32.0 & 90.1 & 27.9 & 30.5 & 76.4 && 48.7 & 36.1 & 7.3 \\
    \midrule
    \multicolumn{16}{@{}l}{\textbf{Qwen3-VL-4B}} \\[1pt]
    \quad Base & 24.6 & 93.1 & 3.3 & 21.4 & -- && 17.2 & 99.9 & 10.6 & 16.3 & -- && 28.4 & 72.1 & 3.1  \\
    \quad Base (w/ belief) & 24.5 & 95.1 & 3.6 & 21.5 & 46.8 && 18.2 & 100.0 & 9.8 & 17.2 & 77.1 && 34.0 & 61.8 & 4.2   \\
    \quad PPO (belief + action) & 25.9 & 94.3 & 3.8 & 23.4 & -- && 15.1 & 100.0 & 3.9 & 14.0 & -- && 38.7 & 59.3 & 4.7   \\
    \quad \method (EM) & 31.5 & 77.4 & 4.3 & 27.4 & 51.0 && 23.1 & 98.2 & 17.3 & 22.4 & 77.0 && 42.6 & 52.4 & 5.0   \\
    \quad \method & \textbf{36.2} & \textbf{34.5} & 8.3 & \textbf{44.6} & 41.0 && \textbf{28.7} & \textbf{90.4} & 31.7 & \textbf{29.5} & 69.3 && \textbf{61.2} & \textbf{21.8} & 10.0   \\
    \quad \method, with the belief term & 35.9 & 52.9 & 6.2 & 38.5 & \textbf{56.9} && 21.7 & 98.3 & 21.6 & 22.7 & \textbf{81.6} && 45.8 & 41.9 & 6.4 \\
    \midrule
    \multicolumn{16}{@{}l}{\textbf{Qwen3-VL-8B}} \\[1pt]
    \quad Base & 22.9 & 97.4 & 3.1 & 19.8 & -- && 14.8 & 100.0 & 5.3 & 15.4 & -- && 36.2 & 62.9 & 4.7  \\
    \quad Base (w/ belief) & 23.9 & 88.9 & 3.5 & 20.6 & 48.4 && 20.5 & 100.0 & 10.6 & 18.2 & \textbf{75.7} && 39.8 & 56.0 & 5.0   \\
    \quad PPO (belief + action) & 26.4 & 81.4 & 3.8 & 26.2 & -- && 16.9 & 99.0 & 10.0 & 17.4 & -- && 51.4 & 37.6 & 7.4   \\
    \quad \method (EM) & 29.0 & 81.3 & 4.1 & 23.6 & 56.1 && \textbf{32.4} & 98.8 & 16.6 & 22.1 & 71.8 && 47.2 & 47.3 & 5.9   \\
    \quad \method & \textbf{31.1} & \textbf{43.1} & 8.4 & \textbf{38.3} & 45.5 && 29.9 & \textbf{88.5} & 33.9 & \textbf{31.5} & 70.8 && \textbf{62.7} & \textbf{20.0} & 10.1   \\
    \quad \method, with the belief term & 27.9 & 50.6 & 7.3 & 31.3 & \textbf{69.9} && 29.9 & 90.1 & 32.6 & 30.4 & 74.6 && 58.0 & 25.3 & 9.2 \\
    \bottomrule
  \end{tabular}
  }
  \caption{Behavior and belief accuracy. \emph{Correct}: success rate, as in \cref{tab:main}. \emph{Prem.}: share of answers given too early, that is, while the posterior of the top answer was below 0.95 (in RoboQuest, while the feedback had not yet met every requirement of the task). \emph{Steps}: evidence-gathering actions per episode. \emph{Post.}: the mean exact posterior of the agent's answer when it answers, that is, how likely the answer is to be right given what the agent has seen. For example, an answer given while two answers are still equally
possible counts as 50\%. On HiddenGames, \emph{Prem.}, \emph{Post.} and \emph{Exact} count the
  five tasks with an answer. \emph{Exact}: share of turns at which the belief line names exactly the answers still possible under the exact posterior, with any line that cannot be read counted as not exact. \method with the belief term also rewards the belief line against the exact posterior, as in \cref{tab:belief_term}. PPO (belief + action) writes its beliefs as free text that cannot be read as a set of answers. Best per model in bold.}
  \label{tab:process}
\end{table*}

\subsection{Ablation}
\label{sec:ablation}

\paragraph{The sample set and the two terms of the reward.}
\Cref{tab:ablation} reports runs that change one setting of \method at a time on Qwen3-VL-4B-Thinking. In every run, the agent still writes
a belief line before each action and receives it at the next turn. Only the training changes. \method trains with the action term of \cref{eq:score}.
\begin{table}[!htbp]
  \centering
  \footnotesize
  \setlength{\tabcolsep}{2.2pt}
  \begin{tabular}{@{}l ccc ccc@{}}
    \toprule
    & \multicolumn{2}{c}{Reward} & Sample & & & \\
    \cmidrule(lr){2-3}
    Method & Belief & Action & set & HiddenGames & Pro-Worlds & RoboQuest \\
    \midrule
    Base (w/ belief)               & --         & --         & --    & 24.5          & 18.2          & 34.0 \\
    \method                 & --         & \checkmark & large & \textbf{36.2} & \textbf{28.7} & \textbf{61.2} \\
    \quad small sample set         & --         & \checkmark & small & 35.3          & 23.9          & 55.5 \\
    \quad\quad with the belief term & \checkmark & \checkmark & small & 35.9          & 21.7          & 45.8 \\
    \quad\quad belief term only    & \checkmark & --         & small & 24.6          & 14.6          & 19.7 \\
    \bottomrule
  \end{tabular}
  \caption{\textbf{The sample set and the two terms of the reward} (Qwen3-VL-4B-Thinking). The numbers are the mean accuracy in percent over three test runs. \emph{Large}: 1,200 histories are sampled
  in each phase and 800 of them are used for training. \emph{Small}: 300 histories are sampled and 200 are used. With the belief
  term only, the acceptance rule is also off. Best of each column in bold.}
  \label{tab:ablation}
\end{table}

We first reduce the sample set to 300 histories and use 200 of them. The accuracy then stays about the same on HiddenGames, at 35.3\% against 36.2\%, but falls from 28.7\% to 23.9\% on Pro-Worlds and from 61.2\% to 55.5\% on RoboQuest. The larger sample set therefore helps on the two benchmarks never seen in training.

Next, we keep the small sample set and add the belief term. The accuracy does not rise with this term. It stays about the same on HiddenGames, at 35.9\% against 35.3\%, and falls from 23.9\% to 21.7\% on Pro-Worlds and from 55.5\% to 45.8\% on RoboQuest.

Finally, we keep the belief term and remove the action term. The reward of a turn is then its belief term alone. We also turn off the acceptance rule of \cref{sec:learning}, because it uses the teacher's probability of the action. The accuracy is then the same as that of Base (w/ belief) on HiddenGames, at 24.6\% against 24.5\%. It is lower on Pro-Worlds, at 14.6\% against 18.2\%, and on RoboQuest, at 19.7\% against 34.0\%. The gain of \method therefore comes from the action term. A reward on the belief line alone does not teach the agent when to continue and when to answer.

\paragraph{Adding a process reward.}
PPO (action-only) and PPO (belief + action) use the same learner and the same outcome reward. Only PPO (belief + action) adds a process reward from an LLM judge that scores the belief at every turn. In \cref{tab:main}, this process reward raises accuracy in eight of the nine settings, by 13.0\% on average. The two runs also differ in the turn format, because only PPO (belief + action) states a belief. In the two base rows, the belief line alone changes accuracy by 3.8\% on average. In six of the nine settings, the gap between the two runs is larger than this change. This suggests that most of the gain comes from the process reward.

\subsection{Analysis}
\label{sec:analysis}

\paragraph{Training time.}
\Cref{tab:cost} gives the cost of each training run. On Qwen3.5-4B with A100 GPUs, PPO (belief + action) uses 1.6$\times$ as many GPU-hours as \method (120.1 against 74.2) and also
makes about 12,600 calls to an LLM judge. \method (EM) uses 29.7 GPU-hours, which is four
times less than PPO (belief + action). \method is cheaper because its reward is computed from the exact posterior and needs no value network and
no reward model. PPO (belief + action) instead trains a value network next to the policy and asks an LLM judge to score every belief. At this lower
cost, \method is still more accurate than PPO (belief + action) in eight of the nine columns of
\cref{tab:main}.

\begin{table}[!htbp]
  \centering
  \footnotesize
  \setlength{\tabcolsep}{0.8\tabcolsep}
 \begin{tabular}{@{}l r r@{}}
    \toprule
    Method & GPU-hours & Judge (calls) \\
    \midrule
    \multicolumn{3}{@{}l}{\emph{Qwen3.5-4B, A100 80GB}} \\
    \quad PPO (belief + action) & 120.1 & \$9.8 (12.6k) \\
    \quad \method (EM) & 29.7 & -- \\
    \quad \method & 74.2 &--\\
    \midrule
    \multicolumn{3}{@{}l}{\emph{Qwen3-VL-4B, H200 141GB}} \\
    \quad \method (EM) & 13.6 & -- \\
    \quad \method & 30.7 & -- \\
    \midrule
    \multicolumn{3}{@{}l}{\emph{Qwen3-VL-8B, H200 141GB}} \\
    \quad \method (EM) & 15.0 & -- \\
    \quad \method & 46.2 & -- \\
    \bottomrule
  \end{tabular}
  \caption{Training cost of one run. Each run has 50 updates, but the step KL limit ended the runs of \method after 20 to 35 updates. The runs of one model use the same type of GPU. The last column gives the API cost of the LLM judge and its number of calls.}
  \label{tab:cost}
\end{table}

\paragraph{\method (EM) against \method.}
The two learners both yield large gains, reaching a mean accuracy of 42.7\% for \method and 36.0\% for \method (EM). \method (EM) outperforms \method in one of the nine columns of \cref{tab:main}, while in \cref{tab:process}, \method produces fewer premature answers than \method (EM) across all three benchmarks. Because \method (EM) retains the belief term, the two rows also differ in their reward. In the rows with the belief term in \cref{tab:belief_term}, where \method is trained with the same reward as \method (EM), its mean accuracy is 38.5\%, compared to 36.0\% for \method (EM). The learner of \method is therefore better under either reward. \method (EM) is the more efficient learner, because it needs only 32--44\% of the GPU-hours of \method to reach its accuracy (\cref{tab:cost}).

\begin{table}[!htbp]
  \centering
  \footnotesize
  \setlength{\tabcolsep}{3pt}
  \begin{tabular}{@{}l ccc cc cc cc@{}}
    \toprule
     & \multicolumn{3}{c}{Accuracy} & \multicolumn{2}{c}{Line exact} & \multicolumn{4}{c}{Answers when the line names one answer} \\
    \cmidrule(lr){2-4}\cmidrule(lr){5-6}\cmidrule(l){7-10}
     & & & & & & \multicolumn{2}{c}{HiddenGames} & \multicolumn{2}{c}{Pro-Worlds} \\
    \cmidrule(lr){7-8}\cmidrule(l){9-10}
    Training & HG & PW & RQ & HG & PW & one left & several left & one left & several left \\
    \midrule
    \multicolumn{10}{@{}l}{\textbf{Qwen3-VL-4B}} \\
    \quad with the belief term    & 35.9 & 21.7 & 45.8 & 57 & 82 & 98 & 95 & 92 & 84 \\
    \quad without the belief term & 35.3 & 23.9 & 55.5 & 39 & 60 & 83 & 44 & 59 & 30 \\
    \quad \method & 36.2 & 28.7 & 61.2 & 41 & 69 & 66 & 32 & 47 & 27 \\
    \midrule
    \multicolumn{10}{@{}l}{\textbf{Qwen3-VL-8B}} \\
    \quad with the belief term    & 27.9 & 29.9 & 58.0 & 70 & 75 & 94 & 91 & 55 & 48 \\
    \quad \method & 31.1 & 29.9 & 62.7 & 46 & 71 & 84 & 57 & 37 & 26 \\
    \midrule
    \multicolumn{10}{@{}l}{\textbf{Qwen3.5-4B}} \\
    \quad with the belief term    & 46.5 & 32.0 & 48.7 & 53 & 76 & 96 & 92 & 95 & 84 \\
    \quad \method & 48.5 & 32.1 & 54.0 & 60 & 75 & 91 & 57 & 89 & 67 \\
    \bottomrule
  \end{tabular}
  \caption{\textbf{What the belief term changes.} All numbers are in percent. \method is trained without the belief term. \emph{Accuracy}: as in \cref{tab:main} (HG: HiddenGames, PW:
  Pro-Worlds, RQ: RoboQuest). \emph{Line exact}: share of the turns when the belief line names exactly the answers still allowed by the
  posterior. As in \cref{tab:process}, a line that cannot be read counts as not exact. \emph{Answers when the line names one answer}: share of these turns when the agent answers, split by whether
  the posterior also allows only one answer (\emph{one left}) or still allows several (\emph{several left}). The
  HiddenGames columns of the last two groups use the five tasks with an answer.}
  \label{tab:belief_term}
\end{table}

\begin{figure}[!htbp]
  \centering
  \includegraphics[width=\textwidth]{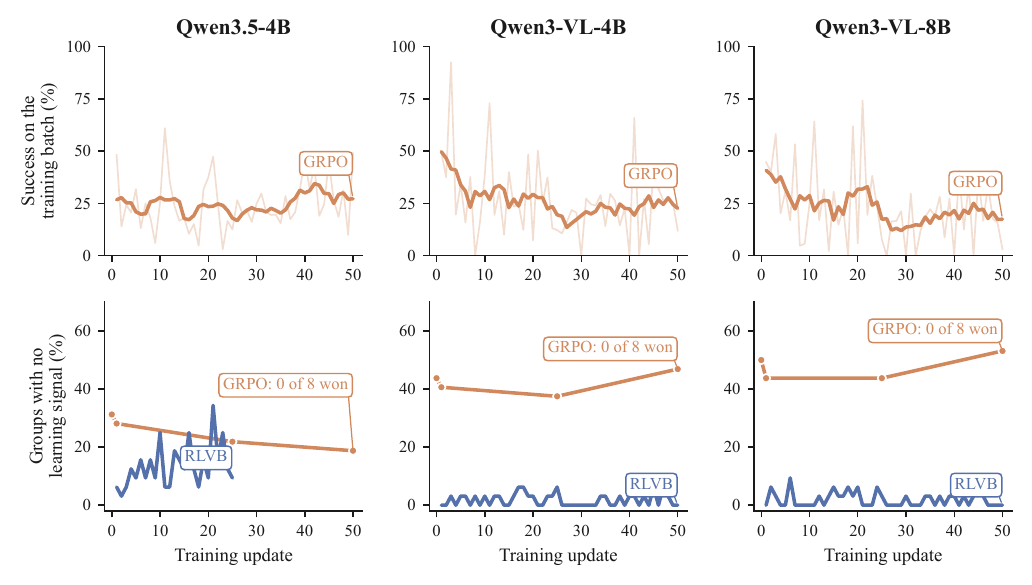}
  \caption{\textbf{Learning signal during training.} Top: success of GRPO on the training batch. Bottom: share of groups that give no learning signal because all their members receive the same reward. A group is a set of episodes for GRPO and the turns at one history for \method.}
  \label{fig:signal}
\end{figure}

\paragraph{Learning signal.}
\Cref{fig:signal} shows why the outcome reward helps little. With GRPO, a group without a winning episode gives every episode the same reward and therefore no learning signal. After 50 updates, 18.8--53.1\% of the groups are of this kind. The success of GRPO on the training batch does not rise. With the posterior reward, few groups of \method give no signal. Their share is about 2\% for the two Qwen3-VL models and 14\% for Qwen3.5-4B.

\paragraph{Fewer premature answers.}
Because the reward of \method comes from the exact posterior, it penalizes a guess even when the guess is right and rewards further steps while the evidence is not enough. We pool HiddenGames, Pro-Worlds and RoboQuest. Each range below spans the three models. Before the evidence is enough, Base answers at 11 to 20\% of the turns and \method at only 3\%. PPO (belief + action) also writes a belief, but its reward checks the belief against the true answer instead of the evidence. This reward can therefore still pay a lucky guess. PPO (belief + action) gives 56 to 83\% of its answers too early. In \cref{tab:failures}, the share of premature answers falls from 76--87\% for Base to 44--46\% for \method. \Cref{sec:more_looks} examines where the gain comes from.

\begin{table}[!htbp]
  \centering
  \footnotesize
  \setlength{\tabcolsep}{2.2pt}
  \begin{tabular}{@{}l rr rr r@{}}
    \toprule
    & \multicolumn{2}{c}{Right} & \multicolumn{2}{c}{Wrong} & \\
    \cmidrule(lr){2-3} \cmidrule(lr){4-5}
    Method & enough & too early & too early & enough & No answer \\
    \midrule
    \multicolumn{6}{@{}l}{\textbf{Qwen3.5-4B}} \\
    \quad Base & 17.8 & 9.9 & 59.1 & 3.5 & 9.6 \\
    \quad GRPO (action-only) & 21.6 & 9.9 & 62.1 & 5.6 & 0.7 \\
    \quad Base (w/ belief) & 26.8 & 11.0 & 54.4 & 6.6 & 1.3 \\
    \quad PPO (belief + action) & 34.8 & 8.8 & 45.4 & 7.8 & 3.2 \\
    \quad \method (EM) & 33.4 & 8.5 & 42.4 & 8.1 & 7.7 \\
    \quad \method & \textbf{40.2} & 8.3 & \textbf{31.0} & 9.8 & 10.6 \\
    \midrule
    \multicolumn{6}{@{}l}{\textbf{Qwen3-VL-4B}} \\
    \quad Base & 11.0 & 11.5 & 74.8 & 1.7 & 0.9 \\
    \quad GRPO (action-only) & 9.3 & 11.5 & 77.1 & 1.5 & 0.6 \\
    \quad Base (w/ belief) & 10.1 & 13.4 & 72.2 & 2.4 & 1.9 \\
    \quad PPO (belief + action) & 15.0 & 12.7 & 68.4 & 1.8 & 2.1 \\
    \quad \method (EM) & 19.0 & 10.9 & 60.8 & 7.1 & 2.3 \\
    \quad \method & \textbf{36.1} & 8.0 & \textbf{32.4} & 14.1 & 9.4 \\
    \midrule
    \multicolumn{6}{@{}l}{\textbf{Qwen3-VL-8B}} \\
    \quad Base & 11.8 & 11.7 & 72.8 & 2.6 & 1.2 \\
    \quad GRPO (action-only) & 11.2 & 11.6 & 73.1 & 2.9 & 1.3 \\
    \quad Base (w/ belief) & 15.4 & 11.4 & 67.4 & 4.8 & 1.0 \\
    \quad PPO (belief + action) & 23.3 & 10.2 & 58.2 & 6.6 & 1.7 \\
    \quad \method (EM) & 20.2 & 9.8 & 62.6 & 6.2 & 1.2 \\
    \quad \method & \textbf{34.2} & 8.1 & \textbf{33.9} & 14.6 & 9.2 \\
    \bottomrule
  \end{tabular}
  \caption{How episodes end. The numbers are percent of all episodes of HiddenGames, Pro-Worlds and RoboQuest together. \emph{Enough}: the answer was given once the evidence was enough, that is, once the exact posterior of the top answer reached 0.95 (in RoboQuest, once the feedback met every requirement of the task). \emph{Too early}: before that. Best per model in bold for right answers with enough evidence
  and wrong answers given too early.}
  \label{tab:failures}
\end{table}

\begin{figure}[!htbp]
  \centering
  \includegraphics[width=\columnwidth]{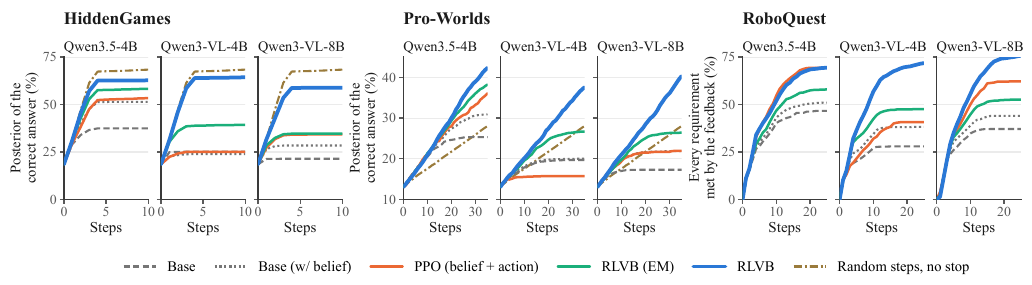}
  \caption{\textbf{Evidence gathered by the \textit{k}-th step.} Each small panel is one model. HiddenGames and Pro-Worlds: the exact posterior
  of the correct answer. RoboQuest: the share of episodes where the feedback meets every requirement of the task. A finished episode keeps its final value. The right end of each curve is therefore the evidence at the end of the episodes. The dash-dot line is an agent that takes random legal steps and never stops.}
  \label{fig:evidence_looks}
\end{figure}

\paragraph{How much evidence the agents gather.} \Cref{fig:evidence_looks} shows how the evidence grows with the number of steps. On HiddenGames and Pro-Worlds, it plots the exact posterior of the correct answer after each step. RoboQuest has no posterior. On RoboQuest, the figure plots the share of episodes where the feedback meets every task requirement. Each curve rises while that method's agents still take steps. It flattens once they stop. As a result, the curves separate where the methods stop. On Pro-Worlds, the curves overlap over the first steps. This happens because every method takes steps for a while, and each step gives about the same evidence for every method. On HiddenGames, \method's curve closely follows the dash-dot line of random steps. On Pro-Worlds, the curves for all methods lie above it by about the same amount. \method takes the most steps for all three models on all three benchmarks. Its evidence keeps growing after the other methods stop. At the end of the episodes, \method has more evidence than Base on all three benchmarks. On HiddenGames, the posterior of the correct answer is 59 to 64\% for \method, compared to 22 to 37\% for Base. On Pro-Worlds, it is 38 to 42\% for \method, compared to 17 to 25\% for Base. On RoboQuest, the feedback meets every task requirement in 68 to 77\% of \method's episodes, compared to 31 to 45\% for Base (\cref{tab:three_questions}). \Cref{sec:more_looks} shows that \method takes these extra steps because it does not commit while its belief line still names several answers.

\paragraph{The belief line and the belief reward.}

For Base (w/ belief), the belief line is already exact on many turns. It is exact on 46.8 to 48.4\% of turns on HiddenGames and on 75.7 to 82.4\% of turns on Pro-Worlds. Yet the model often does not act on this belief. It commits on 3 to 68\% of the turns where the evidence is insufficient and its own belief line still names several answers (\cref{sec:more_looks}). We test whether rewarding the belief line fixes this problem. For this test, \cref{tab:belief_term} compares training with and without the belief term in \cref{eq:score}. We replay every test episode using the exact posterior. At each turn, we count how many answers the belief line names and how many answers the evidence still allows. On HiddenGames, we look at Qwen3-VL-4B trained with the belief term and select the turns where its belief line names a single answer. The agent answers on 98\% of these turns when the evidence allows only that answer, and on 95\% of them when the evidence still allows several. With the belief term, the agent commits on almost every single-answer line, whether or not the evidence supports it. This matters because the evidence still allows several answers on about half of the turns where the belief line names one answer. \method is trained without the belief term, and it treats these two kinds of turns differently. It answers on 66\% of the turns where the evidence allows only one answer, and on 32\% of the turns where the evidence still allows several. The other two models show the same change. On HiddenGames, the belief term makes the belief line exact more often for the two Qwen3-VL models (57 against 41\% and 70 against 46\% of turns). It does not do so for Qwen3.5-4B (53 against 60\%). With the belief term, the agent also answers earlier. On HiddenGames, 43.0 to 52.9\% of its answers are premature, compared to 34.5 to 43.1\% for \method, as \cref{tab:process} shows. The belief term does not raise accuracy on any model. Removing the belief term helps most on RoboQuest. Without the belief term, 20.0 to 28.9\% of the agent's RoboQuest answers come before it has seen what the task requires, compared to 25.3 to 41.9\% with the belief term.

With the belief term, the agent commits whenever its belief line names a single answer. With the action term alone, the agent rarely commits while the evidence is insufficient and its belief line names several answers. Removing the belief term does not remove the belief from the reward. The action term is still computed from the exact belief verified by the simulator. Therefore, \method's main effect is on the decision to commit. As \cref{tab:process} shows, \method changes the share of exact belief lines over all turns by $-5.8$ to $+12.5$\% on HiddenGames and by $-7.8$ to $-4.9$\% on Pro-Worlds. The rise on Qwen3.5-4B comes from fewer belief lines that cannot be read, at 1.3\% of turns compared to 29.5\%.

\subsection{Why \method works}
\label{sec:more_looks}

\begin{tcolorbox}[colback=gray!6, colframe=gray!55, boxrule=0.5pt, arc=2pt, left=6pt, right=6pt, top=5pt, bottom=5pt]
\textbf{Summary.} \method gains because it answers with more evidence. It gets this evidence by acting on its own belief line. When the evidence is insufficient and its belief line still names several answers, \method commits at 0.1 to 2.2\% of turns, against 3 to 68\% for Base (w/ belief). As a result, its answers come mostly after its belief line names a single answer (RQ3). The belief line itself is not consistently more accurate. Over all turns, compared to Base (w/ belief), it is exact more often in some settings and less often in others, and the single answers it names are often wrong (RQ1). \method's steps add about as much evidence as the base model's steps (RQ2).
\end{tcolorbox}

There are three possible explanations for \method's gain. The trained agent could read its evidence better (RQ1), take steps that bring more evidence (RQ2), or act on what its belief line says (RQ3). In terms of metacognition, RQ1 and RQ2 look at monitoring the evidence. RQ3 looks at whether controlling the decision follows this monitoring. At every turn, the exact posterior measures how strongly the evidence gathered so far settles the answer. This lets us test each explanation.

We call each evidence-gathering action a \emph{step}. We call the evidence \emph{sufficient} when the exact posterior of the top answer is at least 0.95. RoboQuest has no posterior, so there the evidence is sufficient once all evidence requirements of the task are met. We define the \emph{evidence at the answer} as the top answer's posterior on the turn when the agent commits. On RoboQuest, it is the share of requirements met on that turn. A belief line is \emph{exact} when it names exactly the answers that are still possible. Base is the base model without a belief line. Base (w/ belief) is the same model with a belief line. In this section, HiddenGames means its five tasks with an answer, and RoboQuest means its eight tasks with countable requirements. The accuracies therefore differ from those in \cref{tab:main}. \Cref{app:working} gives details of these measures and further tests.

\begin{figure}[!htbp]
  \centering
  \includegraphics[width=\linewidth]{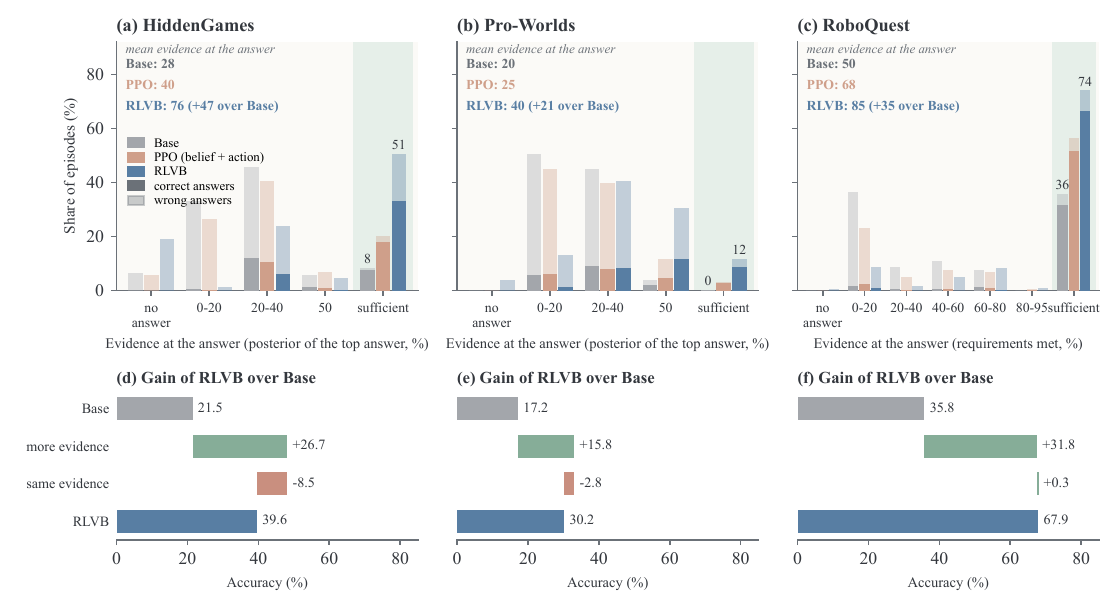}
  \caption{\textbf{The evidence at the answer.} We group each method's episodes by the evidence at the answer. This evidence is the exact posterior of the top answer, or the share of requirements met on RoboQuest. (a) to (c) show the share of episodes answered at each level. The solid part of each bar marks correct answers. The shaded level shows sufficient evidence. The numbers above it report the shares for Base and \method. The upper left corner reports the mean evidence at the answer. (d) to (f) show \method's gain over Base, split by \cref{eq:split}. We pool the results across the three models.}
  \label{fig:answer_dist}
\end{figure}

\paragraph{The evidence at the answer.} \Cref{fig:answer_dist} groups each method's episodes by the evidence at the answer. HiddenGames and Pro-Worlds use four levels because the posterior takes only a few values there. RoboQuest uses six levels. Across all methods, answers given with sufficient evidence are correct in 66 to 92\% of cases, against 0 to 13\% at the lowest level. The base model gives most of its answers at low evidence levels. It answers with a posterior below 40\% in 79\% of its HiddenGames episodes and 96\% of its Pro-Worlds episodes. In contrast, \method commits at much higher evidence levels. It answers with sufficient evidence in 51\% of HiddenGames episodes, 12\% on Pro-Worlds and 74\% on RoboQuest, against 8\%, 0.4\% and 36\% for Base and 20\%, 3\% and 57\% for PPO (belief + action). As a result, the mean evidence at the answer is higher for \method than for Base by 47.4\% in HiddenGames (95\% interval from 45.4 to 49.4\%), 20.6\% on Pro-Worlds (from 19.4 to 21.8\%), and 34.8\% on RoboQuest (from 32.5 to 37.0\%). Across individual models, this difference is between 18 and 51\%, as \cref{fig:answer_dist_models} shows.

\paragraph{Where the gain comes from.} A method can improve its accuracy in two ways. First, it can answer at higher levels of evidence by taking more steps before it commits. Second, it can answer more accurately at the same level of evidence, for example by reading its observations better. To measure both effects, we write a method's accuracy as a sum over evidence levels $e$. Each term multiplies the share of episodes answered at that level, $f(e)$, by the accuracy of the answers at that level, $a(e)$. We count episodes with no answer as wrong. The gain of a method $M$ over Base $B$ then splits exactly into two parts,
\begin{equation}
\mathrm{Acc}_M-\mathrm{Acc}_B=\underbrace{\sum_e \bigl(f_M(e)-f_B(e)\bigr)\,\bar a(e)}_{\text{more evidence}}+\underbrace{\sum_e \bigl(a_M(e)-a_B(e)\bigr)\,\bar f(e)}_{\text{same evidence}},
\label{eq:split}
\end{equation}
where $\bar a(e)$ and $\bar f(e)$ are the two methods' means at level $e$. \Cref{fig:answer_dist}(d) to \cref{fig:answer_dist}(f) show these two parts for \method. \method is more accurate because a larger share of its answers come at high levels of evidence, where every method's answers are often correct. At the same level of evidence, \method's answers are no more accurate than Base's. In HiddenGames, the 18.1\% gain splits into $+26.7$\% from more evidence and $-8.5$\% at the same evidence. On Pro-Worlds, the two parts are $+15.8$ and $-2.8$\%. On RoboQuest, they are $+31.8$ and $+0.3$\%. In HiddenGames, the second part is negative because \method misreads more of its sufficient evidence, as RQ1 shows. \Cref{tab:split} reports both parts for each model and for PPO (belief + action). Answers with sufficient evidence are also correct much more often than other answers in the same test game. \method gives a correct answer with sufficient evidence in 33.3\% of HiddenGames episodes, 8.7\% of Pro-Worlds episodes, and 66.6\% of RoboQuest episodes. Base does so in 7.5\%, 0.3\%, and 31.7\% of episodes, as \cref{tab:within_game} shows.

\paragraph{RQ1: Does \method read its evidence better?} We measure what the agent reads from its evidence through its belief line. Because \method's update raises the likelihood of the entire turn, this belief line is trained alongside the actions even though the reward scores only the action. Over all turns, \method's belief line is not consistently more accurate than the belief line of Base (w/ belief). \Cref{tab:exact_turns} reports the share of turns where the belief line is exact. We count every turn, and we count a line that cannot be read as not exact. In HiddenGames, \method's belief line is exact more often on Qwen3.5-4B, at 60.3\% against 47.8\%, but it is exact less often on the two Qwen3-VL models. On Pro-Worlds, it is exact less often on all three models, by 4.9 to 7.8\%. On RoboQuest, it is exact more often on the two Qwen3-VL models, by 5.9 and 12.2\%, and about as often on Qwen3.5-4B. Within HiddenGames, it is exact more often on the Tabletop tasks for every model, by 1.6 to 11.9\%, as \cref{tab:reading_all} shows.
\begin{table}[!htbp]
  \centering
  \footnotesize
  \setlength{\tabcolsep}{4pt}
  \begin{tabular}{@{}l rr c rr c rr@{}}
    \toprule
     & \multicolumn{2}{c}{HiddenGames} & & \multicolumn{2}{c}{Pro-Worlds} & & \multicolumn{2}{c}{RoboQuest} \\
    \cmidrule(lr){2-3}\cmidrule(lr){5-6}\cmidrule(l){8-9}
    Model & Base & \method && Base & \method && Base & \method \\
    \midrule
    Qwen3.5-4B & 47.8 & 60.3 && 82.4 & 75.0 && 83.8 & 84.2 \\
    Qwen3-VL-4B & 46.8 & 41.0 && 77.1 & 69.3 && 72.0 & 77.9 \\
    Qwen3-VL-8B & 48.4 & 45.5 && 75.7 & 70.8 && 73.7 & 85.9 \\
    \bottomrule
  \end{tabular}
  \caption{\textbf{Belief accuracy over all turns.} We report the share of turns (\%) where the belief line is exact. A belief line is exact when it names exactly the answers still possible under the exact posterior. We count a line that cannot be read as not exact. RoboQuest has no posterior. On RoboQuest, a line counts as exact when it names a single answer exactly when every requirement is met and does not exclude the true answer. Language-model readers judge this on every turn of the 361 instances that at least one method solves, as \cref{app:working} describes. Base means Base (w/ belief). \Cref{tab:reading_all} reports the HiddenGames task groups.}
  \label{tab:exact_turns}
\end{table}
\begin{figure}[!htbp]
  \centering
  \includegraphics[width=\linewidth]{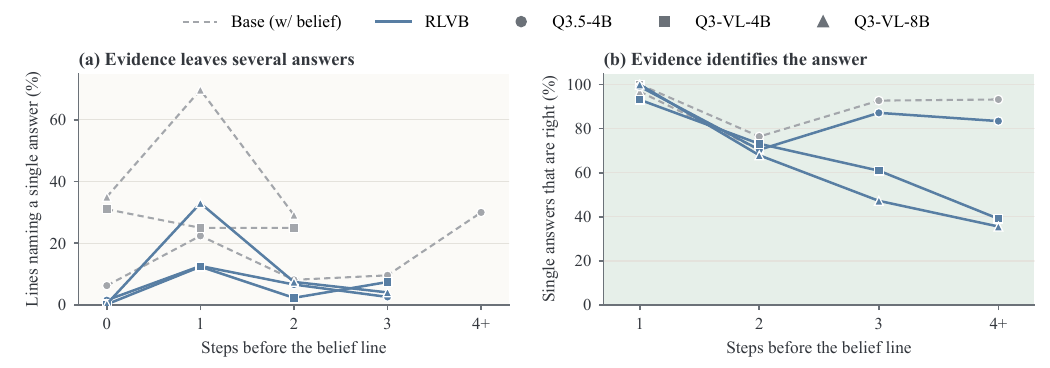}
  \caption{\textbf{The belief line after the same number of steps} on the Tabletop tasks. (a) On turns where the evidence still leaves several answers, the share of belief lines that name a single answer. (b) On turns with sufficient evidence, the share of single-answer belief lines that name the identified answer. Each curve corresponds to one model, and each plotted point requires at least 20 belief lines. Base denotes Base (w/ belief). \Cref{tab:reading_steps_all} reports results for all task groups.}
  \label{fig:reading_steps}
\end{figure}

At the same states, \method's belief line is more cautious. \Cref{tab:reading} compares the agents on states that all of them reach while several answers remain. On these states, \method's belief line names a single answer less often than Base's belief line in all nine settings. The 95\% interval of the difference excludes zero in eight of those settings. \method's belief line is also exact more often in all nine settings. \Cref{fig:reading_steps}(a) shows the same pattern after every number of steps on the Tabletop tasks. Over all turns, this caution does not make the belief line more exact for two reasons. First, \method spends many turns at later states that Base rarely reaches. Second, its belief line more often still names several answers after the evidence has become sufficient, as \cref{tab:reading_all} shows.
\begin{table}[!htbp]
  \centering
  \footnotesize
  \setlength{\tabcolsep}{3.5pt}
  \begin{tabular}{@{}ll rrr c rrr@{}}
    \toprule
     & & \multicolumn{3}{c}{Exact line (\%)} & & \multicolumn{3}{c}{Single answer (\%)} \\
    \cmidrule(lr){3-5}\cmidrule(l){7-9}
    Tasks & Model & Base & \method & + belief && Base & \method & + belief \\
    \midrule
    Tabletop & Qwen3.5-4B & 78.4 & 83.6 & 84.4 && 12.8 & 6.4 & 4.7 \\
     & Qwen3-VL-4B & 45.8 & 85.3 & 86.0 && 28.5 & 4.7 & 8.4 \\
     & Qwen3-VL-8B & 48.9 & 78.4 & 73.6 && 44.1 & 16.0 & 13.5 \\
    \midrule
    MiniGrid, MineSweeper & Qwen3.5-4B & 30.0 & 34.0 & 53.9 && 48.1 & 36.9 & 25.3 \\
     & Qwen3-VL-4B & 36.2 & 44.5 & 44.7 && 37.1 & 29.4 & 26.3 \\
     & Qwen3-VL-8B & 49.7 & 60.1 & 84.4 && 42.3 & 32.2 & 3.1 \\
    \midrule
    Pro-Worlds & Qwen3.5-4B & 73.8 & 75.4 & 73.7 && 8.4 & 3.4 & 4.4 \\
     & Qwen3-VL-4B & 69.1 & 74.6 & 73.7 && 3.2 & 2.8 & 7.2 \\
     & Qwen3-VL-8B & 66.4 & 77.9 & 79.5 && 5.5 & 0.9 & 0.5 \\
    \bottomrule
  \end{tabular}
  \caption{\textbf{The belief line while several answers remain}, evaluated on states reached by all three agents. A state is defined by the game together with the number of answers still possible, with every state weighted equally. \emph{Exact line}: share of belief lines that name exactly the answers still possible. \emph{Single answer}: share of belief lines that name a single answer. Unreadable lines are excluded. Base denotes Base (w/ belief), and \emph{+ belief} denotes \method trained additionally with the belief term of \cref{eq:score}, as in \cref{tab:belief_term}. \Cref{tab:belief_diag_full} reports the 95\% intervals.}
  \label{tab:reading}
\end{table}

Once the belief line narrows to a single answer, this answer is often wrong. At the last turn of an episode that ends with an answer, \method's belief line names a single answer in 83\% of the Tabletop episodes. That answer is correct in 57\% of those cases, or in 70\% when the evidence is sufficient, as \cref{tab:last_line} shows. After the same number of steps, the two agents' single answers are correct about equally often. With sufficient evidence, these answers are correct in 93--100\% of cases after one step for both agents. After two steps, they are correct in 68--73\% of cases for \method and 70--76\% for Base, as \cref{fig:reading_steps}(b) shows. \method takes more steps. On the two Qwen3-VL models, its single answers are correct less often as steps add up. After four or more steps, they are correct in 36--39\% of cases. Base almost never reaches that many steps. When \method gives a wrong answer with sufficient evidence on the Tabletop tasks, its belief line excludes the true answer in 85--97\% of cases, as \cref{tab:belief_errors} shows.

\begin{tcolorbox}[colback=blue!4, colframe=blue!35!black!45, boxrule=0.5pt, arc=2pt, left=6pt, right=6pt, top=4pt, bottom=4pt]
\textbf{Answer to RQ1.} Not consistently. At the same states, \method's belief line names a single answer too early less often. Over all turns, it is exact more often in some settings and less often in others. The single answers it names are no more often right.
\end{tcolorbox}

\begin{figure}[!htbp]
  \centering
  \includegraphics[width=\linewidth]{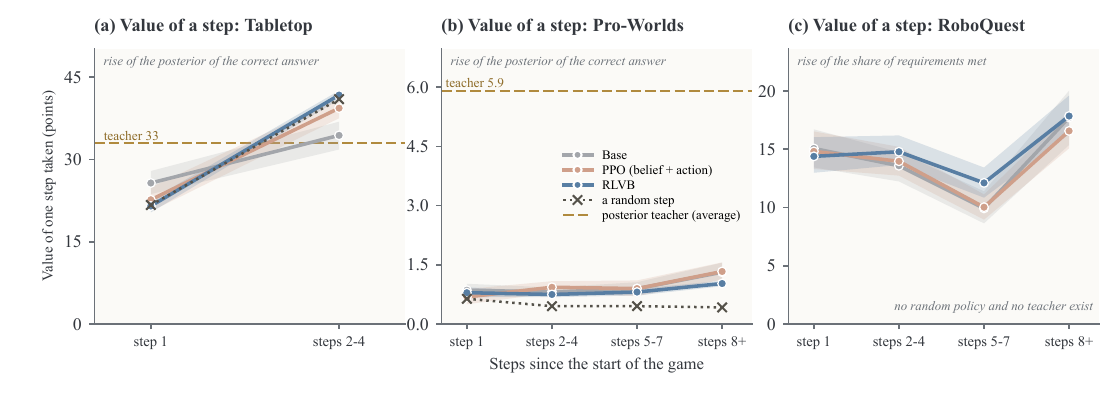}
  \caption{\textbf{The value of a step} when every episode is followed from the start of its test game, with the steps grouped into four windows. The value is the increase in the exact posterior of the correct answer in percent, or on RoboQuest the increase in the share of requirements met, averaged over the steps taken in a window before the evidence is sufficient. The dotted line indicates a random step and the dashed line indicates the average step of the posterior teacher. RoboQuest has neither line. The bands represent 95\% intervals.}
  \label{fig:step_value}
\end{figure}

\paragraph{RQ2: Does \method judge what further steps can bring?}
If \method had learned to anticipate upcoming evidence, it would choose steps that add more evidence than those of Base and recognize when its own steps stop adding evidence. We measure the value of a step by how much it raises the exact posterior of the correct answer, or by the share of requirements met on RoboQuest. By this measure, steps from \method add about as much evidence as those from Base. From the states reached by Base, PPO (belief + action), and \method, the first step raises the posterior by 15.6, 14.9, and 14.8\% in HiddenGames and by 1.5, 1.9, and 1.5\% on Pro-Worlds, while meeting 19.1, 19.3, and 19.8\% of the requirements on RoboQuest, as \cref{app:working} shows. Later steps have similar value across every window of steps from the start of the game, as \cref{fig:step_value} shows. Across states reached by both Base and \method, about 20\% of each agent's actions meet one more requirement on RoboQuest, and 4--6\% of their moves add evidence on Pro-Worlds, as \cref{tab:same_state} shows. Only in MiniGrid and MineSweeper does the first step of \method add less evidence, raising the posterior by 0.14\%, against 1.24\% for Base.

\method also does not choose better actions. Measured against the actions ranked best by the posterior teacher, \method and Base choose a best action about equally often, in 40 versus 37\% of HiddenGames states and 53 versus 52\% of Pro-Worlds states. A better belief line does not lead to better steps either. From the same state and with an exact belief line, a step from \method raises the posterior of the correct answer by 0.9\% more than a step from Base (w/ belief) on the Tabletop tasks, where a step adds about 30\%, and by 0.02\% less on Pro-Worlds, where a step adds about 0.9\%. The 95\% intervals for both differences include zero. The same holds when the posterior teacher continues playing to the end after each step. With an exact belief line, a step from \method leaves the teacher with as many remaining steps as one from Base (w/ belief) on Pro-Worlds and in the Tabletop tasks, and 0.53 more in MiniGrid and MineSweeper, as \cref{app:working} shows.

\method also does not recognize when its own steps stop adding evidence. In MineSweeper, the posterior teacher reaches sufficient evidence in 83\% of scenes, as \cref{tab:explore_limit} shows. \method's steps rarely reach sufficient evidence, and the agent answers in only 3 to 16\% of episodes. On Qwen3-VL-4B, 44 of its 81 MineSweeper episodes without an answer add no evidence across their last ten turns, as \cref{tab:stall} shows.
\paragraph{Why the steps do not improve.} Three things explain this result. First, \method gains the most on the Tabletop tasks. On those tasks, every step that tests a new object splits the possible answers in the same way, as \cref{app:lookahead} shows. The choice of step does not matter on these tasks. Second, on MiniGrid and MineSweeper, the reward hardly separates a step without evidence from the best step. That kind of step only delays the teacher's best plan by one turn. As a result, its score is lower than the best step's score by about one time discount. This discount is 0.2 nats, or 0.5 nats in MineSweeper. An early answer scores several nats lower, for example $-7.42$ in \cref{fig:turn}. Because of this, the reward mostly teaches when to answer. Third, Pro-Worlds and RoboQuest never appear in training. Better steps on those tasks would have to carry over from the training tasks. A test on Qwen3.5-4B supports these points. With the three rules in \cref{app:rules}, the teacher rejects steps without evidence. On the HiddenGames tasks with an answer, the share of steps that add evidence then rises from 21.2 to 28.0\%. On Pro-Worlds, this share changes from 5.1 to 5.0\%. On RoboQuest, it changes from 21.3 to 21.9\%. Accuracy changes by $-0.2$ to $+2.9$\%.

\begin{tcolorbox}[colback=blue!4, colframe=blue!35!black!45, boxrule=0.5pt, arc=2pt, left=6pt, right=6pt, top=4pt, bottom=4pt]
\textbf{Answer to RQ2.} No. \method's steps add about as much evidence as the base model's steps. \method also does not notice when its own steps stop adding evidence. The reward mostly scores when to answer. It hardly separates a step without evidence from the best step.
\end{tcolorbox}

\begin{table}[!htbp]
  \centering
  \footnotesize
  \setlength{\tabcolsep}{3.5pt}
  \resizebox{\textwidth}{!}{%
  \begin{tabular}{@{}l rr rr c rr rr c rr rr@{}}
    \toprule
     & \multicolumn{4}{c}{RoboQuest} & & \multicolumn{4}{c}{Tabletop} & & \multicolumn{4}{c}{Pro-Worlds} \\
    \cmidrule(lr){2-5}\cmidrule(lr){7-10}\cmidrule(l){12-15}
     & \multicolumn{2}{c}{Useful (\%)} & \multicolumn{2}{c}{Commits (\%)} & & \multicolumn{2}{c}{Evidence (bits)} & \multicolumn{2}{c}{Commits (\%)} & & \multicolumn{2}{c}{Useful (\%)} & \multicolumn{2}{c}{Commits (\%)} \\
    \cmidrule(lr){2-3}\cmidrule(lr){4-5}\cmidrule(lr){7-8}\cmidrule(lr){9-10}\cmidrule(lr){12-13}\cmidrule(l){14-15}
    Model & Base & \method & Base & \method && Base & \method & Base & \method && Base & \method & Base & \method \\
    \midrule
    Qwen3.5-4B & 21 & 21 & 30 & 13 && 1.30 & 1.29 & 45 & 5 && 6 & 5 & 9 & 2 \\
    Qwen3-VL-4B & 20 & 21 & 49 & 7 && 0.92 & 0.85 & 75 & 5 && 4 & 4 & 16 & 2 \\
    Qwen3-VL-8B & 21 & 21 & 38 & 7 && 0.81 & 0.87 & 87 & 12 && 5 & 5 & 33 & 2 \\
    \bottomrule
  \end{tabular}%
  }
  \caption{\textbf{Base and \method at the same state.} A state is defined by the task together with the evidence gathered so far, and we compare the two models across states reached by both. \emph{Useful}: share of actions that add evidence. \emph{Evidence}: evidence added by one action, in bits. \emph{Commits}: share of cases in which the agent commits to an answer although its evidence is not sufficient.}
  \label{tab:same_state}
\end{table}

\begin{figure}[!htbp]
  \centering
  \includegraphics[width=\linewidth]{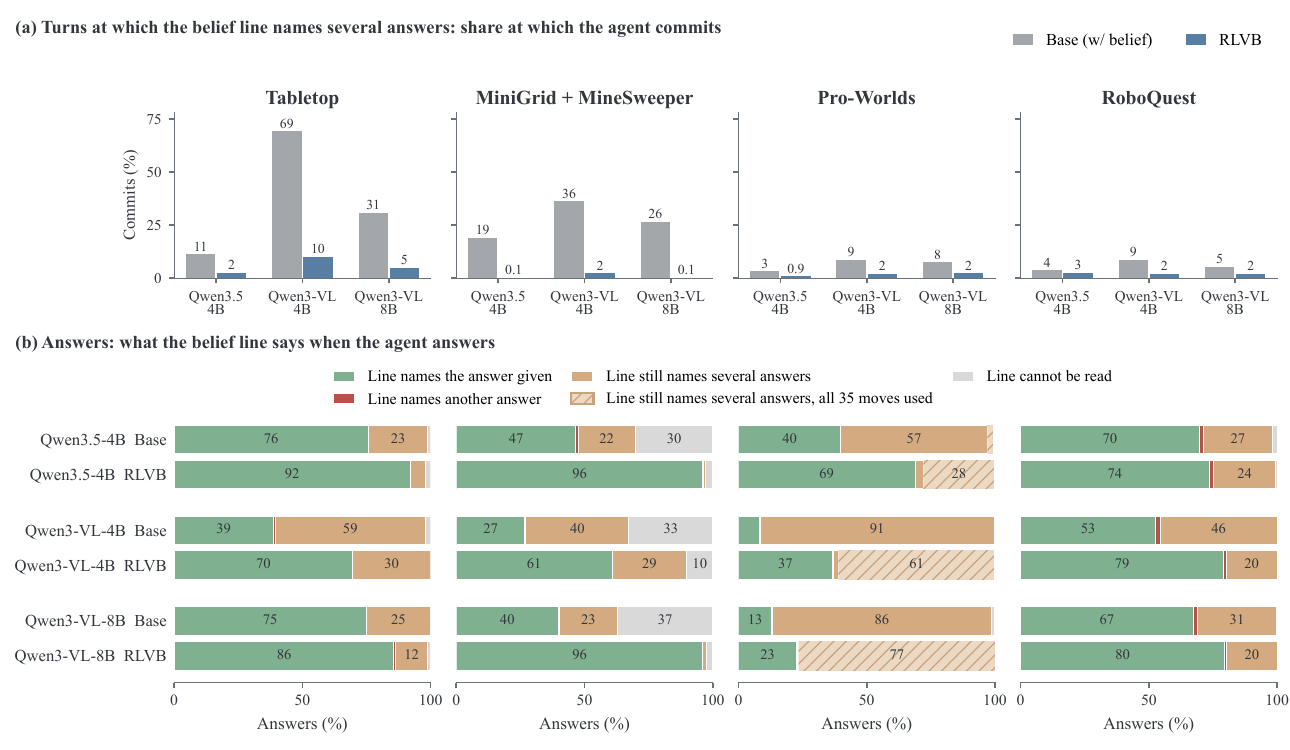}
  \caption{\textbf{Acting on the belief line.} (a) Among turns where the belief line names several answers, at any level of evidence, the share where the agent commits. (b) For every answer, the belief line written on that turn either names the given answer, names another answer, or still names several answers. On Pro-Worlds, answers given after all 35 moves are used are hatched because the agent can only answer then. Base means Base (w/ belief). Results are pooled across the three test seeds. On RoboQuest, language-model readers label every belief line, as \cref{app:working} describes.}
  \label{fig:line_commit}
\end{figure}

\paragraph{RQ3: Does \method act on its belief line?} The base model often acts against its own belief line. Over all turns where the belief line names several answers, Base (w/ belief) commits at 3.2 to 69.1\% of these turns, while \method commits at 0.1 to 10.0\%. On RoboQuest, Base (w/ belief) commits at 3.8 to 8.7\% of these turns, while \method commits at 1.9 to 2.6\%, as \cref{fig:line_commit}(a) shows. As a result, many of the base model's answers go against its own belief line. When the base model answers, its belief line still names several answers in 23 to 59\% of its Tabletop answers, 22 to 40\% in MiniGrid and MineSweeper, 57 to 91\% on Pro-Worlds, and 27 to 46\% on RoboQuest. For \method, these shares are 6 to 30\%, 1 to 29\%, 0.3 to 3.3\%, and 20 to 24\%, as \cref{fig:line_commit}(b) shows. On Pro-Worlds, \method gives most of its other answers with such a line after all 35 moves are used. When the evidence is also insufficient, \method commits at only 0.1 to 2.2\% of these turns, against 3.2 to 67.9\% for Base (w/ belief). In all nine settings, the 95\% intervals do not overlap, as \cref{tab:commit_full} shows. From the same states with insufficient evidence, Base commits in 30--49\% of cases on RoboQuest, against 7--13\% for \method. On the Tabletop tasks, Base commits in 45--87\% of cases, against 5--12\% for \method, as \cref{tab:same_state} shows. Instead of committing, \method keeps working. It takes 2.6 to 3.1 steps on the Tabletop tasks, which is close to the posterior teacher's 2.5 steps. Base takes 0.1 to 0.8 steps.

Like Base, \method gives the answer that its belief line names when it commits after a single-answer line. Both agents give the named answer in at least 97\% of these commits in all nine settings, and in at least 96\% on RoboQuest, as \cref{fig:line_commit} shows. \method rarely commits when the evidence is insufficient and its belief line names several answers. Because of this, it answers with a single-answer belief line far more often than Base. This happens in 83 against 63\% of answered Tabletop episodes, 40 against 20\% on Pro-Worlds, and 79 against 66\% on RoboQuest, as \cref{tab:last_line} shows. To separate acting on the belief line from forming it, we combine the share of turns where one agent's belief line names a single answer, taken from \cref{tab:reading_all}, with the other agent's commit rates for each kind of line, taken from \cref{tab:commit_full}. We average the result over both orders. Acting on the belief line accounts for 53 to 83\% of the drop in early commits on the Tabletop tasks, nearly all of it on Pro-Worlds, and about four fifths on RoboQuest. The rest comes from the belief line naming a single answer too early less often.
\begin{table}[!htbp]
  \centering
  \footnotesize
  \setlength{\tabcolsep}{4pt}
  \begin{tabular}{@{}l rr c rr@{}}
    \toprule
     & \multicolumn{2}{c}{Answers with a single-answer line (\%)} & & \multicolumn{2}{c}{\method: single answer correct (\%)} \\
    \cmidrule(lr){2-3}\cmidrule(l){5-6}
    Tasks & Base & \method && All & Sufficient evidence \\
    \midrule
    Tabletop & 63.4 & 82.7 && 57.2 & 69.6 \\
    MiniGrid, MineSweeper & 43.4 & 84.4 && 32.2 & -- \\
    Pro-Worlds & 20.0 & 40.0 && 36.9 & 77.6 \\
    RoboQuest & 66.1 & 78.6 && 87.8 & 98.7 \\
    \bottomrule
  \end{tabular}
  \caption{\textbf{The belief line at the answer.} The last turn's belief line for every episode that ends with an answer. \emph{Answers with a single-answer line} is the share of these episodes where the belief line names a single answer. \emph{Single answer correct} is the share of these single answers that are the true answer. We report it over all such episodes and over those with sufficient evidence at the answer. -- means fewer than 20 episodes. On RoboQuest, the belief lines are read as in \cref{tab:exact_turns}. Base means Base (w/ belief). Results are pooled across the three models.}
  \label{tab:last_line}
\end{table}
\begin{figure}[!htbp]
  \centering
  \includegraphics[width=\linewidth]{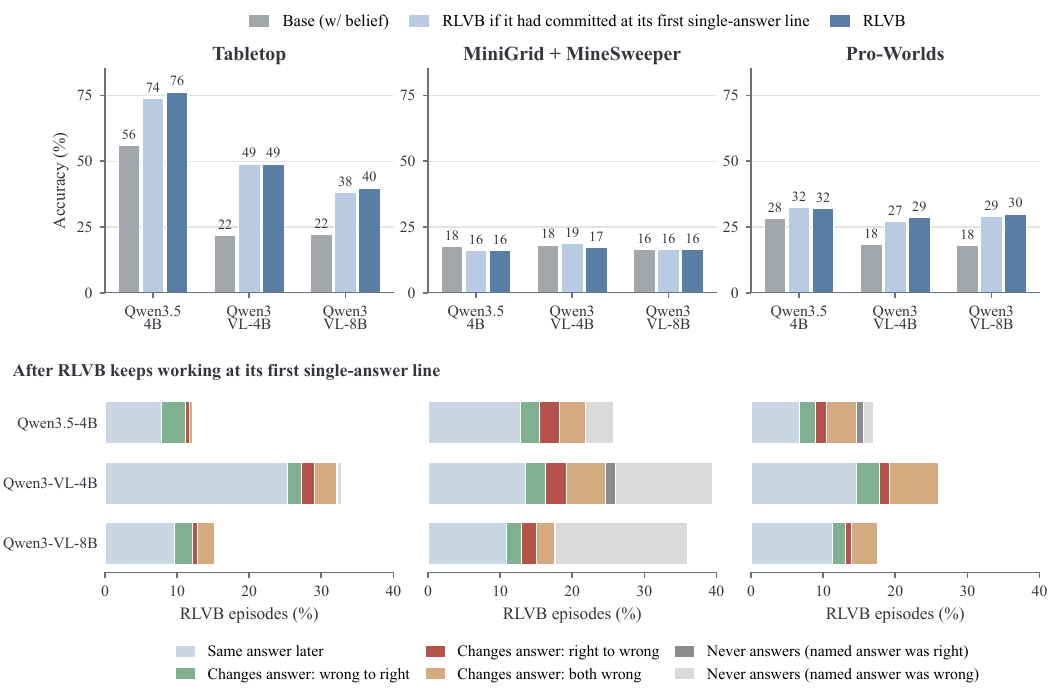}
  \caption{\textbf{Committing at the first single-answer line.} The upper row shows the accuracy of Base (w/ belief) and \method. It also shows the accuracy \method would reach by committing to the answer in its first single-answer belief line. The lower row shows what follows in episodes where \method keeps working at its first single-answer line, as a share of its episodes. Results are pooled across the three test seeds.}
  \label{fig:follow}
\end{figure}

\method acts on its belief line almost entirely through its first single-answer line. \Cref{fig:follow} compares its accuracy with the accuracy it would reach by committing to that line's answer. The two accuracies differ by at most 2.3\% in all nine settings. When \method keeps working after such a line, it most often gives the same answer later. Changes from a wrong answer to a right answer happen about as often as changes from a right answer to a wrong answer. Its extra step most often tests the named answer. On Qwen3-VL-4B, it lifts the named cup in 46 of 48 such steps on cups. On lamps, it presses the last button in 177 of 198 such steps, and the belief line has already inferred that button's lamp. After this step, its belief line names the same answer in 74\% of cases, and it commits on the next turn in 88\%. This check rarely corrects a wrong belief line because the next line names the same wrong answer in 66\% of cases. The check also depends little on the evidence. Pooled over the three models on the Tabletop tasks, \method commits at 78.5\% of single-answer turns with insufficient evidence and at 78.6\% with sufficient evidence. Because \method gives the answer that its belief line names, errors in the belief line become wrong answers. This produces the negative second part of \cref{eq:split} in HiddenGames.

A likely reason for this behavior lies in the reward. The reward never reads the belief line, and it scores every action against the exact posterior. A commit gets a low reward when the evidence leaves several answers open, even if the guess happens to be correct. In contrast, an outcome reward gives a lucky guess the full reward. At these states, the belief line usually names several answers. The update raises the likelihood of the whole turn, including the line. As a result, the agent learns to keep working when its own line names several answers. Taking one more step after the evidence is sufficient costs only that step's small time discount in the action term, as \cref{sec:score} describes. This makes the extra check cheap. Adding a reward on the belief line itself does not raise accuracy, as \cref{tab:belief_term} shows.

The on-policy variant of \method doubts its belief line more and keeps exploring. On Qwen3-VL-4B, after it takes a step while its belief line names a single answer, its next line keeps that answer in 61\% of cases, against 74\% for \method. It commits on the next turn in 71\%, against 88\%. Of these steps, 20\% are rejected as invalid, against 5\%. On the Tabletop tasks, its belief line still names several answers on 62\% of turns with sufficient evidence, against 46\% for \method. It commits at only 9\% of these turns, against 37\% for \method. It ends half of its HiddenGames episodes without an answer, against 18\% for \method. It is less accurate than \method on all three benchmarks, as \cref{sec:onpolicy} shows.

\begin{tcolorbox}[colback=blue!4, colframe=blue!35!black!45, boxrule=0.5pt, arc=2pt, left=6pt, right=6pt, top=4pt, bottom=4pt]
\textbf{Answer to RQ3.} Yes. The base model often commits while its own belief line still names several answers, and \method rarely does. This accounts for 53 to 83\% of the drop in early commits on the Tabletop tasks and nearly all of it on Pro-Worlds. It is the source of the accuracy gain.
\end{tcolorbox}

\paragraph{A check in the harness.} Such a check cannot replace an agent that acts on its own belief. Our harness does not accept an episode's first answer. It asks the agent once to check whether its evidence identifies the answer. The agent can then repeat its answer, change it, or keep working. This check raises Base's accuracy by 6.1 to 11.3\% on RoboQuest. On HiddenGames and Pro-Worlds, it changes accuracy by at most 1.9\%, as \cref{tab:harness} shows. On the Tabletop tasks, Base keeps working after the check in only 2 to 17\% of the checked episodes, as \cref{app:working} shows. Otherwise, it repeats its answer. In these tasks, Base (w/ belief) has a belief line that already names a single answer in 38 to 72\% of its early commits. Because the harness does not read the observations, it cannot tell the agent when its evidence is sufficient. \method writes its belief line from its own observations, and that line guides most of its decisions to commit.

\paragraph{Limits.}
An agent that keeps working gains only where its steps bring evidence. Among the HiddenGames tasks, \method gains most on the Tabletop tasks, where every valid step adds evidence, with 51\% on lamps and 35\% on cups, as \cref{app:pertask} shows. On RoboQuest, where about one action in five meets a requirement, the gain on the three models is 14 to 33\%, and on Pro-Worlds, where about one move in twenty adds evidence, the gain is 12 to 15\%. In MiniGrid and MineSweeper, a step raises the posterior of the correct answer by about 1\% or less, and \method shows no gain on these tasks. As RQ2 shows, \method has not learned to commit to its best answer in MineSweeper when its steps stop adding evidence or when no step remains. Across the three models, \method ends 19\% of its HiddenGames episodes without an answer, compared with 7\% for Base, almost all of them in MineSweeper.

\begin{tcolorbox}[colback=yellow!6, colframe=yellow!40!black!55, boxrule=0.5pt, arc=2pt, left=6pt, right=6pt, top=5pt, bottom=5pt]
\textbf{Reading the evidence is the next bottleneck.} \method answers with sufficient evidence far more often than the base model. Even so, only 65.7\% of these answers are correct in HiddenGames, 73.7\% in Pro-Worlds, and 89.8\% in RoboQuest. Most of these wrong answers come from the two Qwen3-VL models. In the Tabletop tasks, \method on these two models is correct on 55.6\% and 52.3\% of its answers with sufficient evidence, against 86.6\% for \method on Qwen3.5-4B. On lamps, Base never reaches sufficient evidence on these two models. \method reaches sufficient evidence in 295 episodes per model and answers correctly in 48\% and 33\% of them. Because \method acts on its belief line, we expect a better reading of the observations to translate into better answers. However, its reward scores only the action. It trains how the agent reads these observations only through the actions that follow.
\end{tcolorbox}

\section{Limitations}
\label{sec:limitations}

\method needs a task with known rules. The environment can compute the exact posterior only when it knows the prior over the hidden answer and the likelihood of each observation. Many puzzles, games, and simulated household tasks meet this condition. For tasks with unknown or open-ended rules, the posterior must be estimated. Very large action spaces need a shallower or sampled lookahead. The cost of the teacher's lookahead grows with the number of actions and observations. Finally, a belief line names a set of possible answers. This set states the posterior exactly only when every possible answer is equally likely. The belief line cannot express finer degrees of certainty. With training, wrong answers given with enough evidence also rise from 2--4\% to 10--15\% of the episodes, mostly in the Tabletop tasks (\cref{tab:failures,tab:belief_errors}). On these tasks, the agent misreads observations that already settle the answer. It then gives the wrong answer that its belief line names (\cref{sec:more_looks}). The action term also scores which step the agent takes, but the trained agent does not choose more informative steps. On Pro-Worlds, a step by the posterior teacher raises the posterior by 5.9\%, compared to 0.7 to 1.3\% for a step by any method. The trained agents also stay far below the posterior teacher (for example 32.4\% against 100\% on Pro-Worlds). The agents have not yet learned much of what the exact posterior allows.

\section{Conclusion}
\label{sec:conclusion}

We showed that a simulator can compute the exact posterior of an evidence-seeking task from its privileged information. This posterior gives a process reward at every turn. The posterior scores the agent's choice between taking more steps and answering. It can also score the agent's stated belief. \method learns from this reward with a policy gradient, and \method (EM) learns from it with EM. Across three vision-language models and three benchmarks, both \method learners improve the base model in every setting. With the same learner, the posterior process reward beats the outcome reward by 18.1\% on average. On each of the three benchmarks, \method also beats a process reward from an LLM judge on average. The trained agent answers with sufficient evidence in 51\%, 12\%, and 74\% of episodes, compared to 8\%, 0.4\%, and 36\% for the base model. It does this by acting on its own belief. With the same belief line, it commits on only 0.1 to 2.2\% of turns with insufficient evidence where this line still names several answers, compared to 3 to 68\% for the base model. Across all turns, its belief line is not consistently more exact than the base model's. Each of its steps adds about as much evidence as a step from the base model. A check in the harness does not raise the base model's accuracy on HiddenGames and Pro-Worlds because the harness cannot read the observations. We expect a better reading of the evidence to carry into the answers when an agent acts on its belief. \method's steps are not yet more informative than the base model's. An agent with sufficient evidence can still misread its observations. We want to extend \method to tasks where the posterior must be learned, to improve the choice of the next step, and to train the agent to read its observations correctly once its evidence is sufficient.

{\let\clearpage\relax\clearpage
}

\bibliographystyle{plainnat}
\bibliography{main}

\clearpage
\xdef\klsupplpage{\number\value{page}}
\makeatletter
\let\dl@suppsection\section
\renewcommand{\section}{\FloatBarrier\dl@suppsection}
\makeatother

\clearpage
\setcounter{page}{1}
\maketitlesupplementary
\appendix

\section{The posterior teacher in detail}
\label{app:teacher}

This section describes the posterior teacher of \mainref{sec:teacher} in detail.

\paragraph{Boundary values.}
In \mainref{eq:values}, $V^{(0)}(b) = \log c(t) + \log \hat\rho^{+}(b)$. Here $\hat\rho^{+}$ is an upper confidence bound on $\hat\rho$ that adds one standard deviation, as given in \cref{eq:ucb}. This bound is high for belief classes with few measured answers. The lookahead thus gives a high value to evidence that leads to these classes. The time of $b^{a,o}$ is $t + 1$. A belief at the step limit $t = T$ and a belief that identifies the answer both have the value $\log c(t) + \log \hat\rho(b)$. For an answer with $b(y) = 0$, we floor the accuracy at $10^{-9}$. Such an answer has the value $\log c(t) + \log 10^{-9}$. An outcome that ends the episode without an answer, such as a revealed mine, has the value of knowing nothing, $\log c(t) + \log(1/|\Theta|)$. A lookahead of depth 2 evaluates every pair of evidence actions with their observations. Its cost thus grows as $(|\mathcal A|\,|\mathcal O|)^2$ Bayes updates per history.

\paragraph{Task lookaheads.}
We use \mainref{eq:values} with depth 2 for the Tabletop tasks. The other tasks have many evidence actions. Each of them computes the same kind of value, $\log c(t) + \log\hat\rho$, with its own lookahead. MiniGrid, FrozenLake and Sokoban value each move by the best view reached on a shortest plan starting with this move. MineSweeper values each reveal by the posterior after it. The cost above holds for the depth-2 recursion only.

\paragraph{Why the logarithm.}
The average of $\log \hat\rho$ counts progress at every step. Each halving of the possible answers adds $\log 2$, whether it comes early or late. For a uniform posterior, this average is the expected information gain on the answer. When the answers are many, the average of $\hat\rho$ itself stays close to 0 until the answer is almost decided. For example, on a $6 \times 6$ MineSweeper board with 6 mines and 1,947,792 mine sets, the average of $\hat\rho$ cannot rank the actions. It also gives almost no cost to a risk that ends the episode.

\paragraph{Answer accuracy.}
$\hat\rho(b)$ is the measured answer accuracy of the agent:
\begin{equation}
\begin{aligned}
  \hat\rho(b) &= b_{\max}\,\hat\varphi_{\kappa(b)}, \qquad r_y = \frac{b(y)}{b_{\max}}, \\
  \hat\varphi_\kappa &= \frac{4 + \sum_{\omega \in \Omega_\kappa} \frac{1}{\bar n}\sum_{y \in \omega} r_y}{4 + \sum_{\omega \in \Omega_\kappa} \frac{n_\omega}{\bar n}}.
\end{aligned}
\label{eq:accuracy}
\end{equation}
$\kappa(b)$ is the belief class. It is the number of answers still possible, capped at 4. $\hat\varphi_\kappa$ is the mean of $r$ over the agent's answers in class $\kappa$, with a prior mean of 1 and 4 pseudo-counts. $\hat\varphi = 1$ when the agent always names the most probable answer. Each update gives one measurement for each cell $\omega = (\text{scene}, \text{belief class})$. This measurement is the mean of $r$ over all of the agent's answers in $\omega$. This mean replaces a single sample by its conditional mean given the cell. As in the Rao--Blackwell theorem~\cite{rao1945information,blackwell1947conditional}, it thus has a smaller variance than one sample. $n_\omega$ is the number of answers in $\omega$. $\bar n$ is the mean of $n_\omega$ over all cells measured in the same update as $\omega$. $\Omega_\kappa$ is the set of cells of class $\kappa$ measured so far over all updates. Each $r_y$ is taken at the belief where the agent gave the answer $y$. Episodes starting with random evidence actions keep each belief class measured.

The estimate $\hat\varphi_\kappa$ is the mean of a Beta posterior with parameters $A_\kappa = 4 + \sum_{\omega \in \Omega_\kappa} \frac{1}{\bar n}\sum_{y \in \omega} r_y$ and $A_\kappa + D_\kappa = 4 + \sum_{\omega \in \Omega_\kappa} \frac{n_\omega}{\bar n}$. The upper confidence bound adds one standard deviation of this posterior:
\begin{equation}
\begin{aligned}
  \hat\rho^{+}(b) &= \min\big\{1,\ b_{\max}\,(\hat\varphi_{\kappa} + \mathrm{sd}_\kappa)\big\}, \quad \kappa = \kappa(b), \\
  \mathrm{sd}_\kappa &= \sqrt{\frac{A_\kappa\, D_\kappa}{(A_\kappa + D_\kappa)^2\, (A_\kappa + D_\kappa + 1)}}.
\end{aligned}
\label{eq:ucb}
\end{equation}

\paragraph{Teacher policy.}
In \mainref{eq:teacher}, we normalize the step level over the groups $\mathcal G(b)$ of evidence actions with equal values $Q^{(2)}(b, \cdot)$. Each action in a group gets the full probability of its group. The step level of $\pi^*$ is thus a distribution over groups. If $b$ identifies the answer, $p_{\text{ans}} = 1$. An answer outside the support of $b$ has accuracy 0. It thus gets the floor value $\log c(t) + \log 10^{-9}$. An action that cannot be executed in the environment uses up the turn. It has the value of an answer one step later, $\log c(t+1) + \log\hat\rho(b)$. The log-probability has the floor $-200$. $\pi^*$ depends on the agent only through $\hat\rho$. Where the agent's answers are not reliable, $Q(b, y)$ is low and $\pi^*$ asks for more evidence. The stopping rule of the teacher thus fits the agent.

\begin{proposition}[What the teacher optimizes]
\label{prop:optimal}
Suppose the lookahead reaches the step limit and $\hat\rho$ is the true answer accuracy of the agent. Then, as $\tau \to 0$, the best action of $\pi^*$ maximizes $\mathbb E\big[\log\big(c(T_{\text{stop}})\,\hat\rho(b_{T_{\text{stop}}})\big)\big]$.
\end{proposition}
\begin{proof}[Proof sketch]
We have $\max_y Q(b, y) = \log c(t) + \log\hat\rho(b)$. With depth $n \ge T - t$, \mainref{eq:values} is therefore the Bellman recursion of the stopping problem with the reward $\log c(t) + \log\hat\rho(b)$ at the stop. By backward induction on the steps left, $V^{(n)}$ is its optimal value. As $\tau \to 0$, $\pi^*$ puts all its mass on the maximizers of the values.
\end{proof}
The best action of $\pi^*$ therefore maximizes the expected log of the discounted answer accuracy. The proposition holds for the recursion of \mainref{eq:values}. The task lookaheads above approximate this recursion. With depth 2, $Q^{(2)}$ approximates this optimum. This criterion differs from the expected return. It prefers steady progress to a gamble that can lose all the evidence. If $\hat\rho$ is the true accuracy, $\mathbb E[G] = \mathbb E\big[c(T_{\text{stop}})\,\hat\rho(b_{T_{\text{stop}}})\big]$. By Jensen's inequality, the teacher's criterion is then a lower bound on $\log \mathbb E[G]$. The gap between the two makes the teacher averse to risk.

\FloatBarrier
\section{The process reward in detail}
\label{app:score}

This section describes the process reward in \mainref{eq:score} in detail. The action term is the reward for \method. It scores the action against the teacher. The teacher acts on the exact belief $b$, and the simulator can verify this belief. The belief term is a variant that also scores the stated belief line against $b$. We use it in \method (EM) and in the ablation in \mainref{sec:ablation}.

\paragraph{Action term.}
The action term scores the agent's action against the teacher policy $\pi^*$. Inside one level of \mainref{eq:teacher},
\begin{equation}
\begin{aligned}
  \log \pi^*(u \mid b) &= \frac{\bar Q(b, u)}{\tau} + \nu_{\mathcal L}(b), \\
  \nu_{\mathcal L}(b) &= \log P_{\mathcal L}(b) - \log \textstyle\sum_{u' \in \mathcal L} e^{\bar Q(b, u')/\tau},
\end{aligned}
\label{eq:action-term}
\end{equation}
where $\bar Q(b, u) = Q(b, y)$ for an answer $u = y$, and $\bar Q(b, u) = Q^{(2)}(b, a)$ for an evidence action $u = a$. The level $\mathcal L$ of $u$ is ``answer'' or ``step''. We set $P_{\text{answer}} = p_{\text{ans}}$ and $P_{\text{step}} = 1 - p_{\text{ans}}$. For the step level, the sum runs over the groups $\mathcal G(b)$. Because $\nu_{\mathcal L}(b)$ depends only on $b$ and on $\mathcal L$, the action term is a soft advantage of the lookahead values in units of $\tau$. An action whose value is one nat below the best costs $1/\tau$ nats. In the belief term, a line that leaves out all of the posterior mass costs $m$ nats in $S$.

\paragraph{Belief language.}
The belief language is the set of lines \bline{Possible: $v_1, \ldots, v_n$} with distinct values $v_i \in \Theta$, plus the open line \bline{Possible: all}, which rules out nothing. We read a line in the language exactly as its set $B \subseteq \Theta$, and the open line as $B = \Theta$. Any other text is outside the language and receives the floor of every term. The open line is thus the safe statement. A set states only the support of $b_t$. It states $b_t$ itself only when $b_t$ is flat on its support, as when each observation is either possible or impossible under each target. The last belief line in the prompt is the agent's own earlier estimate. In the variant with the belief term, the teacher scores each new line against $b_t$, which comes only from the evidence. A line that copies an earlier mistake thus scores low.

\paragraph{Properties of the log score $S$.}
The stated distribution $\hat b_\beta$ is uniform on $B$ with the log-probability floor $-m$ outside $B$. Then $S(\beta; b) = \mathbb E_{\theta \sim b}[\log \hat b_\beta(\theta)] = -b(B)\log|B| - m\,(1 - b(B))$.
\begin{itemize}
  \item The log score is proper~\cite{gneiting2007strictly}. Among normalized distributions, none gets a higher expected score under $b$ than $b$ itself. Because $\hat b_\beta$ is not normalized and names only a set, the property holds in the two forms below.
  \item When $b$ is flat on its support and $m > 1 + \log|\Theta|$, the exact line $\beta^*$ gets the highest score.
  \item When $b$ is not flat, the best set can leave out targets of small mass. Adding a target of mass $\delta$ to $B$ raises $S$ only when $\delta\,\big(m - \log(|B| + 1)\big) > b(B)\,\log\big((|B| + 1)/|B|\big)$. This holds when $\delta$ is above about $1/(m|B|)$.
\end{itemize}

\paragraph{The decision score $S_{\text{dec}}$.}
Let $\tilde b_\beta$ be the belief stated by the line. It holds the targets that agree with the line. Outside MineSweeper, it also holds the targets already excluded by the evidence. Let $a_\beta$ be the teacher's action when $\tilde b_\beta$ is its posterior. Then $S_{\text{dec}}(\beta; b) = \log \pi^*(a_\beta \mid b)$, averaged over $a_\beta$ when the teacher's best actions tie. Every decision problem has its own proper scoring rule, the value of the act implied by a belief~\cite{grunwald2004game,gneiting2007strictly}. $S_{\text{dec}}$ is this rule for the teacher's problem. The exact line implies the teacher's own action and receives the highest score. $S_{\text{dec}}$ is proper but not strictly proper, because lines that imply the same act tie. The sum of a strictly proper and a proper score is strictly proper. Adding $\eta\,S$ with any $\eta > 0$ thus keeps the exact line the unique best line. If the task cannot read a line as a belief, the line receives the floor $-200$.

\paragraph{The mirror term $W$.}
$W(\beta; b) = -m\,|B \setminus B^*|/|B|$ is $-m$ times the line's probability outside the support $B^*$ of $b$, with the line read as uniform on its options. Each task maps the line to $B$ and the posterior to $B^*$ with its own parser. $W$ mirrors the miss term $-m\,(1 - b(B))$ of $S$. A line two times too wide pays about $m/2$, as does a line that leaves out half of the posterior. Because $W = 0$ for the exact line and for every line inside it, the exact line keeps its score and the term needs no new constant. Under $S + W$, the best line under the agent's own uncertainty keeps an option when the option is more likely possible than not.

\paragraph{An example with three lines.}
\Cref{tab:belief} gives the log-score part of the belief term for three lines.

\begin{table}[!htbp]
  \centering
  \footnotesize
  \begin{tabular}{@{}lccc@{}}
    \toprule
    Belief line & $S$ & $W$ & $S + W$ \\
    \midrule
    \bline{Possible: 3, 6, 7} (exact) & $-1.10$ & $0$ & $-1.10$ \\
    \bline{Possible: all} & $-2.77$ & $-40.6$ & $-43.4$ \\
    \bline{Possible: 1, 2, 8} & $-50.0$ & $-50.0$ & $-100.0$ \\
    \bottomrule
  \end{tabular}
  \caption{The log-score part of the belief term for three lines, when $b$ is uniform on the targets 3, 6 and 7 out of $|\Theta| = 16$ and $m = 50$. Under $S$ alone, the open line is almost as good as the exact line. With the mirror term $W$, a line that is too wide is penalized like a wrong line.}
  \label{tab:belief}
\end{table}

\paragraph{Likelihood view.}
Every term of $R$ is a log-probability or a non-positive score, and hence $R \le 0$. With $\alpha = 1$, $e^{R}$ lies in $(0, 1]$ and acts as the likelihood of the decision under the teacher. This view holds for $R$ without the length reward, because that reward adds $C\ell_i \le C/2$ to the weights. $R$ does not depend on the reasoning $z$. The reasoning gets credit only through its effect on the probability of the decisions.

\FloatBarrier
\section{The two learners in detail}
\label{app:learning}

Both learners of \method use the sampled turns, rewards, phases and step limit described in \cref{app:shared}. \Cref{tab:settings} gives the settings of each reported run. They differ only in how they weight the sampled turns. \Cref{app:em} describes EM, and \cref{app:pg} describes the policy gradient. \Cref{app:family} shows that their weights are two limits of one family.

\subsection{Shared by both learners}
\label{app:shared}

\begin{proposition}[Improved policy]
\label{prop:gibbs}
For each history $h$,
\begin{equation}
\begin{aligned}
  \mathbb E_{\pi} R - \alpha\, \mathrm{KL}(\pi \,\|\, \pi_k) &= \alpha \log Z_k(h) - \alpha\, \mathrm{KL}(\pi \,\|\, q_k), \\
  Z_k(h) &= \mathbb E_{x \sim \pi_k}\big[e^{R(h,x)/\alpha}\big],
\end{aligned}
\label{eq:gibbs-identity}
\end{equation}
with $q_k$ as in \mainref{eq:gibbs}. Thus $q_k$ is the unique maximizer of $\mathcal J_k$.
\end{proposition}
\begin{proof}
This is the Gibbs variational principle~\cite{donsker1975asymptotic}. Write $\pi$ for $\pi(\cdot \mid h)$. Then $\mathbb E_\pi R - \alpha\,\mathrm{KL}(\pi \,\|\, \pi_k) = -\alpha\, \mathbb E_\pi \log \frac{\pi}{\pi_k e^{R/\alpha}} = -\alpha\, \mathbb E_\pi \log \frac{\pi}{q_k Z_k}$, which is the right side. The last KL is zero only at $\pi = q_k$.
\end{proof}

\paragraph{Accepted actions.}
The self-normalized weights show only the order of the samples and do not change when the same constant is added to every $R$. We therefore accept a turn only when $\pi^*(u \mid b) \ge \pi^*(u_{\text{best}} \mid b)/20$, where $u_{\text{best}}$ is the action of highest probability under $\pi^*$. Only accepted turns get weight. A turn with a rejected action gets weight 0 whatever its belief line. In the Bayes view, its likelihood is 0. With the belief term, the floor of the line score could otherwise give the weight to a rejected action. This could happen when every line leaves out part of the posterior. Formally, the E-step uses the score $\tilde R(h, x) = R(h, x)$ for an accepted turn and $\tilde R(h, x) = -\infty$ otherwise. \Cref{prop:gibbs}, the Bayes view and \cref{eq:mirror} then hold with $\tilde R$ in place of $R$ at every history with an accepted turn.

\paragraph{Covered histories.}
With the action term alone, a history is covered when one of its turns has an accepted action. With the belief term, that turn must also have a belief line within 2 nats of the exact line. If no turn of the first round covers the history, the trainer samples one more round of $N$ turns, with $2K$ reasonings in all. In either case, it then fits the history with its accepted turns. With the belief term, the two thresholds have two roles. The loose threshold of 2 nats decides whether to sample more. The tight threshold of 0.1 nats in the M-step decides whether a line is a target.

\paragraph{Histories without a target.}
A history has no positive target when it has no accepted turn. With the belief term, it also has no positive target when none of its accepted turns names any of the posterior mass. Such a history gets no weight in the M-step.

\paragraph{Length reward.}
Within one history, let $|z_i|$ be the number of reasoning tokens of turn $i$ and $s_i = \tfrac12 - (|z_i| - \min_j |z_j|)/(\max_j |z_j| - \min_j |z_j|)$, as in Kimi k1.5~\cite{team2025kimi}. A turn is correct when its action is accepted. With the belief term, its line must also be within 2 nats of the exact line. A correct turn gets $\ell_i = s_i$. Shorter correct reasoning thus gets more weight. An incorrect turn gets $\ell_i = \min(0, s_i)$. A short incorrect turn thus gets no bonus, and a long one gets a penalty. All $\ell_i = 0$ when every reasoning has the same length. From the first phase, the E-step weights add $C\,\ell_i$ to $R$ with $C = 4$.

\paragraph{Fitting the turns.}
For both learners, three rules decide how the trainer fits a turn. For the policy gradient, the first rule applies only to turns with positive weights. A turn with a negative weight keeps all of its tokens.
First, with the belief term, a fitted turn trains its whole text only when its belief line is verified. The belief line must satisfy $S(\beta; b) \ge S(\beta^*; b) - 0.1$. Otherwise, its belief-line tokens are not counted and its reasoning gets weight 0. The turn then trains its accepted action alone, with the reasoning and the line kept as context. With the action term alone, every fitted turn trains its whole text.
Second, a turn has two blocks of tokens, the reasoning and the decision. Each block takes its own gradient step with its own clip of the gradient norm. The objective and its minimizer stay the same. The split keeps the long and noisy gradient of the reasoning from setting the pace of the short decision block.
Third, the likelihood $\pi_\phi$ uses the sampler of the policy. For the reasoning, it uses the sampling temperature, the top-$p$ and top-$k$ cuts and the presence penalty. For the decision, it uses the same temperature and presence penalty with no cut. Under a cut, the likelihood of a token is its tempered probability renormalized over the kept tokens. The sampled token always counts as kept. The fixed point is thus the policy used by the agent.

\paragraph{Histories and sample sets.}
In phase $k$, the teacher action replaces the agent action with probability $\xi_k = \xi_0\,\gamma^k$. As in DAgger~\cite{ross2011reduction}, the share of teacher actions in $\mu_k$ thus decreases with $k$. In our runs of two phases, the probability $\xi_k$ falls from 0.5 to 0.465, and in the five phases on Qwen3.5-4B to 0.374. On half of the training scenes, one more episode starts with random evidence actions. Because $q_k$ does not change inside a phase, the samples of $\pi_k$ stay valid for all of its updates. One sample set of $H$ histories thus serves the whole phase. Following the dynamic sampling of DAPO~\cite{yu2025dapo}, we draw the batches in random passes and only from histories that give a gradient. These are the histories in which one of the agent's own turns has an accepted action.

\paragraph{Step limit.}
The M-step fits only a few samples of $q_k$ at each history. It can thus also fit their noise. To limit this, the trainer computes $\mathrm{KL}(\pi_k \,\|\, \pi_\phi)$ after each update. It computes this KL exactly at every position of the probe turns, which are turns of $\pi_k$ from the first batch of the phase:
\begin{equation}
  \Delta_k(\phi) = \frac{1}{|\mathcal P|} \sum_{p \in \mathcal P} \sum_{v} \pi_k(v \mid x_{<p}) \log \frac{\pi_k(v \mid x_{<p})}{\pi_\phi(v \mid x_{<p})} \le \varepsilon,
  \label{eq:steplimit}
\end{equation}
where $\mathcal P$ is the set of token positions of the probe turns, $x_{<p}$ is the prefix before position $p$, and $v$ runs over the vocabulary at the sampling temperature. When $\Delta_k > \varepsilon$, the phase stops. The model after that update is $\pi_{k+1}$. This rule follows the early stop of PPO implementations at 1.5 times a target KL~\cite{schulman2017proximal,achiam2018spinning} and the KL limit in the M-step of MPO~\cite{abdolmaleki2018maximum}.

\subsection{The EM learner}
\label{app:em}

\Cref{alg:pl} gives all the steps of \method with the EM learner.

\paragraph{Bayes view.}
Let $O$ be the event ``the teacher accepts the turn'', with $p(O \mid h, x) = e^{R(h,x)/\alpha}$. This is a probability because $R \le 0$. Then $q_k(x \mid h) = p(x \mid h, O)$ under the prior $\pi_k$. The E-step is Bayes' rule, and the M-step is the M-step of EM~\cite{dempster1977maximum}.

\paragraph{Factors of $q$.}
$R$ depends on the decision $d$ only. Thus
\begin{equation}
\begin{aligned}
  q_k(z, \beta, u \mid h) \;\propto\;& \underbrace{\pi_k(z \mid h)}_{\text{reasoning}}\ \cdot\ \underbrace{\pi_k(\beta \mid h, z)\, e^{R_\beta(\beta;b)/\alpha}}_{\text{belief line}} \\
  &\cdot\ \underbrace{\pi_k(u \mid h, z, \beta)\, \pi^*(u \mid b)^{1/\alpha}}_{\text{action}},
\end{aligned}
\label{eq:factors-full}
\end{equation}
with $R_\beta = S_{\text{dec}} + \eta\,(S + W)$ the belief term of \mainref{eq:score}. For the action, with $\alpha = 1$, $q_k$ is the normalized product of the policy and the teacher policy. With the action term alone, $R_\beta = 0$. A line then gains weight only through the actions that follow it. With the belief term, $q_k$ also moves the policy toward lines with a high belief score. For the reasoning, $q_k(z \mid h) \propto \pi_k(z \mid h)\,\mathbb E_{d \sim \pi_k(\cdot \mid h, z)}[e^{R(h,d)/\alpha}]$. Thus no teacher text is a target for $z$.

\paragraph{The M-step.}
The M-step projects $q_k$ onto the model class by a weighted maximum likelihood on the samples of the agent:
\begin{equation}
\begin{aligned}
  \pi_{k+1} &= \arg\min_{\phi}\ \mathbb E_{h}\, \mathrm{KL}\big(q_k(\cdot\mid h) \,\|\, \pi_\phi(\cdot\mid h)\big) \\
  &\approx \arg\max_{\phi}\ \textstyle\sum_{h}\sum_{i=1}^{N} \hat w_i\, \log \pi_\phi(x_i \mid h).
\end{aligned}
\label{eq:mstep}
\end{equation}
The trainer draws $J$ turns from the weights by systematic resampling~\cite{kitagawa1996monte} and fits each draw with the weight $1/J$. The expected gradient is then that of the weighted loss. The weights are not negative and sum to 1 at each history. The probability of a turn with a small weight goes down only because the probabilities sum to 1. The update thus needs no critic and no group baseline.

\paragraph{Mirror descent.}
The prior $\pi_k$ stays fixed for the updates of phase $k$. Then $\pi_{k+1} = \pi_\phi$. Phase 0 uses the base model $\pi_0$. If each projection is exact and $R$ is fixed,
\begin{equation}
\begin{gathered}
  \pi_k \;\propto\; \pi_{k-1}\, e^{R/\alpha} \quad\Rightarrow\quad \pi_k \;\propto\; \pi_0\, e^{kR/\alpha}, \\
  \pi_k(u \mid h, z, \beta) \;\propto\; \pi_0(u \mid h, z, \beta)\; \pi^*(u \mid b)^{\,k/\alpha}.
\end{gathered}
\label{eq:mirror}
\end{equation}
Phase $k$ thus has the effective temperature $\alpha/k$. As $k$ increases, the policy moves to the maximizers of $R$. These are the teacher's best action, and also the exact line $\beta^*$ with the belief term. In practice, $R$ is not fixed, because $\hat\rho$ is measured again at every update. In addition, $\mu_k$ changes from phase to phase. \Cref{eq:mirror} thus describes the ideal case. Each phase is a mirror-descent step on $R$, as in policy mirror descent~\cite{lan2023policy,tomar2022mirror}. Unlike standard policy mirror descent, the step uses the teacher score $R$ in place of the Q-function of the current policy.

\begin{algorithm*}[t]
\caption{\method with all its steps and the EM learner. This is the full form of \mainref{alg:main}.}
\label{alg:pl}
\begin{algorithmic}[1]
\Statex \textbf{Input:} the base model $\pi_0$, the training scenes, $k_{\max}$ phases, $U$ updates for each phase, the step limit $\varepsilon$
\For{$k = 0, 1, \ldots, k_{\max} - 1$}
  \State $\pi_k \gets$ the current model, the prior of phase $k$.
  \State Run own episodes of $\pi_k$ on new scenes, with the teacher action at probability $\xi_0\,\gamma^k$, and random-step episodes on half of the scenes.
  \State $\mathcal D \gets H$ histories from the turns of these episodes.
  \For{each history $h$ in $\mathcal D$} \Comment{E-step}
    \State Sample $K$ reasonings $z \sim \pi_k(\cdot \mid h)$, and $M$ decisions $d \sim \pi_k(\cdot \mid h, z)$ after each.
    \State Score each turn: $R_i = S_{\text{dec}}(\beta_i; b) + \eta\,[S(\beta_i; b) + W(\beta_i; b)] + \log \pi^*(u_i \mid b)$.
    \State If no accepted turn has a line within 2 nats of the exact line, sample one more round.
    \State $w_i = \mathrm{softmax}_i\big((R_i + C\,\ell_i)/\alpha\big)$ over the accepted turns, $w_i = 0$ for the others.
  \EndFor
  \For{each of the $U$ updates} \Comment{M-step}
    \State Take $L$ histories from $\mathcal D$ in random passes, and $J$ draws from $w$ at each.
    \State Update $\phi$ to increase $\sum_h \sum_i w_i \log \pi_\phi(x_i \mid h)$. A turn with an unverified line trains its action only.
    \State If $\Delta_k(\phi) > \varepsilon$ on the probe turns of $\pi_k$: stop the phase.
  \EndFor
\EndFor
\State \Return $\pi_\phi$
\end{algorithmic}
\end{algorithm*}

\subsection{The policy-gradient learner}
\label{app:pg}
The policy gradient uses the histories, turns and rewards above. It replaces the weights of the E-step by those of \mainref{eq:adv}. Let $r_i = R(h, x_i) + C\ell_i$ be the reward of turn $i$ with its length term. At each history, we compute the advantages $A_i = \tilde r_i - \tfrac{1}{N-1}\sum_{j \ne i} \tilde r_j$ over all of its sampled turns, the extra round included. Here $\tilde r_i = \max(r_i, \max_j r_j - c)$ raises every reward to at least the best one minus $c = 20$ nats. A turn with a very low reward thus cannot raise the advantages of all the others. The advantages also stay in $[-c, c]$. The weight of a turn is $w_i = s A_i / N$, with one factor $s$ per mini-batch. We do not divide by the standard deviation of the group, because the rewards are already in nats. Dividing would also give large weights to a group of nearly equal rewards. A turn with a rejected action gets a weight of at most zero. Its probability can thus be lowered but never raised. As with the weights of EM, the factor $s$ makes the absolute weights of a mini-batch sum to the number of its histories with a target. This factor only balances the mini-batches, because we use Adam~\cite{kingma2015adam} and clip the gradient norm of each block. Unlike EM, the policy gradient does not draw turns. It fits every turn with a nonzero weight. A phase uses its turns for several updates. A block with a positive weight therefore stops contributing once its mean log-ratio to $\pi_k$ per counted token rises above $\log(1+\delta)$. A block with a negative weight stops once this log-ratio falls below $\log(1-\delta)$. We use $\delta = 0.2$ and compute the log-ratio once per phase. This rule follows the clipped probability ratio of PPO~\cite{schulman2017proximal}.

In our runs, the policy gradient fitted about 1,300 turns per update and EM about 300. This is because the policy gradient gives a weight to every sampled turn, and the draws of EM fall on the few turns of high reward. The steps of the policy gradient were also larger.
\paragraph{The reported runs of \method.}
\method in \mainref{tab:main} is trained with the action term in \mainref{eq:score} alone. As a result, the rules above for the belief term do not apply to it. They apply to \method (EM) and to the ablation runs with the belief term in \mainref{tab:belief_term}. For the two Qwen3-VL models, a phase has 25 updates and a sample set of 1,200 histories, of which 800 are used. The training takes 50 updates. Qwen3.5-4B learns faster. With phases of 25 updates, its reasoning became very short. We therefore give it a new sample set of 240 histories every 5 updates, of which 160 are used. We train it for 25 updates with no length reward. \Cref{app:rules} describes three more rules in the reward that we tested on Qwen3.5-4B.

\subsection{Rules against steps without evidence}
\label{app:rules}
\paragraph{Why we tested the rules.}
With phases of 5 updates, the agent of Qwen3.5-4B learned a second way to delay its answer. It took steps that bring no evidence, such as turning on the spot in MiniGrid. The action term lets such a step pass. In the teacher's score, a step that brings no evidence costs only its time discount, which is 0.2 nats in most tasks and 0.5 nats in MineSweeper. The acceptance rule accepts every turn whose score is within $\log 20 = 3.0$ nats of the teacher's best action. A step without evidence is therefore still accepted. The measured answer accuracy $\hat\rho$ of \cref{eq:accuracy} can add to this. Where the agent's answers are often wrong, $\hat\rho$ is low. Answering then has a low value, and the teacher itself asks for more steps.

\paragraph{The three rules.}
All three rules change the reward of the action term, and all three are computed from the exact posterior.
\begin{enumerate}[leftmargin=*,itemsep=1pt,topsep=2pt]
\item \emph{Exact answer accuracy.} The estimate $\hat\varphi$ of \cref{eq:accuracy} stays at its prior mean of 1, so $\hat\rho(b) = b_{\max}$. This is the accuracy of an agent that always reads its evidence correctly. The teacher's choice between a step and an answer then depends on the evidence alone and not on the agent's own errors.
\item \emph{A cap on steps without evidence.} An illegal action gets a teacher score of at most the best action's score minus 3.5 nats. So does a step that brings no expected evidence and is not one of the teacher's best steps. The acceptance rule therefore rejects both. The teacher's own moves in the training games follow the same capped scores.
\item \emph{The teacher's turn as a positive example.} With the cap, more histories have no accepted turn of the agent. At such a history, the update can only lower the probability of the agent's turns. We then add the teacher's own belief line and action to the turns of this history as a positive example.
\end{enumerate}

\paragraph{Results.}
We trained Qwen3.5-4B with the three rules in the same way as the reported run, with phases of 5 updates and 25 updates, and we tested it on the same three test seeds. \Cref{tab:rules} compares the two runs. With the rules, the accuracy is 1.6\% higher on HiddenGames, 0.2\% lower on Pro-Worlds and 2.9\% higher on RoboQuest. In a paired test over the test games, only the gain on RoboQuest is significant ($p = 0.014$). On the HiddenGames tasks with an answer, a larger share of the steps adds evidence, 28.0\% compared to 21.2\%, and the agent takes fewer steps per episode. Fewer episodes end without an answer on all three benchmarks, for example 6.0\% compared to 11.6\% on Pro-Worlds. On HiddenGames and Pro-Worlds, the agent reaches sufficient evidence about as often as without the rules. On RoboQuest, it answers with sufficient evidence in 71.3\% of the episodes, compared to 68.8\%. We report the run without the rules in \mainref{tab:main}, so that all three models are trained with the same reward.

\begin{table}[!htbp]
  \centering
  \footnotesize
  \setlength{\tabcolsep}{4pt}
  \begin{tabular}{@{}l rrr@{}}
    \toprule
    & HiddenGames & Pro-Worlds & RoboQuest \\
    \midrule
    \multicolumn{4}{@{}l}{\emph{Accuracy (\%)}} \\
    \quad Base (w/ belief) & 38.1 & 23.9 & 41.3 \\
    \quad \method & 48.5 & 32.1 & 54.0 \\
    \quad \method with the rules & 50.0 & 31.9 & 56.9 \\
    \quad Difference & $+1.6$ & $-0.2$ & $+2.9$ \\
    \quad 95\% interval & $[-0.4, 3.6]$ & $[-4.2, 3.9]$ & $[0.5, 5.3]$ \\
    \quad $p$ & 0.11 & 0.91 & 0.014 \\
    \midrule
    \multicolumn{4}{@{}l}{\emph{Steps that add evidence (\%)}} \\
    \quad \method & 21.2 & 5.1 & 21.3 \\
    \quad \method with the rules & 28.0 & 5.0 & 21.9 \\
    \multicolumn{4}{@{}l}{\emph{Steps per episode}} \\
    \quad \method & 7.8 & 29.3 & 9.2 \\
    \quad \method with the rules & 6.9 & 28.8 & 9.2 \\
    \multicolumn{4}{@{}l}{\emph{Episodes without an answer (\%)}} \\
    \quad \method & 19.3 & 11.6 & 7.1 \\
    \quad \method with the rules & 15.1 & 6.0 & 5.6 \\
    \multicolumn{4}{@{}l}{\emph{Episodes answered with sufficient evidence (\%)}} \\
    \quad \method & 52.7 & 14.4 & 68.8 \\
    \quad \method with the rules & 53.6 & 12.8 & 71.3 \\
    \bottomrule
  \end{tabular}
  \caption{\textbf{Rules against steps without evidence (Qwen3.5-4B).} \method is the run of \mainref{tab:main}. \method with the rules adds the three rules of \cref{app:rules}. Both runs are tested on the same three seeds. \emph{Difference}: accuracy with the rules minus accuracy without them. The 95\% interval and $p$ come from a paired test over the test games, with each game's accuracy averaged over the three seeds. Accuracy covers all tasks. The other rows cover the HiddenGames tasks with an answer (Tabletop, MiniGrid and MineSweeper) and the RoboQuest tasks with countable requirements. On RoboQuest, a step adds evidence when it meets a new requirement, and the evidence is sufficient when every requirement is met.}
  \label{tab:rules}
\end{table}

\subsection{One family of weights}
\label{app:family}
The two weightings are two limits of one family. Let $r_1, \ldots, r_N$ be the rewards of the turns at a history, $\bar r$ their mean, and $\omega_i^\alpha = e^{r_i/\alpha} / \sum_j e^{r_j/\alpha}$ the EM weights at temperature $\alpha$, without the acceptance rule.
\begin{proposition}
\label{prop:limits}
(i) As $\alpha \to 0$, $\omega^\alpha$ puts all of its weight on the turns of highest reward. (ii) As $\alpha \to \infty$, $\alpha\,(\omega_i^\alpha - 1/N) \to (r_i - \bar r)/N$. (iii) The leave-one-out advantage is $r_i - \frac{1}{N-1}\sum_{j \ne i} r_j = \frac{N}{N-1}(r_i - \bar r)$.
\end{proposition}
\begin{proof}
For (i), $\omega_i^\alpha / \omega_j^\alpha = e^{(r_i - r_j)/\alpha}$ tends to $0$ whenever $r_i < r_j$. For (ii), $e^{r/\alpha} = 1 + r/\alpha + O(\alpha^{-2})$, which gives $\omega_i^\alpha = \big(1 + (r_i - \bar r)/\alpha\big)/N + O(\alpha^{-2})$. For (iii), $\sum_{j \ne i} r_j = N\bar r - r_i$.
\end{proof}
By (ii) and (iii), the weights of the policy gradient at each history are $\lim_{\alpha \to \infty} \alpha\,(\omega_i^\alpha - 1/N)$ times the positive factor $sN/(N-1)$. This holds without the floor, the limit $c$ and the acceptance rule. The policy gradient thus uses the centered EM weights $\omega_i^\alpha - 1/N$ at high temperature. EM uses the uncentered weights. At the first update of a phase, $\pi_\phi = \pi_k$ and $\mathbb E_{x \sim \pi_k}\big[\nabla_\phi \log \pi_\phi(x \mid h)\big] = 0$. The centered and the uncentered weights thus give the same expected gradient of \mainref{eq:wll}. Their steps differ by a term of mean zero. By (i), EM at a low temperature fits only the best turn of each history, as rejection-sampling fine-tuning does~\cite{zelikman2022star}. Our EM uses $\alpha = 1$. At this temperature, $e^{R}$ is the teacher's probability of approving the turn.

The two learners differ in how far they move from $\pi_k$ on the same turns. The EM weights are non-negative and sum to one at each history. The loss of \mainref{eq:wll} is thus smallest when $\pi_\phi$ matches them on the sampled turns. The fit then settles near $q_k$. The policy gradient also has negative weights, which keep lowering the probability of poor turns. Only the ratio clip and the step limit of \cref{eq:steplimit} bound its steps.

\FloatBarrier
\FloatBarrier
\section{What the lookahead adds}
\label{app:lookahead}

To see what the lookahead adds without training a model, we play the 756 test games of HiddenGames with the teacher alone. The teacher knows the exact posterior and answers like a perfect answerer. Its $\hat\rho$ therefore stays at its prior. It takes the action of highest value. \Cref{tab:lookahead} compares the step chosen by the lookahead with a random legal step. Both rules stop at the same point, where the teacher would answer.

\begin{table}[!htbp]
  \centering
  \begin{minipage}[t]{0.52\linewidth}
    \centering
    \footnotesize
    \setlength{\tabcolsep}{2.5pt}
    \begin{tabular}{@{}lcccc@{}}
      \toprule
      & \multicolumn{2}{c}{Accuracy (\%)} & \multicolumn{2}{c}{Steps} \\
      \cmidrule(lr){2-3} \cmidrule(l){4-5}
      Task & lookahead & random & lookahead & random \\
      \midrule
      MineSweeper & 83.0 & 10.7 & 11.0 & 21.3 \\
      MiniGrid KeyCorridor & 100 & 49.0 & 8.3 & 57.6 \\
      FrozenLake & 100 & 26.8 & 6.5 & 8.9 \\
      Sokoban & 98.4 & 24.2 & 6.1 & 12.5 \\
      Tabletop lamps & 100 & 100 & 3.5 & 3.5 \\
      Tabletop cups & 100 & 100 & 2.2 & 2.2 \\
      Tabletop blocks & 100 & 100 & 1.9 & 1.9 \\
      \midrule
      All 756 games & 97.5 & 56.2 & 5.7 & 15.1 \\
      \bottomrule
    \end{tabular}
    \caption{The teacher alone on the test games of HiddenGames. It takes the step of its lookahead or a random legal step. The results of the random steps are averaged over three seeds.}
    \label{tab:lookahead}
  \end{minipage}\hfill
  \begin{minipage}[t]{0.46\linewidth}
    \centering
    \footnotesize
    \setlength{\tabcolsep}{3pt}
    \begin{tabular}{@{}l c c l@{}}
      \toprule
      Game & Board & Answers & Split \\
      \midrule
      Face, Turtle, Rocket, Balloon & $10\times10$ & 8 & training \\
      Sailboat, Barn, Lantern & $10\times10$ & 8 & training \\
      Fish & $9\times9$ & 8 & training \\
      Bird, Bell, Bee, Candle & $10\times10$ & 8 & other \\
      Kaleidoscope, Recipe & $9\times9$ & 8 & other \\
      Jigsaw & $8\times8$ & 10 & other \\
      \bottomrule
    \end{tabular}
    \caption{\textbf{The 15 Pro-Worlds test games.} The board size is in sites. The split follows the benchmark.}
    \label{tab:proworlds}
  \end{minipage}
\end{table}

As \cref{tab:lookahead} shows, the lookahead matters where the steps differ in what they reveal. In MineSweeper, MiniGrid, FrozenLake and Sokoban, random steps lose most games. In MineSweeper, every failure is a revealed mine. Every step in the Tabletop tasks splits the possible answers in the same way. The choice of step thus does not matter there. A lookahead of depth 1 gives the same result as depth 2. With a perfect answerer, the teacher answers exactly when the posterior identifies the answer. It does so in all 500 games with an answer action. Answering at the first turn instead solves 18.0\% of these games. With a less reliable answerer, $\hat\rho$ lowers the value of answering. As \cref{app:teacher} shows, the teacher then asks for more evidence before it answers.

\FloatBarrier
\section{Benchmarks in detail}
\label{app:benchmarks}

\paragraph{HiddenGames.}
HiddenGames is a mixture of seven tasks from different agent benchmarks. In each task, the agent must find a hidden target from what it observes. MineSweeper~\cite{wang2025popgym} hides the mines of a small board. In MiniGrid KeyCorridor~\cite{chevalier2023minigrid}, the agent must find the room that holds the key. In FrozenLake~\cite{wang2025vagen}, we hide the goal and the distant holes. In Sokoban~\cite{wang2025vagen}, we hide the target square. The agent must thus explore before it can plan. The three Tabletop tasks are rendered scenes of a table~\cite{tao2024maniskill3}. The agent must find which button lights which lamp, which cup hides a named object as in \mainref{fig:turn}, or which block slides farther. It sees only two camera images after each action. The environment of every task knows the prior and the likelihood of each observation. The teacher of \mainref{sec:teacher} can thus compute the exact posterior. We train and test on all seven tasks. The test split consists of 756 games whose seeds, maps and Tabletop colors are never seen in training.

\paragraph{Pro-Worlds.}
We introduce Pro-Worlds, a benchmark of how visual agents gather, retain and combine evidence before they commit to an answer. Each game hides one of a few answers, for example which emotion a face shows or which fish a picture shows. The game draws its world as a grid of sites, each a tile of $16\times16$ color cells. The agent sees one site at a time through a window, rendered as a $256\times256$ picture. At each turn, the agent either moves the window by one site with \texttt{ACTION: MOVE UP}, \texttt{DOWN}, \texttt{LEFT} or \texttt{RIGHT}, or submits an answer with \texttt{ACTION: SUBMIT}. A move at the edge keeps the window in place but still counts. A play allows at most 35 moves. It ends with a submission, with a move when no moves are left, or after 43 turns. The prompt gives the rules of the game, the list of answers and every picture seen so far with its position. We place the start where the first view never decides the answer.

The engine knows every possible world of a game and can thus compute the exact posterior. This posterior holds the worlds consistent with every site seen so far, with the same prior weight for each answer. As in the other benchmarks, a submission is premature when the posterior of the top answer is below 0.95. \Cref{tab:proworlds} lists the 15 test games. We test on 20 plays of each game, 300 plays in all. Eight of the games are the training games of the benchmark, drawn at random from the 15. However, no model in this paper trains on Pro-Worlds.

\paragraph{RoboQuest.}
RoboQuest~\cite{liu2026roboquest} consists of ten mobile manipulation tasks built around search, inspection and interactive testing. In each task, such as finding the odd parcel by weighing, the agent must gather information through physical interaction before it decides to finish. Liu~\etal report that agents often stop exploring too early~\cite{liu2026roboquest}. Unlike the other two benchmarks, RoboQuest runs in a physics simulator of real kitchens. Its possible states cannot be listed, and there is no exact posterior. Many of its answers are also actions in the scene rather than a choice among listed options. The belief lines of all agents are therefore free text, such as \texttt{Possible: blue water bottle may be in drawer 2 or another compartment}. We use RoboQuest to test whether what an agent learns from exact posteriors transfers to a task without one. We use the text version of its decision-level setting, where the agent chooses high-level actions and scripted helpers carry them out. We test on 50 games of each task, 500 games in all.

\FloatBarrier
\section{Results per task}
\label{app:pertask}

\Cref{tab:task-hg,tab:task-pw,tab:task-rq} break down the accuracy of \mainref{tab:main} by task. A learner of \method is the best method in five of the seven HiddenGames tasks, 14 of the 15 Pro-Worlds games, and 9 of the 10 RoboQuest tasks. On HiddenGames, the largest gains of \method over Base occur in the Tabletop tasks, with 51.0\% on lamps and 34.8\% on cups, while on RoboQuest the largest gains occur in the search tasks, with 39.6\% on Search Room and 55.5\% on Blackout Search. On Pro-Worlds, \method reaches 30.2\% on both the seen and the unseen games, compared to 17.9\% and 16.9\% for Base.

\begin{table*}[!htbp]
  \centering
  \footnotesize
  \setlength{\tabcolsep}{4pt}
  \begin{tabular}{@{}l rrrrrrrr@{}}
    \toprule
    Task & Base & Base (w/ b.) & SFT & PPO & GRPO & PPO (b.+a.) & \method (EM) & \method \\
    \midrule
    Odd Parcel & 24.4 & 30.2 & 28.2 & 17.8 & 22.2 & 35.1 & 39.3 & \textbf{60.7} \\
    Painted Cubes & 8.0 & 13.1 & 14.0 & 5.1 & 10.0 & 27.6 & 18.9 & \textbf{45.6} \\
    Marked Mugs & 26.2 & 29.3 & 38.7 & 19.5 & 25.8 & 38.5 & 42.2 & \textbf{51.8} \\
    Unfamiliar Containers & 4.5 & 4.4 & 4.4 & 3.3 & 2.5 & 12.9 & 9.8 & \textbf{36.2} \\
    Search Room & 46.2 & 57.6 & 58.2 & 21.1 & 46.2 & 84.4 & 75.1 & \textbf{85.8} \\
    Blackout Search & 27.6 & 41.8 & 44.5 & 14.2 & 24.0 & 77.6 & 58.9 & \textbf{83.1} \\
    Locked Storage & 75.1 & 71.6 & 71.6 & 52.0 & 68.9 & 78.6 & 81.1 & \textbf{83.6} \\
    Puzzle Box & 41.1 & 41.3 & 33.3 & 26.9 & 31.3 & 39.1 & 34.5 & \textbf{46.7} \\
    Stamps & 1.8 & 2.5 & 0.5 & 0.5 & 0.4 & \textbf{3.5} & 2.7 & 3.1 \\
    Wobbly Stand & 92.4 & 91.8 & 90.9 & 81.8 & 88.7 & 93.3 & 94.4 & \textbf{96.7} \\
    \midrule
    \textit{Overall} & 34.7 & 38.4 & 38.4 & 24.2 & 32.0 & 49.1 & 45.7 & \textbf{59.3} \\
    \bottomrule
  \end{tabular}
  \caption{\textbf{Accuracy (\%) per task on RoboQuest.} Each number is the mean over the three models and three test seeds. Bold marks the best method in each row. Base (w/ b.) denotes Base (w/ belief), PPO (b.+a.) denotes PPO (belief + action), and SFT, PPO and GRPO are the action-only baselines.}
  \label{tab:task-rq}
\end{table*}

\begin{table*}[!htbp]
  \centering
  \footnotesize
  \setlength{\tabcolsep}{4pt}
  \begin{tabular}{@{}l rrrrrrrr@{}}
    \toprule
    Task & Base & Base (w/ b.) & SFT & PPO & GRPO & PPO (b.+a.) & \method (EM) & \method \\
    \midrule
    Face & 15.0 & 17.2 & 11.7 & 10.0 & 18.9 & 16.7 & 18.3 & \textbf{19.4} \\
    Fish & 8.3 & 14.5 & 6.1 & 1.1 & 7.8 & 13.3 & \textbf{16.7} & 11.1 \\
    Turtle & 22.8 & 27.2 & 19.5 & 18.9 & 14.4 & 23.3 & 37.8 & \textbf{38.3} \\
    Balloon & 21.7 & 30.6 & 26.1 & 13.3 & 14.5 & 27.2 & \textbf{42.2} & 40.6 \\
    Rocket & 22.2 & 21.1 & 21.7 & 20.6 & 17.8 & 23.9 & \textbf{32.2} & 31.1 \\
    Sailboat & 20.0 & 30.0 & 16.1 & 9.4 & 15.5 & 25.6 & \textbf{33.9} & 32.8 \\
    Barn & 12.8 & 16.1 & 18.9 & 10.5 & 12.2 & 22.8 & 30.5 & \textbf{40.6} \\
    Lantern & 20.0 & 19.4 & 18.3 & 11.7 & 15.5 & 17.8 & 15.6 & \textbf{27.8} \\
    \midrule
    \textit{Training split (8 games)} & 17.9 & 22.0 & 17.3 & 11.9 & 14.6 & 21.3 & 28.4 & \textbf{30.2} \\
    \midrule
    Jigsaw & 5.0 & 6.1 & 8.3 & 3.3 & 6.1 & 6.1 & 7.2 & \textbf{19.4} \\
    Kaleidoscope & 16.1 & 17.2 & 17.8 & 22.2 & 16.1 & 21.1 & \textbf{27.2} & 25.0 \\
    Bird & 15.6 & 13.3 & \textbf{16.1} & 11.1 & 15.0 & 11.7 & 13.3 & 6.7 \\
    Recipe & 8.9 & 17.2 & 11.7 & 11.1 & 13.3 & 17.2 & 24.4 & \textbf{27.2} \\
    Bell & 25.6 & 30.6 & 23.9 & 18.9 & 27.2 & 33.3 & 51.1 & \textbf{56.1} \\
    Bee & 25.6 & 30.6 & 22.2 & 17.2 & 16.7 & 26.1 & 32.2 & \textbf{39.4} \\
    Candle & 21.7 & 21.7 & 18.3 & 13.3 & 15.0 & 19.5 & 35.5 & \textbf{37.8} \\
    \midrule
    \textit{Other split (7 games)} & 16.9 & 19.5 & 16.9 & 13.9 & 15.6 & 19.3 & 27.3 & \textbf{30.2} \\
    \midrule
    \textit{Overall} & 17.4 & 20.9 & 17.1 & 12.9 & 15.1 & 20.4 & 27.9 & \textbf{30.2} \\
    \bottomrule
  \end{tabular}
  \caption{\textbf{Accuracy (\%) per task on Pro-Worlds.} Each number is the mean over the three models and three test seeds. Bold marks the best method in each row. The training split and the other split follow the benchmark. No model in this paper trains on Pro-Worlds.}
  \label{tab:task-pw}
\end{table*}

\begin{table*}[!htbp]
  \centering
  \footnotesize
  \setlength{\tabcolsep}{4pt}
  \begin{tabular}{@{}l rrrrrrrr@{}}
    \toprule
    Task & Base & Base (w/ b.) & SFT & PPO & GRPO & PPO (b.+a.) & \method (EM) & \method \\
    \midrule
    MineSweeper & 1.3 & 1.7 & 1.2 & 0.8 & 0.7 & \textbf{2.8} & 1.0 & 0.2 \\
    MiniGrid KeyCorridor & 31.5 & 31.6 & 30.8 & 32.9 & 32.4 & \textbf{33.4} & 30.9 & 33.0 \\
    FrozenLake & 41.7 & 43.1 & 40.3 & 35.3 & 37.9 & 41.6 & 41.9 & \textbf{48.0} \\
    Sokoban & 24.6 & \textbf{26.1} & 24.1 & 25.8 & 24.3 & 25.7 & \textbf{26.1} & 25.3 \\
    Tabletop lamps & 7.4 & 24.0 & 20.6 & 1.2 & 18.9 & 29.4 & 44.7 & \textbf{58.4} \\
    Tabletop cups & 35.0 & 39.8 & 53.0 & 41.9 & 38.3 & 49.9 & 63.3 & \textbf{69.8} \\
    Tabletop blocks & 26.6 & 32.6 & 26.6 & 29.1 & 33.0 & 32.3 & 33.7 & \textbf{36.7} \\
    \midrule
    \textit{Overall} & 24.7 & 28.8 & 28.4 & 24.4 & 26.8 & 30.9 & 34.5 & \textbf{38.6} \\
    \bottomrule
  \end{tabular}
  \caption{\textbf{Accuracy (\%) per task on HiddenGames.} Each number is the mean over the three models and three test seeds. Bold marks the best method in each row.}
  \label{tab:task-hg}
\end{table*}

\section{The posterior teacher on every task}
\label{app:teacher-tasks}

Every task has its own teacher. The value, the lookahead and the choice between answering and taking a step are the same in all tasks. Each teacher uses the posterior of its own task. \Cref{tab:teacher-tasks} gives the accuracy of each teacher alone with the exact posterior and a perfect answerer, as in the reference row of \mainref{tab:main}. In MineSweeper, every failure of the teacher is a revealed mine. In Sokoban, its two failures used up the move budget. The teacher of Pro-Worlds builds the posterior of each game from its game engine. RoboQuest has no exact posterior and thus no teacher. It tests whether what \method learns transfers to a task without one.

\begin{table*}[!htbp]
  \centering
  \footnotesize
  \begin{tabular}[t]{@{}lcc@{}}
    \toprule
    \multicolumn{3}{@{}l}{\textit{HiddenGames, 756 test games}} \\
    Task & Accuracy (\%) & Steps \\
    \midrule
    MineSweeper & 83.0 & 11.0 \\
    MiniGrid KeyCorridor & 100 & 8.3 \\
    FrozenLake & 100 & 6.5 \\
    Sokoban & 98.4 & 6.1 \\
    Tabletop lamps & 100 & 3.5 \\
    Tabletop cups & 100 & 2.2 \\
    Tabletop blocks & 100 & 1.9 \\
    \midrule
    All & 97.5 & 5.7 \\
    \bottomrule
  \end{tabular}
  \hspace{2em}
  \begin{tabular}[t]{@{}lcc@{\hspace{1.5em}}lcc@{}}
    \toprule
    \multicolumn{6}{@{}l}{\textit{Pro-Worlds, 15 games $\times$ 20 test plays}} \\
    Game & Acc.\ (\%) & Moves & Game & Acc.\ (\%) & Moves \\
    \midrule
    Face & 100 & 14.2 & Jigsaw & 100 & 15.2 \\
    Kaleido & 100 & 17.8 & Bird & 100 & 17.4 \\
    Fish & 100 & 13.2 & Recipe & 100 & 14.6 \\
    Turtle & 100 & 13.9 & Balloon & 100 & 12.4 \\
    Rocket & 100 & 19.4 & Bell & 100 & 10.9 \\
    Bee & 100 & 11.2 & Candle & 100 & 13.8 \\
    Sailboat & 100 & 15.3 & Barn & 100 & 18.2 \\
    Lantern & 100 & 12.3 & & & \\
    \midrule
    All & 100 & 14.7 & & & \\
    \bottomrule
  \end{tabular}
  \caption{The posterior teacher alone on every test task. It uses the exact posterior and a perfect answerer.}
  \label{tab:teacher-tasks}
\end{table*}

\section{Settings of the \method runs}
\label{app:settings}

\paragraph{Shared configuration.}
All methods share one configuration for each model. Every run of every method caps the reasoning at 8,192 tokens in training and in testing. Each run also samples with the settings of \cref{tab:settings}. Each run trained on one node with 8 GPUs, either H200 GPUs with 141\,GB or A100 GPUs with 80\,GB.

\begin{table}[!htbp]
  \centering
  \begin{minipage}[t]{0.46\linewidth}
    \centering
    \footnotesize
    \setlength{\tabcolsep}{3pt}
    \begin{tabular}{@{}lr@{}}
      \toprule
      Setting & Value \\
      \midrule
      KL weight $\alpha$ & 1 \\
      Teacher temperature $\tau$ (nats) & 0.1 \\
      Lookahead depth in Tabletop & 2 \\
      Miss penalty $m$ (nats) & 50 \\
      Weight $\eta$ of $S + W$ & 0.1 \\
      Reasonings $K$, decisions $M$ per history & 8, 4 \\
      Draws $J$ per history & 32 \\
      Acceptance ratio & 1/20 \\
      Length reward $C$ & 4 \\
      \quad \method, Qwen3.5-4B & 0 \\
      Line verification margin (nats) & 0.1 \\
      Histories per update $L$ & 32 \\
      Sample set: drawn / kept histories & 300 / 200 \\
      \quad \method, Qwen3-VL & 1,200 / 800 \\
      \quad \method, Qwen3.5-4B & 240 / 160 \\
      Updates per phase $U$, total updates & 25, 50 \\
      \quad \method, Qwen3.5-4B & 5, 25 \\
      Step limit $\varepsilon$ (nats per token), probe turns & 0.015, 32 \\
      Teacher action rate $\xi_0$, decay $\gamma$ & 0.5, 0.93 \\
      Learning rate, weight decay (AdamW) & $10^{-5}$, 0.01 \\
      LoRA rank & 16 \\
      Sampling temperature, top-$p$, top-$k$ & 1.0, 0.95, 20 \\
      Presence penalty & 1.5 \\
      Thinking budget (training and test, tokens) & 8,192 \\
      Move cost $c(1)$ & \\
      \quad MiniGrid, FrozenLake, Sokoban, Tabletop & 0.98 \\
      \quad MineSweeper & 0.95 \\
      \bottomrule
    \end{tabular}
    \caption{Settings of the reported \method runs.}
    \label{tab:settings}
  \end{minipage}\hfill
  \begin{minipage}[t]{0.53\linewidth}
    \centering
    \footnotesize
    \setlength{\tabcolsep}{3pt}
    \begin{tabular}{@{}l ccc@{}}
      \toprule
      Setting & SFT & GRPO & PPO \\
      \midrule
      Trained tokens & action & all & all \\
      Reward & -- & $0.999^t\cdot\mathbb{1}[\text{correct}]$ & same \\
      Learning rate & $10^{-5}$ & $10^{-5}$ & $10^{-5}$ \\
      Weight decay & 0 & 0.01 & 0.01 \\
      LoRA rank, alpha & 16, 32 & 16, 32 & 16, 32 \\
      Per update & 32 won games & 4 starts $\times$ 8 & 32 episodes \\
      KL loss & -- & 0.01 & -- \\
      Value function & -- & -- & own LoRA \\
      GAE $\gamma$, $\lambda$ & -- & -- & 1, 1 \\
      Value clip & -- & -- & 0.5 \\
      \bottomrule
    \end{tabular}
    \caption{\textbf{Settings of the baselines.} ``All'' means the reasoning and the action.}
    \label{tab:baselines}
  \end{minipage}
\end{table}

\section{The PPO (belief + action) baseline}
\label{app:ppo-belief}

PPO (belief + action) states a belief before each action. An LLM judge rewards this belief at every turn. A turn consists of the reasoning, a belief state and an action. The belief state is a list of bullet points inside \texttt{<belief\_state>} tags, each with one word of certainty such as \texttt{confirmed}, \texttt{probable} or \texttt{ruled out}. The action is inside \texttt{<action>} tags. The turn receives the same input as the other methods, with the rules of the task, the actions and observations so far and the current observation. It also receives the belief state of the previous turn.

The belief reward of a turn is $R_b = f\,(s_t + s_c + d + o)/4$. Here $f$ scores the format of the belief state by the number of bullets, a certainty word in each bullet, no repeats and the length. The judge score $s_t$ measures whether the new belief state takes in the newest observation. The judge score $s_c$ measures whether its claims are true. $d$ is the entropy of the certainty words, and $o$ is the outcome reward. To score $s_c$, the judge checks the claims against a text description of the true hidden state. Only the judge sees this description. The judge is DeepSeek V4.1 Flash at temperature 0, without thinking. A failed judge call receives the batch mean of its term.

The tokens of the belief state are trained on $R_b$, normalized over the belief states of the batch. The reasoning and action tokens are trained with PPO on the outcome reward $0.999^t\cdot\mathbb{1}[\text{correct}]$. PPO uses a learned value function and GAE~\cite{schulman2016high} with $\gamma=\lambda=1$. Training uses LoRA with rank 16 and alpha 32, a learning rate of $10^{-5}$ and 50 updates of 16 episodes each, with no warm start. At test time, the prompt keeps the rules of the belief state but drops the list of task-specific fields and the format example used in training.

\section{Settings of the baselines}
\label{app:baseline-settings}

\Cref{tab:baselines} gives the settings of the trained baselines, which are all action-only. They train LoRA adapters for 50 updates on the training split of HiddenGames, with the sampling settings and the thinking budget of \cref{tab:settings}. GRPO and PPO train with verl~\cite{sheng2025hybridflow}. SFT trains on the teacher's action in its winning games, after the base model's own reasoning. Its learning rate follows a cosine schedule with 10 warm-up steps.

\section{\texorpdfstring{Why \method works: details}{Why RLVB works}}
\label{app:working}

This section provides the details and additional tests underlying \mainref{sec:more_looks}. Unless a table or figure states otherwise, all numbers are for Qwen3-VL-4B.

\paragraph{Measures.}
A step refers to a move on Pro-Worlds, a lift or press in the Tabletop tasks, and an action such as opening a drawer or weighing two parcels on RoboQuest. A step adds evidence when it changes the exact posterior or, on RoboQuest, when it meets a new evidence requirement of the task, such as the agent now seeing a target object. On HiddenGames, we use the five tasks with an answer: the three Tabletop tasks, MiniGrid, and MineSweeper. On RoboQuest, requirements can be counted in eight of the ten tasks, so we omit Stamps and Puzzle Box from these measures. The \emph{chance of a correct answer} is the exact posterior of the top answer, representing the accuracy of an agent that reads its evidence without error.

\paragraph{Reading the belief lines on RoboQuest.} On RoboQuest, belief lines are free text. The task has no posterior. We read every belief line in the test episodes of Base (w/ belief) and \method on the 361 of 400 instances that at least one method solves. These are 56,556 lines from the three models and three test seeds. Language-model readers judge each line against the trace of an episode that solved the same instance. The readers do not see which agent wrote the line. They record whether the line names a single answer or leaves part of the answer open. They also record whether it keeps the true answer. When the agent submits at that turn, they record whether the submission gives the answer that the line names. A second reader labels 2,595 of these lines. It agrees on whether the line names a single answer for 2,539 of them, on whether it keeps the true answer for 2,514, and on whether the submission gives the named answer for 2,562.

\subsection{The evidence at the answer}

\begin{figure}[!htbp]
  \centering
  \includegraphics[width=0.97\linewidth]{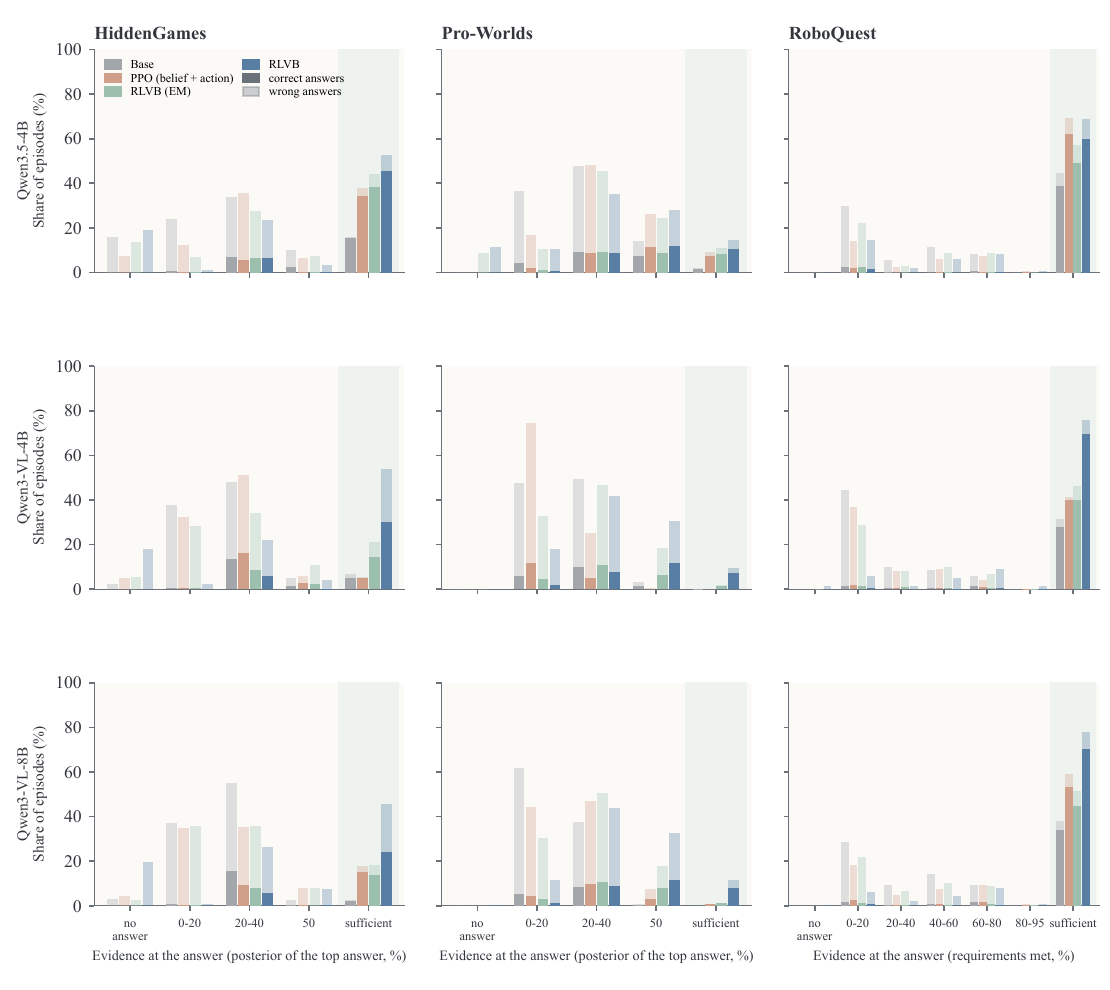}
  \caption{Evidence when each model answers. Each bar shows the share of episodes answered at that evidence level. The solid part shows correct answers. The shaded level marks sufficient evidence.}
  \label{fig:answer_dist_models}
\end{figure}

\paragraph{The evidence at the answer for each model.}
\Cref{fig:answer_dist_models} repeats \cref{fig:answer_dist} for each of the three models and adds \method (EM). Across every model and benchmark, \method gives fewer answers at low levels of evidence and more answers with sufficient evidence than Base. \Cref{tab:split} reports the split from \cref{eq:split} for \method and PPO (belief + action) on each model, with 95\% intervals from a bootstrap over the test games. For \method, the part from answering with more evidence is positive on every model and benchmark, while the part from accuracy at the same evidence is negative or not different from zero. This second part is most negative in HiddenGames on the two Qwen3-VL models ($-7.9$ and $-12.9$\%), because after training these models are less often correct with sufficient evidence in the Tabletop tasks. For PPO (belief + action), the second part is positive on RoboQuest, at $+2.8$\% over the three models.

\begin{table}[!htbp]
  \centering\footnotesize
  \setlength{\tabcolsep}{3pt}
  \resizebox{\textwidth}{!}{%
  \begin{tabular}{@{}ll rr rr rr@{}}
    \toprule
     & & \multicolumn{2}{c}{Accuracy} & \multicolumn{2}{c}{Gain of \method over Base} & \multicolumn{2}{c}{Gain of PPO (belief + action) over Base} \\
    \cmidrule(lr){3-4}\cmidrule(lr){5-6}\cmidrule(l){7-8}
    Benchmark & Model & Base & \method & more evidence & same evidence & more evidence & same evidence \\
    \midrule
    HiddenGames & All three models & 21.5 & 39.6 & $+26.7$ [$+$24.8, $+$28.5] & $-8.5$ [$-$10.4, $-$6.6] & $+9.4$ [$+$7.9, $+$10.9] & $-1.0$ [$-$2.9, $+$1.0] \\
     & Qwen3.5-4B & 25.1 & 52.2 & $+29.6$ [$+$25.9, $+$33.6] & $-2.5$ [$-$5.4, $+$0.3] & $+19.9$ [$+$16.4, $+$23.8] & $-4.6$ [$-$8.1, $-$1.7] \\
     & Qwen3-VL-4B & 20.5 & 36.3 & $+23.7$ [$+$20.1, $+$27.1] & $-7.9$ [$-$10.9, $-$4.6] & $-0.0$ [$-$1.2, $+$1.5] & $+3.8$ [$+$0.3, $+$7.3] \\
     & Qwen3-VL-8B & 19.0 & 30.4 & $+24.3$ [$+$20.8, $+$27.8] & $-12.9$ [$-$16.6, $-$8.9] & $+8.5$ [$+$6.0, $+$11.1] & $-2.5$ [$-$5.8, $+$0.8] \\
    \midrule
    Pro-Worlds & All three models & 17.2 & 30.2 & $+15.8$ [$+$13.0, $+$18.4] & $-2.8$ [$-$5.7, $+$0.6] & $+4.2$ [$+$3.0, $+$5.5] & $+0.0$ [$-$2.9, $+$3.0] \\
     & Qwen3.5-4B & 22.3 & 32.1 & $+11.1$ [$+$7.4, $+$14.9] & $-1.3$ [$-$6.9, $+$4.9] & $+9.4$ [$+$6.1, $+$13.2] & $-2.4$ [$-$8.5, $+$4.6] \\
     & Qwen3-VL-4B & 17.2 & 28.7 & $+14.3$ [$+$11.0, $+$17.5] & $-2.8$ [$-$7.5, $+$2.3] & $-1.6$ [$-$3.4, $-$0.4] & $+1.1$ [$-$3.1, $+$6.1] \\
     & Qwen3-VL-8B & 14.7 & 29.9 & $+22.6$ [$+$12.1, $+$29.0] & $-7.4$ [$-$14.6, $+$3.4] & $+5.4$ [$+$2.8, $+$8.0] & $-1.8$ [$-$7.4, $+$2.9] \\
    \midrule
    RoboQuest & All three models & 35.8 & 67.9 & $+31.8$ [$+$29.5, $+$34.3] & $+0.3$ [$-$1.3, $+$2.0] & $+17.4$ [$+$15.3, $+$19.7] & $+2.8$ [$+$1.4, $+$4.2] \\
     & Qwen3.5-4B & 42.2 & 61.5 & $+19.4$ [$+$15.8, $+$22.9] & $-0.2$ [$-$2.6, $+$2.1] & $+19.8$ [$+$16.2, $+$23.1] & $+2.9$ [$+$0.0, $+$5.3] \\
     & Qwen3-VL-4B & 31.8 & 70.8 & $+37.7$ [$+$33.6, $+$42.3] & $+1.3$ [$-$1.6, $+$4.0] & $+8.6$ [$+$5.7, $+$12.2] & $+3.6$ [$+$1.4, $+$5.6] \\
     & Qwen3-VL-8B & 38.5 & 71.4 & $+32.8$ [$+$28.9, $+$36.9] & $+0.1$ [$-$2.4, $+$2.7] & $+17.2$ [$+$13.8, $+$20.7] & $+3.3$ [$+$0.8, $+$5.5] \\
    \bottomrule
  \end{tabular}%
  }
  \caption{Accuracy gain over Base, split by \cref{eq:split}. \emph{More evidence}: the part from answering with more evidence. \emph{Same evidence}: the part from accuracy at the same evidence. The two parts sum to the gain, and brackets give 95\% intervals from a bootstrap over the test games. Evidence levels follow \mainref{fig:answer_dist}, and an episode without an answer counts as wrong. At any level where only one of the two methods answers, we use that method's accuracy for both, placing that level in the part from more evidence. HiddenGames: the five tasks with an answer. RoboQuest: the eight tasks whose requirements can be counted.}
  \label{tab:split}
\end{table}

\paragraph{Answers with sufficient evidence.} Answers with sufficient evidence might be correct more often only because they come from easier games. \Cref{tab:within_game} tests this by comparing the two kinds of answers within the same test game. For each model, it uses the games that have both kinds of answers across all methods and test episodes. Within the same game, answers with sufficient evidence are correct 49.5 to 76.1\% more often than answers with insufficient evidence. This also holds for Base's episodes alone and for \method's episodes alone. Most MiniGrid and MineSweeper games are not part of this test because no method reaches sufficient evidence in them. The lower part of \cref{tab:within_game} gives the share of episodes in which an agent gives a correct answer with sufficient evidence. \method's share is 4.5 times Base's share on HiddenGames, 26 times on Pro-Worlds, and 2.1 times on RoboQuest. Compared to Base (w/ belief), its share is 2.3, 7.8, and 1.7 times as large.

\begin{table}[!htbp]
  \centering
  \small
  \setlength{\tabcolsep}{5pt}
  \begin{tabular}{@{}l rrr@{}}
    \toprule
     & HiddenGames & Pro-Worlds & RoboQuest \\
    \midrule
    \multicolumn{4}{@{}l}{\emph{Accuracy within the same game (\%)}} \\
    \quad Sufficient evidence & 71.7 & 72.7 & 89.1 \\
    \quad Insufficient evidence & 12.1 & 23.2 & 13.0 \\
    \quad Difference & $+59.5$ & $+49.5$ & $+76.1$ \\
    \quad 95\% interval & [57.1, 62.0] & [44.8, 54.2] & [73.6, 78.3] \\
    \quad Difference, Base only & $+71.0$ & -- & $+69.7$ \\
    \quad Difference, \method only & $+82.8$ & $+42.0$ & $+80.8$ \\
    \quad Games & 890 of 1,481 & 314 of 900 & 854 of 1,200 \\
    \midrule
    \multicolumn{4}{@{}l}{\emph{Correct answers with sufficient evidence (\% of episodes)}} \\
    \quad Base & 7.5 & 0.3 & 31.7 \\
    \quad Base (w/ belief) & 14.5 & 1.1 & 38.3 \\
    \quad \method & 33.3 & 8.7 & 66.6 \\
    \quad \method / Base & 4.5$\times$ & 26$\times$ & 2.1$\times$ \\
    \quad \method / Base (w/ belief) & 2.3$\times$ & 7.8$\times$ & 1.7$\times$ \\
    \bottomrule
  \end{tabular}
  \caption{\textbf{Answers with sufficient evidence.} Upper part: accuracy of answers given with sufficient evidence and with insufficient evidence within the same test game. \emph{Games}: test games that have both kinds of answers across one model's episodes from Base, PPO (belief + action), \method (EM), and \method. The 95\% interval comes from a bootstrap over these games. \emph{Base only} and \emph{\method only} use one method's episodes. Base has both kinds of answers in only one Pro-Worlds game. Lower part: the share of all episodes (\%) with a correct answer given with sufficient evidence, and \method's share divided by the share for Base or Base (w/ belief). HiddenGames covers the five tasks with an answer, and RoboQuest covers the eight tasks where requirements can be counted. Results are pooled across the three models and the three test seeds.}
  \label{tab:within_game}
\end{table}

\subsection{When the agents commit}

\begin{figure}[!tp]
  \centering
  \includegraphics[width=\linewidth]{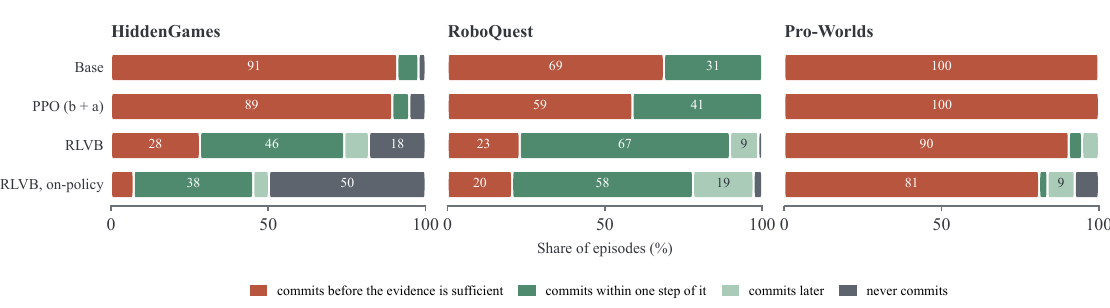}
  \caption{\textbf{When the agent commits} (Qwen3-VL-4B). Each episode is placed according to the turn on which the agent commits, relative to the first turn with sufficient evidence. \emph{Within one step}: the agent commits on that turn or after one further step.}
  \label{fig:when_stop}
\end{figure}

\Cref{fig:when_stop} shows the turn at which the agent commits in each episode, relative to the first turn with sufficient evidence. On Qwen3-VL-4B, \method commits within one step of that turn in 46\% of HiddenGames episodes and 67\% of RoboQuest episodes, compared with 7\% and 31\% for Base.

\paragraph{Steps, useful steps and sufficient evidence for every method.}
\Cref{tab:three_questions} reports, for each model and six methods, the steps per episode, the share of steps that add evidence, the share of episodes reaching sufficient evidence, and the share of turns on which the agent answers. It also includes the on-policy variant of \method from \cref{sec:onpolicy}.

\begin{table}[!htbp]
  \centering
  \footnotesize
  \setlength{\tabcolsep}{3pt}
  \resizebox{\textwidth}{!}{%
  \begin{tabular}{@{}l rrr rrr rrr rrr rr rr@{}}
    \toprule
     & \multicolumn{12}{c}{Steps, useful steps and sufficient evidence} & \multicolumn{4}{c}{Answers at a turn (\%)} \\
    \cmidrule(lr){2-13}\cmidrule(l){14-17}
     & \multicolumn{3}{c}{Tabletop} & \multicolumn{3}{c}{MiniGrid, MineSweeper} & \multicolumn{3}{c}{Pro-Worlds} & \multicolumn{3}{c}{RoboQuest} & \multicolumn{2}{c}{Tabletop} & \multicolumn{2}{c}{RoboQuest} \\
    \cmidrule(lr){2-4}\cmidrule(lr){5-7}\cmidrule(lr){8-10}\cmidrule(lr){11-13}\cmidrule(lr){14-15}\cmidrule(l){16-17}
    Method & Steps & Useful & Suff. & Steps & Useful & Suff. & Steps & Useful & Suff. & Steps & Useful & Suff. & Not suff.$\downarrow$ & Suff.$\uparrow$ & Not suff.$\downarrow$ & Suff.$\uparrow$ \\
    \midrule
    \multicolumn{17}{@{}l}{\textbf{Qwen3.5-4B}} \\[1pt]
    \quad Base & 0.8 & 100 & 19 & 0.9 & 37 & 0 & 12.4 & 6 & 2 & 6.4 & 21 & 45 & 45 & 100 & 8 & 98 \\
    \quad GRPO (action-only) & 1.6 & 98 & 46 & 1.4 & 19 & 0 & 7.9 & 4 & 0 & 6.1 & 23 & 46 & 25 & 93 & 8 & 95 \\
    \quad Base (w/ belief) & 1.9 & 99 & 58 & 1.7 & 33 & 0 & 18.0 & 6 & 4 & 7.0 & 21 & 52 & 19 & 95 & 7 & 72 \\
    \quad PPO (belief + action) & 1.9 & 96 & 62 & 3.8 & 26 & 2 & 24.1 & 5 & 9 & 8.8 & 23 & 67 & 17 & 89 & 4 & 89 \\
    \quad \method (EM) & 2.3 & 94 & 75 & 13.2 & 11 & 2 & 25.4 & 5 & 10 & 7.8 & 23 & 57 & 10 & 84 & 6 & 82 \\
    \quad \method & 2.6 & 92 & 88 & 15.6 & 4 & 0 & 29.3 & 5 & 14 & 9.2 & 21 & 68 & 5 & 81 & 4 & 67 \\
    \midrule
    \multicolumn{17}{@{}l}{\textbf{Qwen3-VL-4B}} \\[1pt]
    \quad Base & 0.3 & 100 & 9 & 0.7 & 46 & 0 & 10.6 & 5 & 0 & 3.6 & 23 & 31 & 77 & 100 & 16 & 96 \\
    \quad GRPO (action-only) & 0.3 & 100 & 8 & 0.5 & 41 & 0 & 8.4 & 5 & 0 & 3.6 & 22 & 27 & 77 & 100 & 17 & 95 \\
    \quad Base (w/ belief) & 0.3 & 99 & 8 & 1.5 & 45 & 0 & 9.8 & 5 & 0 & 4.9 & 24 & 40 & 80 & 95 & 11 & 88 \\
    \quad PPO (belief + action) & 0.4 & 100 & 9 & 1.6 & 39 & 0 & 3.9 & 5 & 0 & 5.4 & 20 & 42 & 69 & 100 & 10 & 97 \\
    \quad \method (EM) & 1.2 & 95 & 34 & 2.8 & 28 & 0 & 17.3 & 5 & 1 & 5.5 & 23 & 46 & 38 & 84 & 9 & 80 \\
    \quad \method & 3.1 & 76 & 89 & 15.1 & 7 & 4 & 31.7 & 4 & 10 & 9.8 & 21 & 76 & 4 & 53 & 2 & 64 \\
    \quad On-policy variant & 3.2 & 75 & 93 & 42.5 & 1 & 0 & 33.3 & 4 & 12 & 10.4 & 21 & 78 & 3 & 47 & 2 & 42 \\
    \midrule
    \multicolumn{17}{@{}l}{\textbf{Qwen3-VL-8B}} \\[1pt]
    \quad Base & 0.1 & 100 & 5 & 0.7 & 58 & 0 & 5.3 & 7 & 0 & 5.5 & 21 & 38 & 87 & 100 & 10 & 94 \\
    \quad GRPO (action-only) & 0.1 & 100 & 3 & 0.8 & 61 & 0 & 7.0 & 6 & 0 & 5.4 & 22 & 37 & 89 & 100 & 10 & 97 \\
    \quad Base (w/ belief) & 0.6 & 98 & 19 & 1.2 & 31 & 0 & 10.6 & 6 & 0 & 5.7 & 23 & 44 & 61 & 93 & 9 & 85 \\
    \quad PPO (belief + action) & 0.9 & 99 & 30 & 1.9 & 34 & 0 & 10.0 & 6 & 1 & 8.1 & 23 & 59 & 46 & 97 & 5 & 94 \\
    \quad \method (EM) & 0.9 & 95 & 28 & 2.3 & 24 & 0 & 16.6 & 5 & 1 & 6.6 & 23 & 53 & 47 & 86 & 7 & 91 \\
    \quad \method & 2.7 & 83 & 76 & 16.7 & 3 & 2 & 33.9 & 4 & 11 & 10.1 & 22 & 77 & 10 & 63 & 2 & 70 \\
    \bottomrule
  \end{tabular}%
  }
  \caption{Work, useful work, and the decision to commit. \emph{Steps}: steps per episode. \emph{Useful}: share of steps that add evidence (\%). \emph{Suff.}: share of episodes reaching sufficient evidence (\%). \emph{Answers at a turn}: share of turns on which the agent answers, reported separately for turns without sufficient evidence and turns with sufficient evidence; a perfect decision yields 0 and 100. RoboQuest covers the eight tasks where evidence can be counted. The on-policy variant from \cref{sec:onpolicy} is \method with new games before every update.}
  \label{tab:three_questions}
\end{table}

\paragraph{Episodes without an answer.} \Cref{tab:no_answer} summarizes the episodes that end without an answer. On the Tabletop tasks, every such episode has sufficient evidence (5 of 5 for \method and 63 of 63 for the on-policy variant). The model finished the work, but it did not answer. In MiniGrid and MineSweeper, almost none of these episodes have sufficient evidence (1 of 83 and 1 of 186). The steps there did not bring the needed evidence. On RoboQuest, forcing an answer at the last turn of these episodes solves 3 of 38 episodes for \method and 21 of 98 for the on-policy variant. Working longer fails in two ways. In MiniGrid and MineSweeper, the steps bring no evidence. In the Tabletop tasks, the model does not act on its evidence. The on-policy variant shows both failures more often than \method.

\begin{table}[!htbp]
  \centering
  \small
  \setlength{\tabcolsep}{4pt}
  \begin{tabular}{@{}l ccc ccc@{}}
    \toprule
     & \multicolumn{3}{c}{\method} & \multicolumn{3}{c}{On-policy variant} \\
    \cmidrule(lr){2-4}\cmidrule(l){5-7}
     & No & Evidence & Evidence not & No & Evidence & Evidence not \\
    Tasks (episodes) & answer & sufficient & sufficient & answer & sufficient & sufficient \\
    \midrule
    HiddenGames (500)                    & 88 & 6 (7\%)   & 82 (93\%) & 249 & 64 (26\%)  & 185 (74\%) \\
    \quad Tabletop (300)                 & 5  & 5 (100\%) & 0 (0\%)   & 63  & 63 (100\%) & 0 (0\%)    \\
    \quad MiniGrid and MineSweeper (200) & 83 & 1 (1\%)   & 82 (99\%) & 186 & 1 (1\%)    & 185 (99\%) \\
    Pro-Worlds (300)                     & 0  & 0         & 0         & 23  & 1 (4\%)    & 22 (96\%)  \\
    RoboQuest (500)                      & 38 & 3 (8\%)   & 35 (92\%) & 98  & 21 (21\%)  & 77 (79\%)  \\
    \quad Puzzle Box (50)                & 10 & 3 (30\%)  & 7 (70\%)  & 39  & 19 (49\%)  & 20 (51\%)  \\
    \quad Stamps (50)                    & 23 & 0 (0\%)   & 23 (100\%) & 48  & 1 (2\%)    & 47 (98\%)  \\
    \bottomrule
  \end{tabular}
  \caption{Episodes ending without an answer (Qwen3-VL-4B). \emph{No answer}: total number of such episodes. \emph{Evidence sufficient}: number and share of these episodes in which the model had sufficient evidence yet still did not answer. \emph{Evidence not sufficient}: number of these episodes that ended before sufficient evidence was reached. HiddenGames covers the five tasks with an answer, and RoboQuest covers all ten tasks. Because evidence cannot be counted in Stamps and Puzzle Box, we treat a RoboQuest episode as having sufficient evidence when a forced answer at its last turn solves the episode, as described in the paragraph on a forced answer.}
  \label{tab:no_answer}
\end{table}

\paragraph{Intervals and commits by belief line.} \Cref{tab:belief_diag_full} reports the 95\% intervals for the differences in \cref{tab:reading}. For all task groups, \cref{tab:commit_full} reports the commit rates by belief line and evidence. Outside the Tabletop tasks, \method also commits less often than Base (w/ belief) when its belief line names a single answer while several remain. \method commits on 6.5 to 55.9\% of these turns, while Base (w/ belief) commits on 61.3 to 74.3\% of them.

\begin{table}[!htbp]
  \centering
  \footnotesize
  \setlength{\tabcolsep}{2.6pt}
  \resizebox{\textwidth}{!}{%
  \begin{tabular}{@{}lll ccccc@{}}
    \toprule
     & & & \multicolumn{2}{c}{Insufficient evidence} & \multicolumn{3}{c}{Sufficient evidence} \\
    \cmidrule(lr){4-5}\cmidrule(l){6-8}
    Tasks & Model & Agent & Several answers & One answer & Several answers & The right answer & A wrong answer \\
    \midrule
    Tabletop & Qwen3.5-4B & Base & 7.9 [6.5, 9.3] & 92.4 [89.2, 95.6] & 72.8 [64.1, 81.7] & 98.6 [97.2, 99.8] & 94.9 [87.2, 100.0] \\
     & & \method & 0.1 [0.0, 0.2] & 77.8 [68.9, 85.6] & 25.0 [18.0, 32.5] & 91.5 [89.3, 93.6] & 72.6 [65.0, 80.6] \\
     & Qwen3-VL-4B & Base & 67.9 [63.2, 72.8] & 97.5 [95.7, 99.1] & 93.9 [86.1, 100.0] & 100.0 [100.0, 100.0] & -- \\
     & & \method & 0.6 [0.3, 1.0] & 72.6 [62.4, 81.5] & 37.3 [34.5, 40.6] & 70.2 [67.2, 73.5] & 59.3 [56.1, 62.5] \\
     & Qwen3-VL-8B & Base & 29.2 [25.1, 33.3] & 99.2 [98.4, 99.8] & 53.5 [42.6, 66.7] & 100.0 [100.0, 100.0] & 100.0 [100.0, 100.0] \\
     & & \method & 0.1 [0.0, 0.2] & 81.4 [74.3, 87.4] & 26.6 [23.8, 29.3] & 89.2 [86.2, 92.4] & 76.0 [72.4, 80.2] \\
    \midrule
    MiniGrid, MineSweeper & Qwen3.5-4B & Base & 15.7 [13.0, 18.5] & 72.1 [66.3, 77.8] & -- & -- & -- \\
     & & \method & 0.1 [0.0, 0.1] & 33.2 [28.8, 38.1] & -- & -- & -- \\
     & Qwen3-VL-4B & Base & 28.9 [24.6, 33.7] & 71.6 [66.0, 78.0] & -- & -- & -- \\
     & & \method & 1.3 [1.1, 1.6] & 6.5 [5.1, 8.3] & -- & -- & 0.0 [0.0, 0.0] \\
     & Qwen3-VL-8B & Base & 21.3 [17.5, 25.1] & 70.3 [63.8, 77.4] & -- & -- & -- \\
     & & \method & 0.1 [0.0, 0.2] & 9.3 [7.1, 12.1] & -- & -- & -- \\
    \midrule
    Pro-Worlds & Qwen3.5-4B & Base & 3.2 [3.0, 3.4] & 66.5 [57.2, 75.6] & -- & 87.1 [77.1, 96.7] & -- \\
     & & \method & 0.9 [0.8, 1.0] & 55.9 [50.0, 62.4] & 1.9 [0.0, 4.6] & 84.1 [77.9, 90.7] & 82.5 [72.1, 94.4] \\
     & Qwen3-VL-4B & Base & 8.5 [8.0, 9.0] & 74.3 [64.5, 84.6] & -- & -- & -- \\
     & & \method & 2.0 [1.9, 2.1] & 25.8 [21.9, 30.2] & 5.2 [2.8, 7.8] & 50.9 [42.9, 61.1] & 31.2 [18.2, 56.0] \\
     & Qwen3-VL-8B & Base & 7.5 [7.1, 7.9] & 61.3 [52.1, 72.1] & -- & -- & -- \\
     & & \method & 2.2 [2.1, 2.3] & 20.1 [15.4, 27.6] & 7.5 [5.0, 10.4] & 38.4 [31.4, 46.7] & 28.6 [17.5, 48.7] \\
    \bottomrule
  \end{tabular}%
  }
  \caption{Share of turns (\%) on which the agent commits, grouped by what its belief line names and whether the evidence is sufficient. Brackets show 95\% bootstrap intervals over the test games, and Base denotes Base (w/ belief). \emph{The right answer}: the belief line names only the identified answer. \emph{A wrong answer}: the belief line names only a different answer. Cells with fewer than 20 such turns are marked --, and turns with an unreadable belief line are excluded.}
  \label{tab:commit_full}
\end{table}

\begin{table}[!htbp]
  \centering
  \footnotesize
  \setlength{\tabcolsep}{2.6pt}
  \begin{tabular}{@{}ll c c@{}}
    \toprule
     & & Exact line (\%) & Single answer (\%) \\
    Tasks & Model & \method $-$ Base & \method $-$ Base \\
    \midrule
    Tabletop & Qwen3.5-4B & $+5.1$ [$+$2.2, $+$8.4] & $-6.4$ [$-$8.6, $-$4.4] \\
     & Qwen3-VL-4B & $+39.4$ [$+$34.7, $+$44.5] & $-23.8$ [$-$28.3, $-$19.6] \\
     & Qwen3-VL-8B & $+29.6$ [$+$24.7, $+$34.9] & $-28.1$ [$-$33.0, $-$23.3] \\
    \midrule
    MiniGrid, MineSweeper & Qwen3.5-4B & $+4.0$ [$-$0.6, $+$8.6] & $-11.2$ [$-$15.7, $-$6.7] \\
     & Qwen3-VL-4B & $+8.3$ [$+$3.1, $+$13.3] & $-7.7$ [$-$14.5, $-$0.8] \\
     & Qwen3-VL-8B & $+10.4$ [$+$6.4, $+$14.3] & $-10.0$ [$-$15.3, $-$4.7] \\
    \midrule
    Pro-Worlds & Qwen3.5-4B & $+1.6$ [$-$0.5, $+$3.6] & $-5.0$ [$-$6.5, $-$3.7] \\
     & Qwen3-VL-4B & $+5.5$ [$+$2.7, $+$8.5] & $-0.4$ [$-$1.6, $+$0.7] \\
     & Qwen3-VL-8B & $+11.5$ [$+$8.3, $+$14.4] & $-4.6$ [$-$6.1, $-$3.0] \\
    \bottomrule
  \end{tabular}
  \caption{Differences in the reading of the belief line between \method and Base (w/ belief), for the values of \cref{tab:reading}. \emph{Exact line}: the belief line names exactly the answers still possible. \emph{Single answer}: the belief line names a single answer while several remain. Both are measured at states reached by Base (w/ belief), \method, and \method with the belief term while several answers remain, as in \cref{tab:reading}, with every state weighted equally. Brackets show 95\% bootstrap intervals over the test games.}
  \label{tab:belief_diag_full}
\end{table}

\Cref{tab:reading_steps_all} compares the belief lines after the same number of steps instead of at the same states. \Cref{tab:line_sep} reports how well the belief line separates turns with sufficient evidence from turns where several answers remain. \Cref{tab:reading_all} evaluates the belief line across all turns. Over all turns, the belief line of \method is more exact in the Tabletop tasks but less exact on Pro-Worlds, because \method spends many turns in later states that Base (w/ belief) rarely reaches.

\begin{table}[!htbp]
  \centering
  \footnotesize
  \setlength{\tabcolsep}{3pt}
  \resizebox{\textwidth}{!}{%
  \begin{tabular}{@{}ll ccccc@{}}
    \toprule
     & & \multicolumn{5}{c}{Steps before the belief line} \\
    \cmidrule(l){3-7}
    Tasks & Model & 0 & 1 & 2 & 3 & 4 or more \\
    \midrule
    \multicolumn{7}{@{}l}{\emph{(a) Evidence leaves several answers: belief lines naming a single answer (\%), Base / \method}} \\
    Tabletop & Qwen3.5-4B & 6.3 / 1.6 & 22.4 / 12.6 & 8.2 / 6.6 & 9.6 / 2.6 & 30.0 / n/a \\
     & Qwen3-VL-4B & 31.0 / 0.2 & 25.0 / 12.4 & 25.0 / 2.3 & n/a / 7.5 & n/a / n/a \\
     & Qwen3-VL-8B & 35.0 / 0.3 & 69.7 / 33.0 & 29.2 / 7.5 & n/a / 4.1 & n/a / n/a \\
    \midrule
    MiniGrid, MineSweeper & Qwen3.5-4B & 45.3 / 29.1 & 43.2 / 44.7 & 29.5 / 48.0 & 26.0 / 42.9 & 31.3 / 10.9 \\
     & Qwen3-VL-4B & 33.1 / 1.0 & 23.0 / 11.4 & 27.2 / 15.3 & 29.0 / 18.5 & 33.4 / 38.4 \\
     & Qwen3-VL-8B & 13.3 / 0.7 & 62.9 / 14.9 & 47.0 / 17.9 & 48.3 / 20.6 & 53.2 / 51.6 \\
    \midrule
    Pro-Worlds & Qwen3.5-4B & 1.1 / 0.2 & 2.0 / 0.3 & 3.0 / 0.3 & 1.7 / 0.6 & 3.1 / 3.2 \\
     & Qwen3-VL-4B & 1.7 / 0.4 & 1.8 / 0.6 & 1.4 / 0.7 & 0.9 / 0.9 & 0.9 / 3.9 \\
     & Qwen3-VL-8B & 2.9 / 0.3 & 2.8 / 0.3 & 3.0 / 0.8 & 2.6 / 0.8 & 1.6 / 2.2 \\
    \midrule
    \multicolumn{7}{@{}l}{\emph{(b) Evidence identifies the answer: single answers that are right (\%), Base / \method\ / + belief}} \\
    Tabletop & Qwen3.5-4B & n/a & 100.0 / 99.1 / 98.9 & 76.4 / 70.4 / 65.2 & 92.7 / 87.2 / 85.7 & 93.2 / 83.4 / 83.7 \\
     & Qwen3-VL-4B & n/a & 100.0 / 93.1 / 97.8 & n/a / 73.1 / 58.0 & n/a / 60.9 / 56.8 & n/a / 39.1 / 34.2 \\
     & Qwen3-VL-8B & n/a & 96.8 / 100.0 / 100.0 & 70.5 / 67.9 / 76.2 & n/a / 47.3 / 49.6 & n/a / 35.6 / 33.0 \\
    \bottomrule
  \end{tabular}%
  }
  \caption{\textbf{The belief line after the same number of steps.} (a) Turns on which the evidence still leaves several answers. (b) Turns with sufficient evidence on which the belief line names a single answer, restricted to the Tabletop tasks because Base (w/ belief) rarely reaches sufficient evidence elsewhere. Before the first step, the evidence is never sufficient. Cells with fewer than 20 belief lines are marked n/a. Base denotes Base (w/ belief), and \emph{+ belief} denotes \method additionally trained with the belief term. Unreadable belief lines are excluded.}
  \label{tab:reading_steps_all}
\end{table}

\begin{table}[!htbp]
  \centering
  \footnotesize
  \setlength{\tabcolsep}{4pt}
  \begin{tabular}{@{}ll rrc@{}}
    \toprule
     & & \multicolumn{2}{c}{Single answer (\%)} & \\
    \cmidrule(lr){3-4}
    Model & Agent & Several remain & Sufficient & Separation \\
    \midrule
    Qwen3.5-4B & Base & 12.2 & 83.3 & 3.58 [3.23, 3.92] \\
     & \method & 6.2 & 79.7 & 4.08 [3.80, 4.35] \\
     & + belief & 4.9 & 69.1 & 3.77 [3.52, 4.06] \\
    \midrule
    Qwen3-VL-4B & Base & 30.0 & 54.2 & 1.01 [0.52, 1.51] \\
     & \method & 5.2 & 54.5 & 3.09 [2.77, 3.44] \\
     & + belief & 12.2 & 75.9 & 3.12 [2.86, 3.41] \\
    \midrule
    Qwen3-VL-8B & Base & 43.4 & 77.6 & 1.51 [1.09, 2.11] \\
     & \method & 12.6 & 60.8 & 2.38 [2.12, 2.64] \\
     & + belief & 12.4 & 41.5 & 1.61 [1.35, 1.86] \\
    \bottomrule
  \end{tabular}
  \caption{\textbf{How well the belief line tells whether the evidence is sufficient}, Tabletop tasks, all turns. \emph{Single answer}: share of belief lines naming a single answer, reported for turns where several answers remain and for turns with sufficient evidence. \emph{Separation}: difference between the two shares on the log-odds scale, with 95\% bootstrap intervals over the test games. Base denotes Base (w/ belief), and \emph{+ belief} denotes \method additionally trained with the belief term. Unreadable belief lines are excluded.}
  \label{tab:line_sep}
\end{table}

\begin{table}[!htbp]
  \centering
  \footnotesize
  \setlength{\tabcolsep}{3pt}
  \resizebox{\textwidth}{!}{%
  \begin{tabular}{@{}ll rrr c rrr c rrr@{}}
    \toprule
     & & \multicolumn{3}{c}{Exact line (\%)} & & \multicolumn{3}{c}{Several $\rightarrow$ single (\%)} & & \multicolumn{3}{c}{Single $\rightarrow$ several (\%)} \\
    \cmidrule(lr){3-5}\cmidrule(lr){7-9}\cmidrule(l){11-13}
    Tasks & Model & Base & \method & + belief && Base & \method & + belief && Base & \method & + belief \\
    \midrule
    Tabletop & Qwen3.5-4B & 74.5 & 76.1 & 75.1 && 12.2 & 6.2 & 4.9 && 16.7 & 20.3 & 30.9 \\
     & Qwen3-VL-4B & 47.9 & 59.8 & 70.4 && 30.0 & 5.2 & 12.2 && 45.8 & 45.5 & 24.1 \\
     & Qwen3-VL-8B & 52.3 & 54.4 & 46.4 && 43.4 & 12.6 & 12.4 && 22.4 & 39.2 & 58.5 \\
    \midrule
    MiniGrid, MineSweeper & Qwen3.5-4B & 22.4 & 51.6 & 42.1 && 35.5 & 15.6 & 7.5 && n/a & 0.0 & 64.5 \\
     & Qwen3-VL-4B & 45.8 & 34.9 & 49.1 && 29.8 & 34.5 & 7.7 && n/a & 27.3 & 77.3 \\
     & Qwen3-VL-8B & 43.8 & 41.8 & 75.2 && 43.6 & 45.1 & 3.6 && n/a & 40.0 & 60.3 \\
    \midrule
    Pro-Worlds & Qwen3.5-4B & 82.4 & 75.0 & 76.4 && 2.9 & 2.8 & 2.6 && 13.3 & 40.9 & 24.6 \\
     & Qwen3-VL-4B & 77.1 & 69.3 & 81.6 && 1.1 & 3.5 & 2.5 && n/a & 58.5 & 5.6 \\
     & Qwen3-VL-8B & 75.7 & 70.8 & 74.6 && 2.0 & 2.0 & 0.8 && n/a & 51.7 & 52.5 \\
    \bottomrule
  \end{tabular}%
  }
  \caption{\textbf{The belief line against the evidence, over all turns}. \emph{Exact line}: share of all turns on which the belief line names exactly the answers still possible, counting unreadable lines as not exact. \emph{Several $\rightarrow$ single}: share of turns with insufficient evidence on which the belief line names a single answer. \emph{Single $\rightarrow$ several}: share of turns with sufficient evidence on which the belief line still names several answers. The two error columns are computed over turns with a readable belief line. Base denotes Base (w/ belief), and \emph{+ belief} denotes \method additionally trained with the belief term in \cref{eq:score}, as in \cref{tab:belief_term}. Cells with fewer than 20 belief lines are marked as n/a.}
  \label{tab:reading_all}
\end{table}

\Cref{tab:belief_errors} examines what the belief line states when agents fail. On Pro-Worlds, most early commits by Base (w/ belief) come with a belief line that still names several answers. On the Tabletop tasks, 38 to 72\% of its early commits occur with an overconfident belief line. In HiddenGames, most early commits by \method likewise occur with an overconfident belief line. On Pro-Worlds, almost all early commits of \method with a belief line that names several answers happen at the limit of 35 moves, in 219 of 244, 536 of 553 and 668 of 671 cases, against at most 4.3\% for Base (w/ belief). Among wrong answers given with sufficient evidence, the belief line excludes the true answer in 71 to 97\% of cases across every agent and model with ten or more such answers, and the answer contradicts a correct belief line in at most one case per model.
\begin{table}[!htbp]
  \centering
  \small
  \setlength{\tabcolsep}{4pt}
  \begin{tabular}{@{}ll cc cc c@{}}
    \toprule
     & & \multicolumn{2}{c}{Early commits with an} & \multicolumn{2}{c}{Wrong with sufficient evidence,} & Answer against \\
     & & \multicolumn{2}{c}{overconfident line (\%)} & \multicolumn{2}{c}{line excludes the true answer} & a correct line \\
    \cmidrule(lr){3-4}\cmidrule(lr){5-6}
    Tasks & Model & Base & \method & Base & \method & \method \\
    \midrule
    Tabletop & Qwen3.5-4B & 62 (372) & 98 (107) & 49 of 60 & 101 of 104 & 1 \\
     & Qwen3-VL-4B & 38 (826) & 86 (95) & 12 of 17 & 298 of 351 & 0 \\
     & Qwen3-VL-8B & 72 (727) & 99.5 (216) & 32 of 34 & 289 of 320 & 1 \\
    \midrule
    MiniGrid, MineSweeper & Qwen3.5-4B & 72 (480) & 99 (312) & -- & -- & 0 \\
     & Qwen3-VL-4B & 51 (462) & 72 (322) & -- & 1 of 2 & 0 \\
     & Qwen3-VL-8B & 72 (455) & 99 (300) & -- & -- & 0 \\
    \midrule
    Pro-Worlds & Qwen3.5-4B & 38 (859) & 63 (666) & 7 of 8 & 33 of 34 & 0 \\
     & Qwen3-VL-4B & 9 (900) & 32 (814) & -- & 15 of 21 & 0 \\
     & Qwen3-VL-8B & 14 (900) & 16 (795) & -- & 24 of 29 & 0 \\
    \bottomrule
  \end{tabular}
  \caption{What the belief line reports when the agents fail. Base denotes Base (w/ belief). \emph{Early commits with an overconfident line}: share of commits with insufficient evidence in which the belief line names a single answer, with the number of these commits in parentheses. \emph{Wrong with sufficient evidence}: wrong answers given with sufficient evidence in which the belief line excludes the true answer. \emph{Answer against a correct line}: wrong answers in which the belief line names only the true answer. --: no such answers.}
  \label{tab:belief_errors}
\end{table}
\paragraph{Following the belief line or the posterior.} At every turn, we compute the posterior teacher's action twice, once from the exact posterior and once from the belief line. In the second case, we read the belief line as the set of answers it names. The two actions differ most at turns where the belief line is wrong about the answer. \Cref{tab:faith} reports the percentage of these turns where the agent follows the belief line and the percentage where it follows the posterior. When the belief line names a single answer under insufficient evidence, Base (w/ belief) follows the belief line and commits on 71.6 to 99.4\% of these turns. \method commits on 9.8 to 81.4\% of these turns and takes the best step under the posterior on 18.6 to 56.5\%. When there is sufficient evidence and the belief line still names several answers, \method follows the belief line and keeps working on 75.6 to 99.0\% of Tabletop turns and 44.1 to 55.2\% of Pro-Worlds turns. Base (w/ belief) does this on 6.2 to 46.5\% of Tabletop turns. \method therefore follows its belief line when the belief line names several answers, while Base (w/ belief) often commits against it. When there is sufficient evidence and its belief line names a single wrong answer, \method commits to this answer on 62.3 to 76.7\% of these Tabletop turns. This turns the misreading into a wrong answer. By contrast, the agents' steps do not closely follow the belief line or the posterior. On Pro-Worlds, before there is sufficient evidence, a step by either agent is a best step under the belief line or under the posterior only 5.6 to 9.5\% more often than a random legal step. There is no difference between the two agents. Finally, we exclude MineSweeper turns where the belief line names a mine set that has already been ruled out. We cannot read such a belief line as a belief.

\begin{table}[!htbp]
  \centering
  \footnotesize
  \setlength{\tabcolsep}{3.5pt}
  \begin{tabular}{@{}ll rr c rr c rr@{}}
    \toprule
     & & \multicolumn{5}{c}{Line commits, posterior steps} & & \multicolumn{2}{c}{Line steps, posterior commits} \\
    \cmidrule(lr){3-7}\cmidrule(l){9-10}
     & & \multicolumn{2}{c}{Follows the line} & & \multicolumn{2}{c}{Best step under the posterior} & & \multicolumn{2}{c}{Follows the line} \\
    \cmidrule(lr){3-4}\cmidrule(lr){6-7}\cmidrule(l){9-10}
    Tasks & Model & Base & \method && Base & \method && Base & \method \\
    \midrule
    Tabletop & Qwen3.5-4B & 93.6 & 80.2 && 5.5 & 19.8 && 23.3 & 75.6 \\
     & Qwen3-VL-4B & 95.7 & 72.6 && 2.5 & 27.4 && 6.2 & 83.6 \\
     & Qwen3-VL-8B & 99.4 & 81.4 && 0.6 & 18.6 && 46.5 & 99.0 \\
    \midrule
    MiniGrid, MineSweeper & Qwen3.5-4B & 97.0 & 51.5 && 0.7 & 23.2 && -- & -- \\
     & Qwen3-VL-4B & 97.3 & 29.3 && 0.0 & 35.0 && -- & -- \\
     & Qwen3-VL-8B & 98.1 & 45.5 && 0.8 & 22.9 && -- & -- \\
    \midrule
    Pro-Worlds & Qwen3.5-4B & 75.5 & 42.5 && 10.6 & 32.9 && -- & 55.2 \\
     & Qwen3-VL-4B & 73.6 & 16.2 && 8.5 & 47.4 && -- & 44.1 \\
     & Qwen3-VL-8B & 71.6 & 9.8 && 11.8 & 56.5 && -- & 47.2 \\
    \bottomrule
  \end{tabular}
  \caption{Actions at turns where the belief line and the exact posterior imply different actions from the posterior teacher (\%). \emph{Line commits, posterior steps}: the belief line names a single answer under insufficient evidence; the columns report the share of these turns on which the agent commits as implied by the line and the share on which it takes the best step under the posterior. \emph{Line steps, posterior commits}: there is sufficient evidence, but the belief line still names several answers; the columns report the share of these turns on which the agent continues working as implied by the line. Base denotes Base (w/ belief). Cells with fewer than 20 such turns are marked --.}
  \label{tab:faith}
\end{table}

\subsection{The value of a step}

\paragraph{Random steps up to the limit.}
To see what the task alone determines, we first run every test scene without a model, taking a random step at each turn until no steps remain. \Cref{tab:explore_limit} shows the result, where an episode ended by a step, as in MineSweeper, counts as wrong from that step onward. In the Tabletop tasks, every step rules out some answers, and three to five steps identify the answer in every scene. On Pro-Worlds, each step is also safe, but evidence accumulates slowly: after all 35 steps, the chance of a correct answer grows only from 13\% to 28\%, with sufficient evidence reached in 2\% of scenes. In MiniGrid, the 64 random steps identify the answer in only 18\% of scenes, while the posterior teacher identifies it in every scene in 8.3 steps; the evidence therefore exists, but only well-chosen steps reach it. In MineSweeper, a step can even end the episode, because opening a cell with a mine loses the game. Finally, on RoboQuest, the steps consist of a limited number of robot actions, some of which cannot be undone.

\begin{table}[!htbp]
  \centering
  \small
  \setlength{\tabcolsep}{5pt}
  \begin{tabular}{@{}lcccccc@{}}
    \toprule
     & & \multicolumn{3}{c}{Random steps, up to the limit} & \multicolumn{2}{c}{Posterior teacher} \\
    \cmidrule(lr){3-5}\cmidrule(l){6-7}
     & Steps & \multicolumn{2}{c}{Chance correct} & Evidence & & Evidence \\
    Tasks & allowed & at start & at limit & sufficient & Steps & sufficient \\
    \midrule
    Tabletop (3 tasks)    & 3--5 & 19 & 100 & 100 & 2.5  & 100 \\
    MiniGrid              & 64   & 33 & 49  & 18  & 8.3  & 100 \\
    MineSweeper           & 32   & 1  & 15  & 6   & 11.0 & 83  \\
    Pro-Worlds (15 tasks) & 35   & 13 & 28  & 2   & 14.7 & 100 \\
    \bottomrule
  \end{tabular}
  \caption{Random steps taken without a model until no further steps are allowed by the task. \emph{Chance correct}: the exact posterior of the top answer (\%). \emph{Evidence sufficient}: share of scenes (\%).}
  \label{tab:explore_limit}
\end{table}

\paragraph{Evidence within the reach of Base.} Base answers early. We test whether the evidence it needed was out of its reach. We replay each Base test episode on Qwen3-VL-4B using Base's steps up to the turn when it answers. From that state, we take random steps without a model until the task allows no more steps. \Cref{tab:base_continue} gives the result. On the Tabletop tasks, sufficient evidence is present in 11\% of episodes when Base answers, and in all of them after the random steps. On Pro-Worlds, the chance of a correct answer grows from 20\% to 31\%. In these tasks, Base could have gathered the evidence in the same episode by not answering yet. In MiniGrid and MineSweeper, random steps reach sufficient evidence in only 18\% and 9\% of episodes. Random steps from the start of the game reach it just as rarely there, as \cref{tab:explore_limit} shows.

\begin{table}[!htbp]
  \centering
  \small
  \setlength{\tabcolsep}{5pt}
  \begin{tabular}{@{}lcccccc@{}}
    \toprule
     & \multicolumn{4}{c}{Base, at the turn of its answer} & \multicolumn{2}{c}{After the random steps} \\
    \cmidrule(lr){2-5}\cmidrule(l){6-7}
     & & & Evidence & Chance & Evidence & Chance \\
    Tasks & Steps & Correct & sufficient & correct & sufficient & correct \\
    \midrule
    Tabletop (3 tasks)    & 0.3  & 22 & 11 & 30 & 100 & 100 \\
    MiniGrid              & 0.4  & 35 & 0  & 33 & 18  & 49  \\
    MineSweeper           & 1.0  & 1  & 0  & 2  & 9   & 19  \\
    Pro-Worlds (15 tasks) & 10.6 & 17 & 0  & 20 & 4   & 31  \\
    \bottomrule
  \end{tabular}
  \caption{\textbf{Base continued with random steps.} Base's test episodes on Qwen3-VL-4B, continued with random steps without a model from the turn when Base answers. \emph{Steps}: Base's steps before its answer. \emph{Correct}: Base's accuracy (\%). \emph{Evidence sufficient}: share of episodes (\%). \emph{Chance correct}: the exact posterior of the top answer (\%).}
  \label{tab:base_continue}
\end{table}

\paragraph{Steps against random steps.} \Cref{tab:matched_random} compares each agent with random steps in the same scene. The random steps stop after the same number of steps that the agent takes in that episode. As a result, the agent and the random steps differ only in which steps they take. On the Tabletop tasks, both agents' steps reach sufficient evidence somewhat more often than random steps. The increase is 5.0 to 16.0\% for \method and 1.5 to 11.5\% for Base (w/ belief). On Pro-Worlds, \method's steps reach sufficient evidence in 9.6 to 14.4\% of episodes, compared to 1.8 to 2.3\% for random steps. The main difference between the two agents is the number of steps. On the Tabletop tasks, \method takes 2.6 to 3.1 steps per episode, while Base (w/ belief) takes 0.3 to 1.9.

\begin{table}[!htbp]
  \centering
  \small
  \setlength{\tabcolsep}{4pt}
  \begin{tabular}{@{}lll rrr@{}}
    \toprule
    Tasks & Model & Agent & Steps & Evidence sufficient & Random, same steps \\
    \midrule
    Tabletop & Qwen3.5-4B & Base (w/ belief) & 1.9 & 58.0 & 46.5 \\
     &  & \method & 2.6 & 88.1 & 72.1 \\
     & Qwen3-VL-4B & Base (w/ belief) & 0.3 & 7.8 & 6.3 \\
     &  & \method & 3.1 & 89.4 & 83.2 \\
     & Qwen3-VL-8B & Base (w/ belief) & 0.6 & 19.1 & 16.0 \\
     &  & \method & 2.7 & 76.0 & 71.0 \\
    \midrule
    Pro-Worlds & Qwen3.5-4B & Base (w/ belief) & 18.0 & 4.2 & 0.6 \\
     &  & \method & 29.3 & 14.4 & 1.8 \\
     & Qwen3-VL-4B & Base (w/ belief) & 9.8 & 0.0 & 0.1 \\
     &  & \method & 31.7 & 9.6 & 2.0 \\
     & Qwen3-VL-8B & Base (w/ belief) & 10.6 & 0.0 & 0.1 \\
     &  & \method & 33.9 & 11.4 & 2.3 \\
    \bottomrule
  \end{tabular}
  \caption{\textbf{Steps against random steps.} Steps per episode and the share of episodes (\%) that end with sufficient evidence. \emph{Random, same steps}: random steps in the same scene that stop after the same number of steps that the agent takes in that episode.}
  \label{tab:matched_random}
\end{table}

\paragraph{The value of a step on all states.}
\Cref{fig:step_value} shows the value of one step across four step windows. \Cref{tab:step_from_start} reports the underlying numbers and adds MiniGrid and MineSweeper. We follow every episode from the start of its game, which is the same state for all methods, and average the value of one step over all games in which a method takes a step in that window. On the Tabletop tasks, Pro-Worlds, and RoboQuest, the value of a step differs little across methods in every window, and where differences appear, \method is neither consistently lower nor consistently higher than Base. In MiniGrid and MineSweeper, the first step of \method is worth 0.14\% compared to 1.24\% for Base, while its steps after the fourth remain at 0.2 to 0.3\%. On the games where every method takes a step in the window, the differences between \method and PPO (belief + action) are not significant on Pro-Worlds or RoboQuest in any window except the first step on RoboQuest, where the difference is 0.6\%. On these games, \method trails Base on RoboQuest by 2.2 and 1.7\% in the last two windows, which contain only the games where Base is still taking steps that late.

Across states reached by both Base (w/ belief) and \method (where a state consists of the game and the number of answers still possible), we also compare the steps taken with an exact belief line. A step taken by \method raises the posterior of the correct answer by 0.9\% more than a step taken by Base (w/ belief) on the Tabletop tasks (95\% interval from $-2.6$ to $+3.8$\%) and by 0.02\% less on Pro-Worlds (from $-0.08$ to $+0.07$\%). Within each agent, a step taken with an exact belief line and a step taken with a misread line at the same state differ by 1.6\% or less on the Tabletop tasks and by 0.17\% or less on Pro-Worlds. We also let the posterior teacher continue playing from each of these steps until it commits, counting how many more steps it needs than it does after its own best step at the same state. With an exact belief line, a step from \method costs the teacher 0.02 fewer steps than one from Base (w/ belief) on Pro-Worlds (95\% interval from $-0.16$ to $+0.13$, with about 0.6 such steps for both agents) and 0.004 fewer in the Tabletop tasks (from $-0.03$ to $+0.02$), where every step is a best step of the teacher. Only in MiniGrid and MineSweeper does a step from \method cost more, by 0.53 steps (from $+0.37$ to $+0.70$). Within each agent, the difference between steps taken with an exact belief line and those with a misread belief line has an interval that includes zero across all three groups of tasks.
\begin{table}[!htbp]
  \centering\small
  
  \begin{tabular}{llcccc}
    \toprule
    Tasks & Method & Step 1 & Steps 2--4 & Steps 5--7 & Steps 8+ \\
    \midrule
    Tabletop & Base & 25.7 & 34.4 & -- & -- \\
     & PPO (belief + action) & 22.7 & 39.4 & -- & -- \\
     & \method & 21.5 & 41.8 & -- & -- \\
    \midrule
    MiniGrid, MineSweeper & Base & 1.24 & 0.29 & -- & -- \\
     & PPO (belief + action) & 1.23 & 1.10 & 1.95 & 2.36 \\
     & \method & 0.14 & 1.18 & 0.28 & 0.22 \\
    \midrule
    Pro-Worlds & Base & 0.85 & 0.81 & 0.90 & 1.31 \\
     & PPO (belief + action) & 0.68 & 0.93 & 0.90 & 1.33 \\
     & \method & 0.79 & 0.75 & 0.81 & 1.03 \\
    \midrule
    RoboQuest & Base & 15.1 & 13.6 & 9.9 & 17.7 \\
     & PPO (belief + action) & 14.8 & 14.0 & 10.0 & 16.6 \\
     & \method & 14.4 & 14.8 & 12.1 & 17.9 \\
    \bottomrule
  \end{tabular}
  \caption{Value of one step when every episode is followed from the start of its test game across all states visited by a method. The value is the increase in the exact posterior of the correct answer in percent, or on RoboQuest the increase in the share of requirements met, counting only steps taken before the evidence is sufficient. Each entry reports the mean over test games in which the method takes a step in that window; a dash marks windows with fewer than 30 such games.}
  \label{tab:step_from_start}
\end{table}

\paragraph{The value of a step from shared states.}
We repeat this comparison starting from states shared by the methods. A shared state is the same test game with the same amount of evidence that Base, PPO (belief + action), and \method all reach (possibly via different preceding steps) and from which each takes a step. We track each agent's subsequent steps across four windows: the first step, steps 2 to 4, steps 5 to 7, and the later steps. The value of a step is how much it raises the exact posterior of the correct answer, or on RoboQuest the share of requirements met. We first evaluate the value of one step in a window across states where every method takes a step in that window. On the first step, all three methods achieve similar value: for Base, PPO (belief + action), and \method, it raises the posterior by 15.6, 14.9, and 14.8\% in HiddenGames and by 1.5, 1.9, and 1.5\% on Pro-Worlds, and meets 19.1, 19.3, and 19.8\% of the requirements on RoboQuest. On RoboQuest, a step by Base and a step by \method also have similar value in later windows, with 15.8 and 16.9\% in steps 2 to 4, 12.8 and 12.1\% in steps 5 to 7, and 17.7 and 16.8\% in the later steps. In HiddenGames, a step by \method in steps 2 to 4 is worth 24.9\% compared to 21.7\% for Base, while on Pro-Worlds a later step by \method is worth 1.0\% compared to 1.3 to 1.5\% for Base. We then measure the total increase since the shared state, where an agent that has committed keeps its value. Even so, this increase is larger for \method than for Base because \method continues taking steps in later windows where Base has already committed. For Base, PPO (belief + action), and \method, the increase reaches 44.6, 59.5, and 70.1\% on RoboQuest, 23.6, 38.6, and 43.7\% in HiddenGames, and 7.6, 12.0, and 22.2\% on Pro-Worlds. In steps 2 to 4, for example, \method takes a step from 83\% of the shared states in HiddenGames and 81\% on RoboQuest, compared to 26\% and 64\% for Base.
\paragraph{Actions against the best action.}
To test whether training alters action selection, we compare each chosen action against the best actions available in that state, as ranked by the posterior teacher. We evaluate on Pro-Worlds and HiddenGames excluding the Tabletop tasks, where all valid actions have equal value under the teacher. Because RoboQuest has no teacher, we instead measure the share of actions that meet one more requirement. A random action is a best action in 33\% of HiddenGames states and 44\% of Pro-Worlds states. Every method selects a best action more often than a random action, exceeding random by 2 to 7\% in HiddenGames and by 8 to 10\% on Pro-Worlds, with all differences significant.

\method and Base select a best action at similar rates. In HiddenGames, the share is 40\% for \method and 37\% for Base, a small difference of 2.5\% (95\% interval from 1.2 to 3.9\%). On Pro-Worlds, the corresponding shares are 53\% and 52\%, with an interval for the difference that includes zero (from $-0.5$ to $+1.5$\%). Across the HiddenGames tasks, the difference runs in both directions: \method exceeds Base in FrozenLake (43\% versus 39\%), falls below Base in MineSweeper (39\% versus 47\%), and matches Base in MiniGrid and Sokoban.

As a second measure, we evaluate the value of the chosen action relative to the best action. Although the share of best moves does not differ on Pro-Worlds, the value of the chosen move is significantly lower for \method by 3.0\% (62.8\% of the best move's value compared to 65.8\% for Base). In MiniGrid, the chosen action falls 0.21 to 0.23 nats below the best action across all methods, while in MineSweeper it falls 0.99 nats below for \method compared to 0.55 for Base and 1.45 for PPO (belief + action). Both measures show only small differences in either direction, so training has little effect on action choice.

\subsection{The check in the harness}

\Cref{tab:harness} reports the accuracy of Base and \method with and without the check described in \mainref{sec:more_looks}. It also reports how often the agent keeps working after the check and how useful its further steps are. The check also helps \method on RoboQuest and raises its accuracy by 1.9 to 7.7\%. On Qwen3.5-4B with the check, \method reaches 61.7\%, while Base reaches 50.9\%. On HiddenGames and Pro-Worlds, the check changes the accuracy of \method by at most 2.8\%. On the Tabletop tasks, Base keeps working after the check in 17\%, 2\% and 3\% of the checked episodes on Qwen3.5-4B, Qwen3-VL-4B and Qwen3-VL-8B.

\begin{table}[!htbp]
  \centering
  \footnotesize
  \setlength{\tabcolsep}{3.5pt}
  \resizebox{\textwidth}{!}{%
  \begin{tabular}{@{}l rrrr c rrrr c rrrr@{}}
    \toprule
     & \multicolumn{4}{c}{HiddenGames} & & \multicolumn{4}{c}{Pro-Worlds} & & \multicolumn{4}{c}{RoboQuest} \\
    \cmidrule(lr){2-5}\cmidrule(lr){7-10}\cmidrule(l){12-15}
    Method & Acc. & Cont. & Steps & Useful && Acc. & Cont. & Steps & Useful && Acc. & Cont. & Steps & Useful \\
    \midrule
    \multicolumn{15}{@{}l}{\textbf{Qwen3.5-4B}} \\
    \quad Base & 26.5 &  &  &  && 20.2 &  &  &  && 39.6 &  &  &  \\
    \quad Base, with the check & 27.4 & 22 & 1.3 & 57 && 22.1 & 83 & 5.1 & 5 && 50.9 & 44 & 2.9 & 28 \\
    \quad \method & 48.5 &  &  &  && 32.1 &  &  &  && 54.0 &  &  &  \\
    \quad \method, with the check & 49.7 & 21 & 2.4 & 14 && 31.2 & 38 & 6.7 & 3 && 61.7 & 27 & 2.8 & 24 \\
    \midrule
    \multicolumn{15}{@{}l}{\textbf{Qwen3-VL-4B}} \\
    \quad Base & 24.6 &  &  &  && 17.2 &  &  &  && 28.4 &  &  &  \\
    \quad Base, with the check & 26.2 & 10 & 1.1 & 48 && 16.7 & 13 & 5.6 & 4 && 34.5 & 40 & 1.7 & 30 \\
    \quad \method & 36.2 &  &  &  && 28.7 &  &  &  && 61.2 &  &  &  \\
    \quad \method, with the check & 37.2 & 18 & 6.2 & 5 && 25.9 & 15 & 6.3 & 2 && 63.1 & 8 & 2.2 & 25 \\
    \midrule
    \multicolumn{15}{@{}l}{\textbf{Qwen3-VL-8B}} \\
    \quad Base & 22.9 &  &  &  && 14.8 &  &  &  && 36.2 &  &  &  \\
    \quad Base, with the check & 24.5 & 10 & 1.2 & 35 && 15.0 & 37 & 2.9 & 7 && 42.6 & 36 & 2.1 & 28 \\
    \quad \method & 31.1 &  &  &  && 29.9 &  &  &  && 62.7 &  &  &  \\
    \quad \method, with the check & 30.7 & 15 & 3.3 & 8 && 31.2 & 10 & 4.8 & 2 && 64.8 & 8 & 2.5 & 25 \\
    \bottomrule
  \end{tabular}%
  }
  \caption{\textbf{A check in the harness.} \emph{With the check}: the harness does not accept an episode's first answer and asks the agent once to check whether the evidence identifies the answer. \emph{Acc.}: accuracy (\%). \emph{Cont.}: share of checked episodes where the agent takes at least one more step (\%). A checked episode is one whose first answer was not accepted. On HiddenGames, only the five tasks with an answer have a first answer to check. \emph{Steps}: mean number of steps taken after the check in these episodes. \emph{Useful}: share of these steps that add evidence (\%).}
  \label{tab:harness}
\end{table}

\subsection{The on-policy variant}

\paragraph{What the on-policy variant knows when it does not answer.}
The exact posterior shows two cases among the episodes that the on-policy variant ends without an answer. In Tabletop Blocks, it gives no answer in 55 of 100 episodes, even though the evidence is sufficient in all of them, with a single answer remaining for four turns on average. Its belief line still names several blocks in 39 of these episodes and a wrong block in 16. In these episodes, the model's reasoning is cut off at the limit of 8,192 tokens in 91\% of turns with sufficient evidence, after which the model takes further steps instead of answering. Across the Tabletop tasks, it answers in 69\% of turns with sufficient evidence when its reasoning finishes within the limit (58\% for \method), but in only 22\% when its reasoning is cut off (43\% for \method). In MiniGrid, by contrast, it gives no answer in 90 of 100 episodes, where the evidence is not sufficient because three answers remain possible in 89 of them after all 64 turns. Here the model is right that it does not know the answer, yet it still does not answer when no step is left.

\paragraph{A forced answer at the last turn.}
For RoboQuest episodes that end without an answer, we test a simple rule: once the agent has used all its steps, we begin its reply as a submit so that the model only writes the answer. With this rule, the on-policy variant solves 21 of its 98 episodes without an answer, raising its accuracy on RoboQuest from 58.8\% to 63.0\%. The rule has almost no effect on Base, which ends at most 1 of 500 episodes without an answer.

\paragraph{Do the further steps of the on-policy variant bring evidence?}
The on-policy variant ends many more episodes without an answer than \method. \Cref{tab:stall} tests whether its final turns in these episodes still add evidence, counting every turn (including rejected actions) as adding evidence whenever the exact posterior changes or, on RoboQuest, a new requirement is met. In MiniGrid, none of the last ten turns adds evidence in any of its 90 episodes without an answer; 89 of these episodes end after 64 turns at the starting posterior of one third. In MineSweeper, the last ten turns add no evidence in 94 of its 96 such episodes, compared to 44 of 81 for \method, whose last ten steps raise the posterior by 3.5\% on average. On the Tabletop tasks, every episode of the on-policy variant without an answer already holds sufficient evidence (63 of 63) for 3.9 turns on average, and 152 of these 247 turns are rejected actions. On Pro-Worlds, both agents use all 35 moves in most episodes (86\% and 69\%), with 3.2\% and 3.5\% of their moves after the tenth still adding evidence, close to 4.2\% for Base. Although \method ends no Pro-Worlds episode without an answer, ten turns without evidence are common there for every agent: the last ten turns before the answer add no evidence in 65\% of episodes for \method, 68\% of answered episodes for the on-policy variant, and 57\% of those for Base (w/ belief). On RoboQuest, the two agents reach sufficient evidence about equally often (78\% and 76\%) and end only 2.8\% and 1.2\% of episodes without an answer.

\begin{table}[!htbp]
  \centering
  \small
  \setlength{\tabcolsep}{4pt}
  \begin{tabular}{@{}ll cccccc@{}}
    \toprule
     & & & Evidence & No & No answer, & No answer, last ten & Later steps \\
    Tasks & Agent & Steps & sufficient & answer & sufficient & turns add nothing & add evidence \\
    \midrule
    Tabletop    & \method & 3.1 & 89 & 1.7 & 5 of 5 & -- & -- \\
                & On-policy & 3.2 & 93 & 21.0 & 63 of 63 & -- & -- \\
    MiniGrid    & \method & 9.4 & 4 & 2.0 & 0 of 2 & 2 of 2 & 0.8 \\
                & On-policy & 60.0 & 1 & 90.0 & 1 of 90 & 90 of 90 & 0.0 \\
    MineSweeper & \method & 21.1 & 3 & 81.0 & 1 of 81 & 44 of 81 & 5.5 \\
                & On-policy & 24.9 & 0 & 96.0 & 0 of 96 & 94 of 96 & 0.9 \\
    Pro-Worlds  & \method & 31.7 & 10 & 0.0 & -- & -- & 3.5 \\
                & On-policy & 33.3 & 12 & 7.7 & 1 of 23 & 19 of 23 & 3.2 \\
    RoboQuest   & \method & 9.8 & 76 & 1.2 & 2 of 5 & 2 of 5 & 11.9 \\
                & On-policy & 10.4 & 78 & 2.8 & 6 of 11 & 4 of 11 & 12.8 \\
    \bottomrule
  \end{tabular}
  \caption{Whether further steps of the on-policy variant bring evidence (Qwen3-VL-4B). \emph{Evidence sufficient}: share of episodes (\%) reaching sufficient evidence. \emph{No answer}: share of episodes (\%) ending without an answer. \emph{No answer, sufficient}: number of these episodes holding sufficient evidence at the end. \emph{No answer, last ten turns add nothing}: number of these episodes adding no evidence in their last ten turns. A turn adds evidence when the exact posterior changes or, on RoboQuest, when it meets a new requirement. Tabletop episodes are shorter than ten turns, and every Tabletop episode without an answer holds sufficient evidence; \method ends no Pro-Worlds episode without an answer. \emph{Later steps add evidence}: share (\%) of steps after the tenth that raise the posterior of the correct answer, change the posterior on Pro-Worlds, or meet a new requirement on RoboQuest, with a dash marking fewer than 0.3 such steps per episode. RoboQuest covers the eight tasks whose requirements can be counted.}
  \label{tab:stall}
\end{table}

\subsection{Examples}

\begin{figure}[!htbp]
  \centering
  \includegraphics[width=\linewidth]{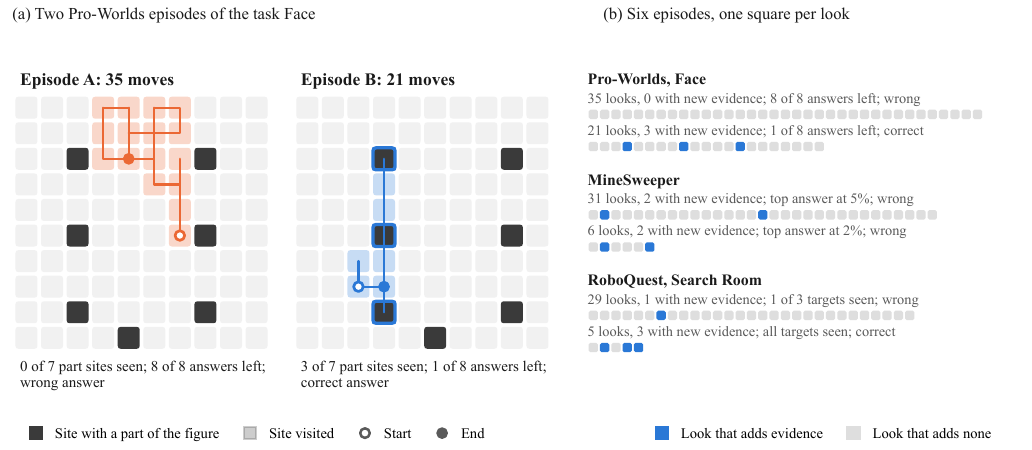}
  \caption{\textbf{Example episodes} of an agent trained with \method on Qwen3-VL-4B. (a) Two Pro-Worlds episodes of the same task, where the agent views one site at a time and only the dark sites show a part of the hidden picture. (b) Two episodes from each benchmark, with the first taking many steps and the second taking few; each square represents one step and is filled when that step adds evidence.}
  \label{fig:more_looks_examples}
\end{figure}

\Cref{fig:more_looks_examples} shows episodes in which many steps yield little evidence. In episode A on Pro-Worlds, the agent takes all 35 steps in one corner of the board without seeing any site that reveals part of the picture, leaving all eight answers possible at the end. In episode B of the same task, it takes 21 steps, sees three such sites, and is left with a single answer. In the first Search Room episode on RoboQuest, the agent takes 29 steps and sees only one of the three objects; thirteen of these steps are not valid actions of the task. It then submits places for the other two objects without having seen them.

\section{Sampling in phases against on-policy sampling}
\label{sec:onpolicy}

\begin{figure}[!htbp]
  \centering
  \includegraphics[width=\linewidth]{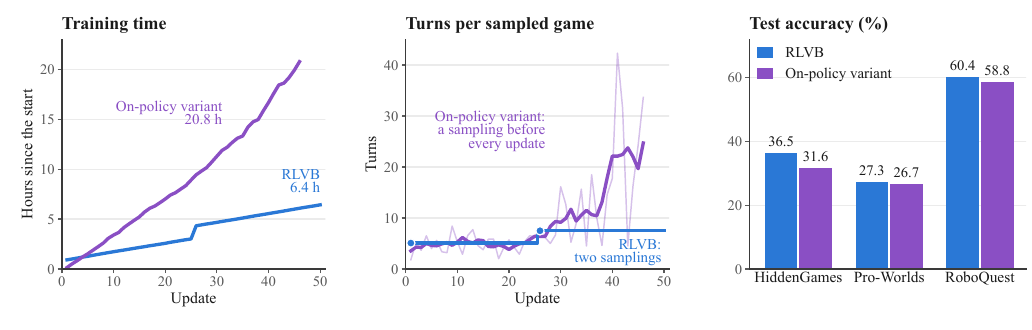}
  \caption{\textbf{Sampling in phases against on-policy sampling} (Qwen3-VL-4B). Left: training time, where the two jumps for \method come from its two sampling passes. Middle: the number of turns in the games sampled for training, with the thin line showing each update and the thick line showing the mean of five updates. Right: accuracy on the three benchmarks.}
  \label{fig:training_dynamics}
\end{figure}

As described in \mainref{sec:method}, \method samples its training games only at the start of a phase and reuses them for all updates in that phase, while on-policy methods such as PPO play new games before every update. Because on-policy sampling is the more common choice, we also train an on-policy variant of \method on Qwen3-VL-4B. The only difference is that this variant samples a small set of new games before every update. \Cref{fig:training_dynamics} compares the two approaches.

The first difference is cost: \method takes 6.4 hours in total, with its two sampling passes taking 47 and 74 minutes and each of its 50 updates then taking about 5 minutes. We stopped the on-policy variant after 46 updates, which took 20.8 hours. Because every update of the variant includes sampling, updates grow slower as training proceeds, averaging 20 minutes over the first ten updates and 39 minutes over the last ten. As the middle panel of \cref{fig:training_dynamics} shows, the games sampled by the on-policy variant lengthen from about 5 turns in the first ten updates to about 20 in the last ten because the model answers less and less often. In contrast, the games sampled by \method have 5 turns in the first phase and 7.5 in the second.

The second difference is what the model learns. The on-policy variant's longer training does not pay off. On the same test episodes, the variant falls behind \method on all three benchmarks (\cref{fig:training_dynamics}). The gap is 4.9\% on HiddenGames, 0.6\% on Pro-Worlds, and 1.6\% on RoboQuest. The variant reaches sufficient evidence as often as \method, but it then often gives no answer (\cref{tab:three_questions}). Its extra steps in MiniGrid and MineSweeper bring almost no evidence (\cref{tab:stall}). A likely reason is the reward. It gives a very low value to an early answer, and it charges very little for one more step after the evidence is sufficient. Under on-policy sampling, each update draws its games from a model that already answers slightly less often than the previous one. The middle panel of \cref{fig:training_dynamics} is consistent with this drift toward fewer answers. With phases, in contrast, the model playing the games is renewed only once during training. The on-policy variant's reasoning is also cut off more often at the 8,192-token limit. After a cutoff, the variant mostly takes more steps instead of answering. We keep the phases because they take a third of the time and produce a model that ends fewer episodes without an answer. \Cref{app:rules} describes three more rules tested against steps that bring no evidence.

In the test episodes, the on-policy variant runs for too long. On Qwen3-VL-4B, it takes 18.9 steps per episode in the five HiddenGames tasks with an answer, compared to 7.9 for \method. It ends half of these episodes without an answer, compared to 18\% for \method (\cref{tab:no_answer}). It commits before its evidence is sufficient in only 7\% of these episodes, compared to 28\%. Unlike \method, it often keeps working after its belief line names a single answer (\cref{sec:more_looks}). \Cref{tab:stall} shows what its extra steps add, and \cref{app:working} describes what it knows when it does not answer.

\section{A larger thinking budget}
\label{app:think16k}

Every run caps the reasoning at 8,192 tokens. To see whether this cap makes the base models answer early, we also test Base with a cap of 16,384 tokens. \Cref{tab:think16k} shows that the larger budget does not help. The accuracy changes by $-0.4$ to 2.6\%, and none of these differences is significant. The base models therefore do not answer too early because their reasoning is cut.

\begin{table}[!htbp]
  \centering
  \footnotesize
  \begin{tabular}{@{}l r rrr@{}}
    \toprule
    Model & Thinking budget & HiddenGames & Pro-Worlds & RoboQuest \\
    \midrule
    Qwen3-VL-8B & 8,192 & 24.6 & 14.3 & 34.8 \\
                & 16,384 & 24.2 & 14.7 & 35.2 \\
    \midrule
    Qwen3-VL-4B & 8,192 & 24.2 & 18.0 & 28.2 \\
                & 16,384 & 24.6 & 19.3 & 28.0 \\
    \midrule
    Qwen3.5-4B  & 8,192 & 27.1 & 22.3 & 39.0 \\
                & 16,384 & 28.4 & 22.7 & 41.6 \\
    \bottomrule
  \end{tabular}
  \caption{Accuracy (\%) of Base with a thinking budget of 8,192 and of 16,384 tokens.}
  \label{tab:think16k}
\end{table}

\section{Two case studies}
\label{app:cases}

\begin{figure*}[p]
  \centering
  \includegraphics[width=\textwidth,height=0.88\textheight,keepaspectratio]{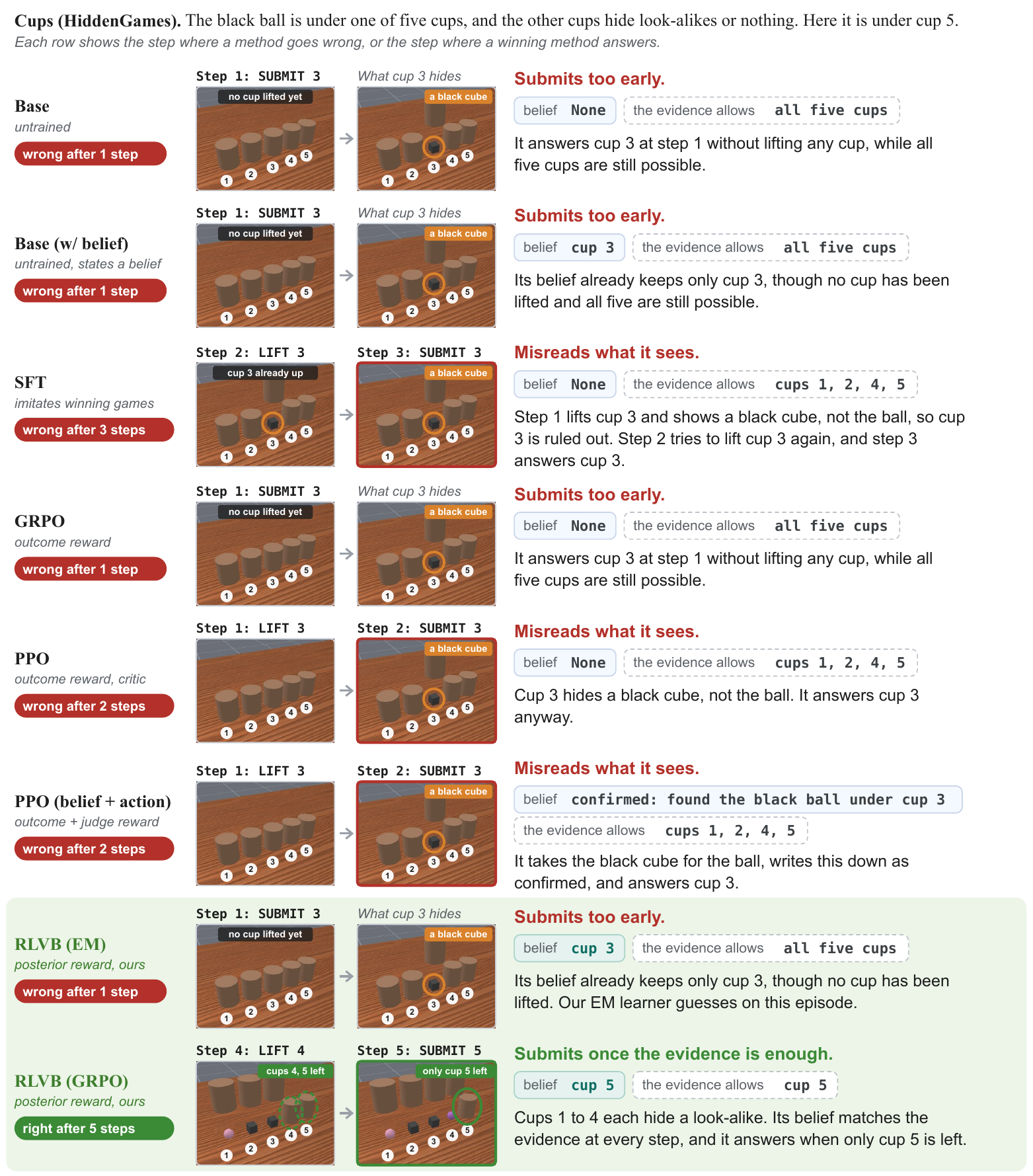}
  \caption{\textbf{Where each method goes wrong on a test episode of Cups.} Each row shows one method of \cref{tab:main} at the step where it goes wrong, or at its answer when it wins. The two frames show the step before and that step. For an answer given at step 1, the second frame shows what the answer missed. Green marks a right answer supported by the evidence, and amber a right answer given before the evidence settles it. The frames are renders of the recorded test runs of Qwen3-VL-4B.}
  \label{fig:case-cups}
\end{figure*}

\begin{figure*}[p]
  \centering
  \includegraphics[width=\textwidth,height=0.88\textheight,keepaspectratio]{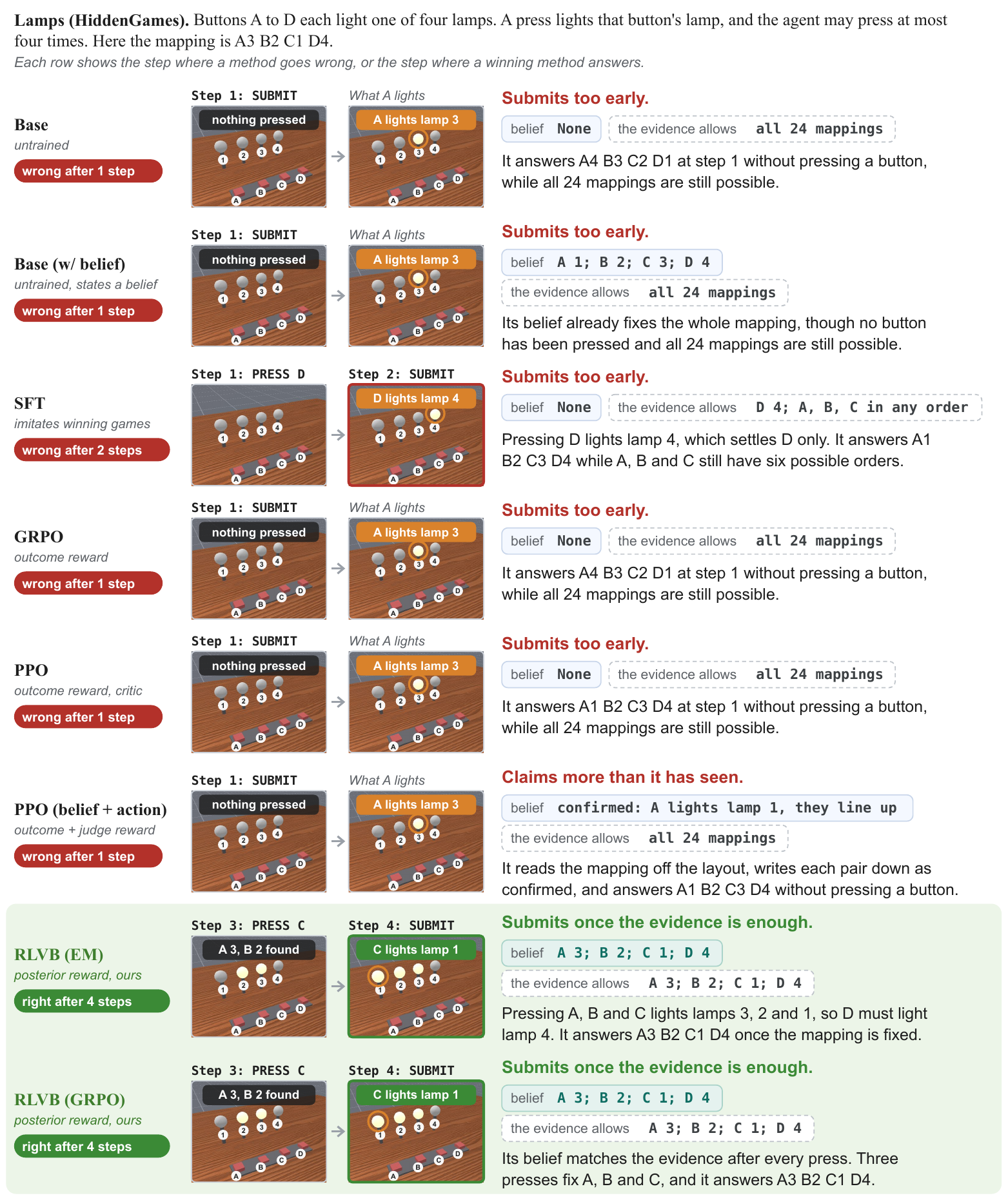}
  \caption{\textbf{Where each method goes wrong on a test episode of Lamps.}}
  \label{fig:case-lamps}
\end{figure*}

\begin{figure*}[p]
  \centering
  \includegraphics[width=\textwidth,height=0.88\textheight,keepaspectratio]{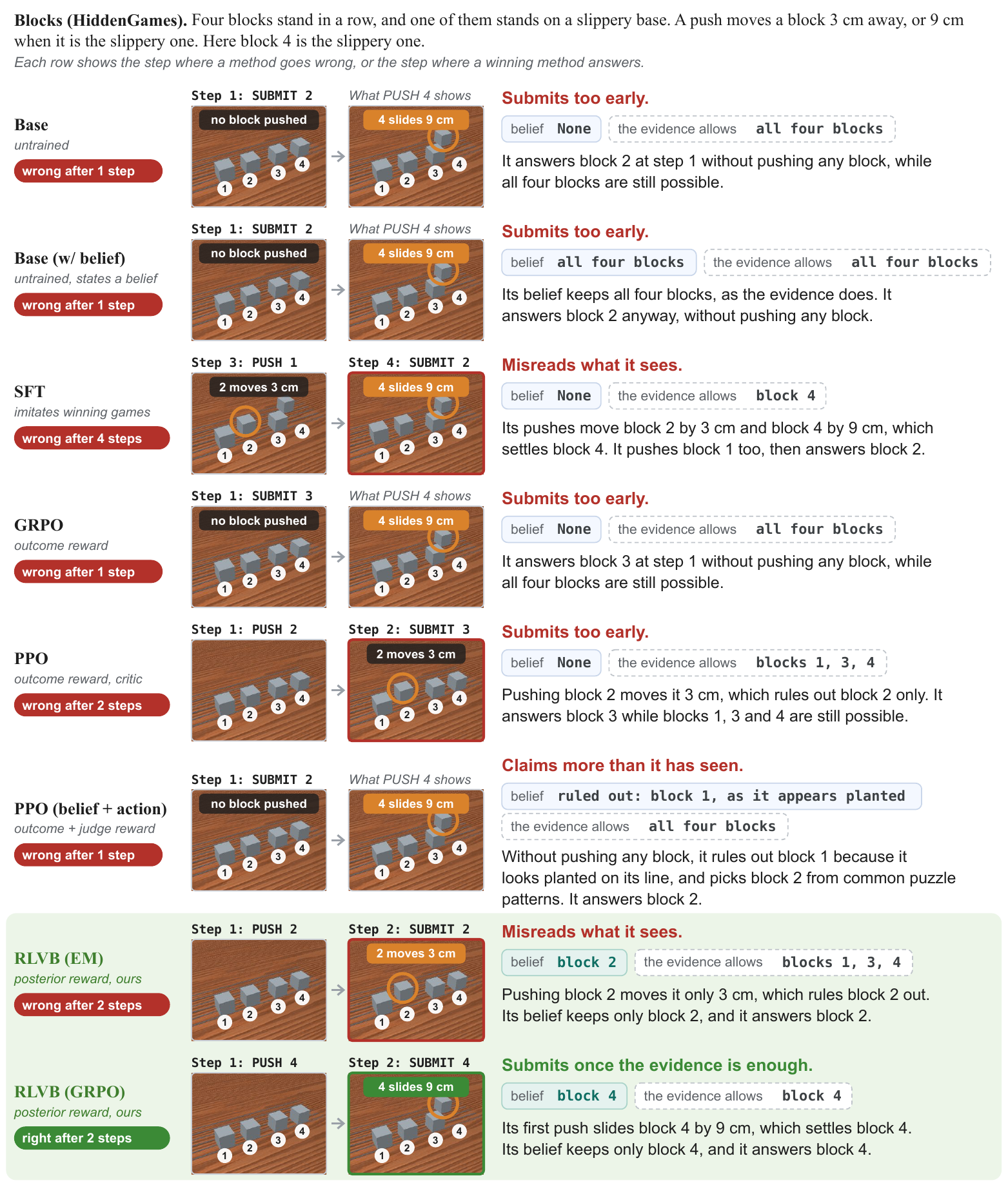}
  \caption{\textbf{Where each method goes wrong on a test episode of Blocks.}}
  \label{fig:case-blocks}
\end{figure*}

\begin{figure*}[p]
  \centering
  \includegraphics[width=\textwidth,height=0.88\textheight,keepaspectratio]{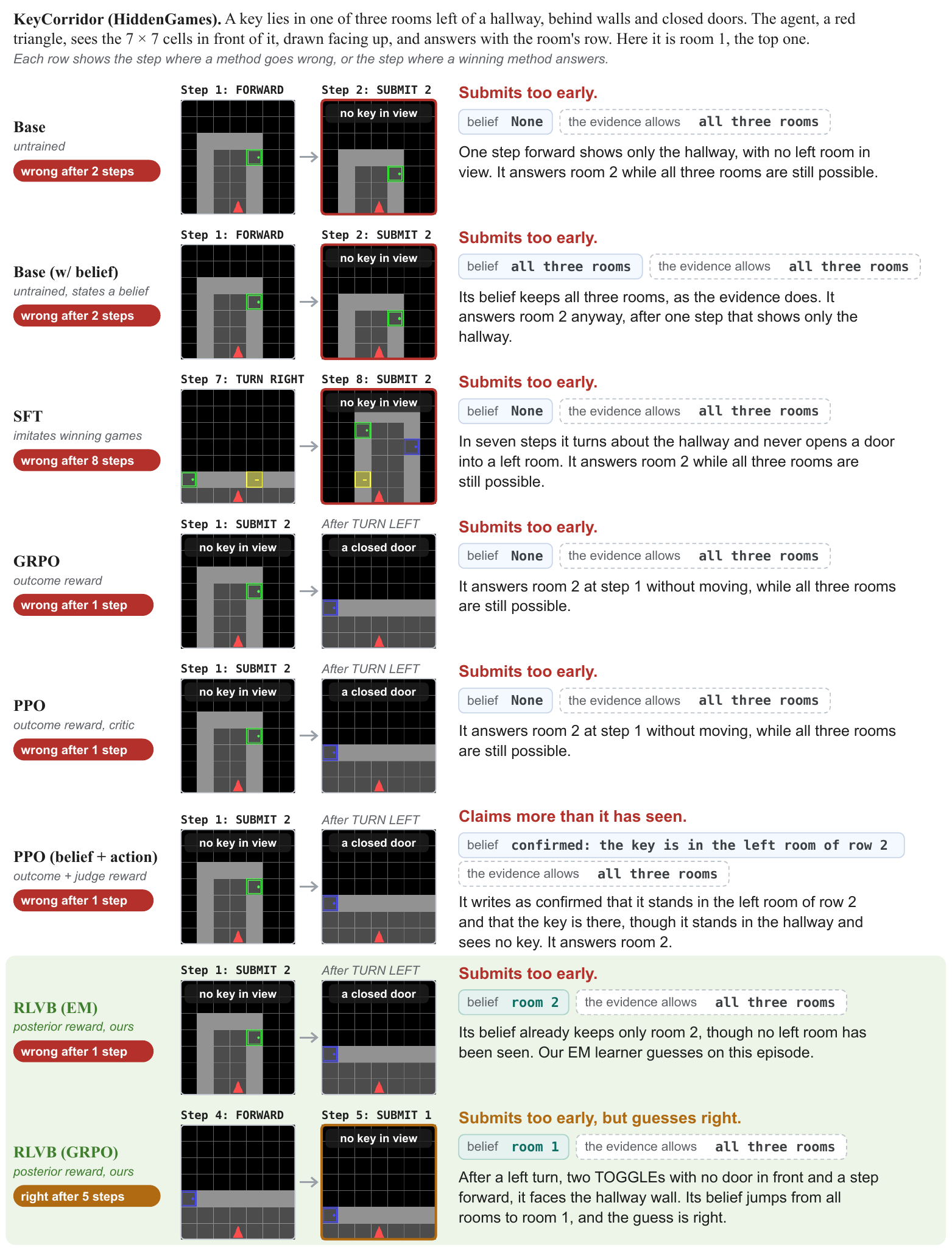}
  \caption{\textbf{Where each method goes wrong on a test episode of KeyCorridor.}}
  \label{fig:case-keycorridor}
\end{figure*}

\begin{figure*}[p]
  \centering
  \includegraphics[width=\textwidth,height=0.88\textheight,keepaspectratio]{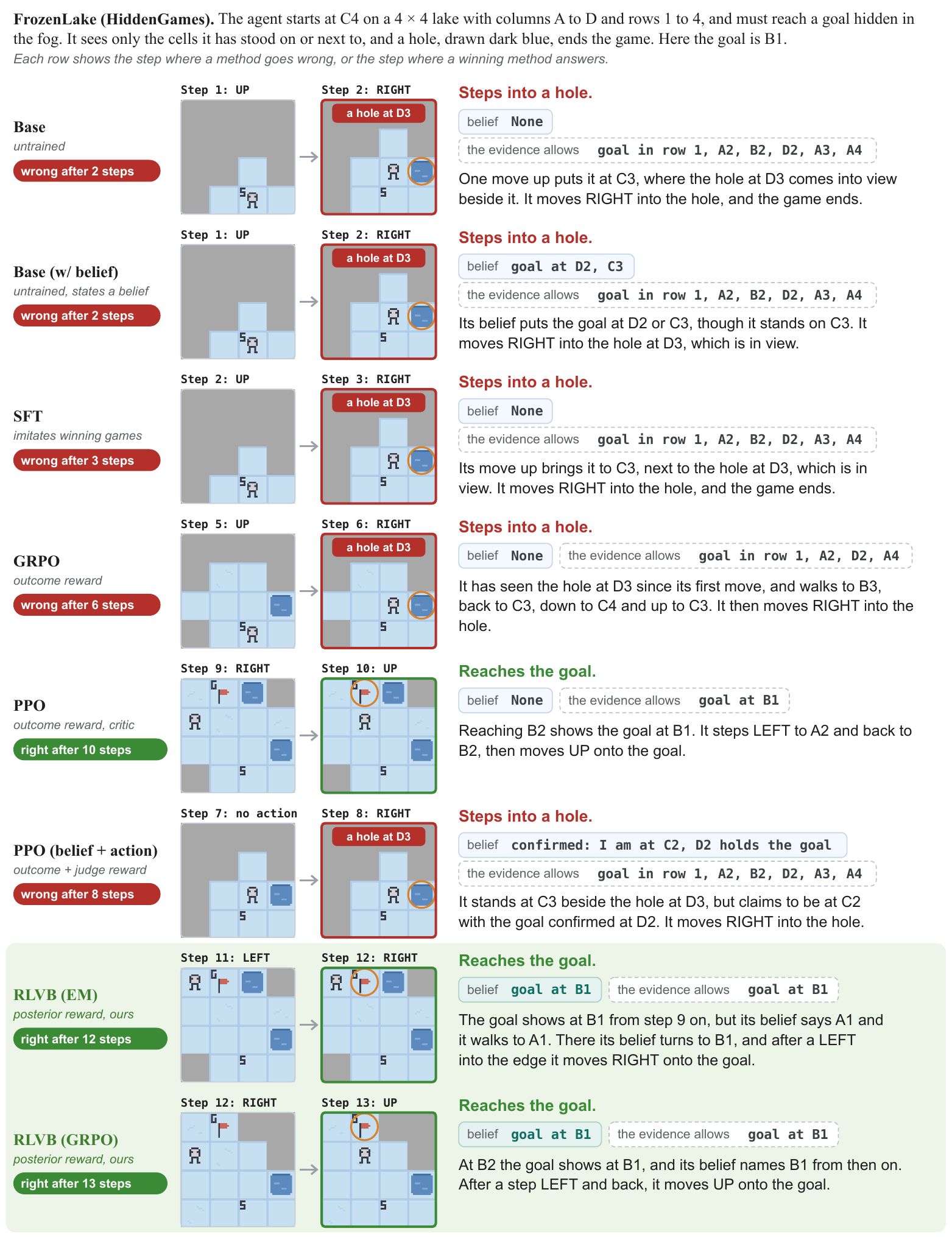}
  \caption{\textbf{Where each method goes wrong on a test episode of FrozenLake.}}
  \label{fig:case-hidden-goal}
\end{figure*}

\begin{figure*}[p]
  \centering
  \includegraphics[width=\textwidth,height=0.88\textheight,keepaspectratio]{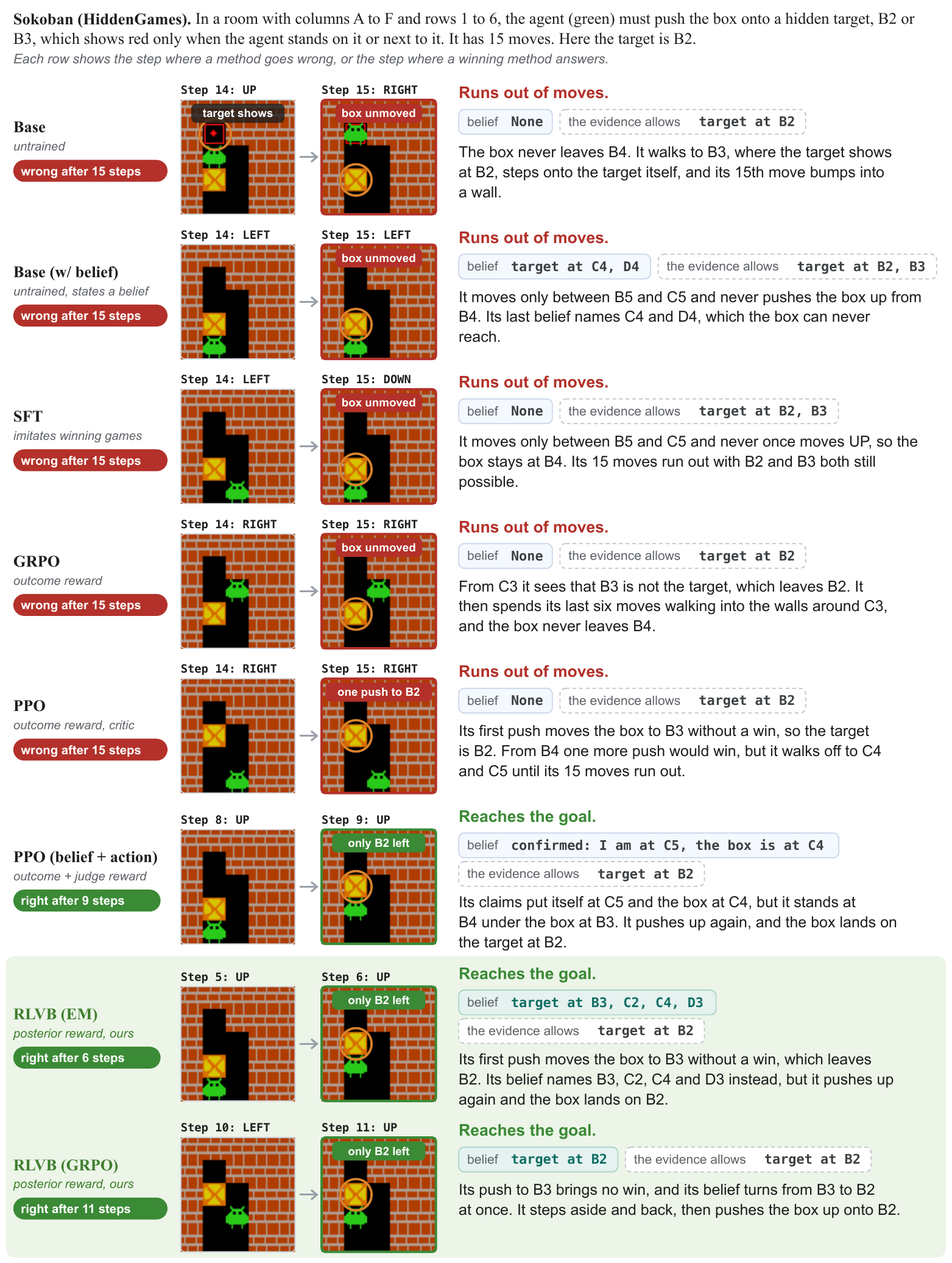}
  \caption{\textbf{Where each method goes wrong on a test episode of Sokoban.}}
  \label{fig:case-hidden-target}
\end{figure*}

\begin{figure*}[p]
  \centering
  \includegraphics[width=\textwidth,height=0.88\textheight,keepaspectratio]{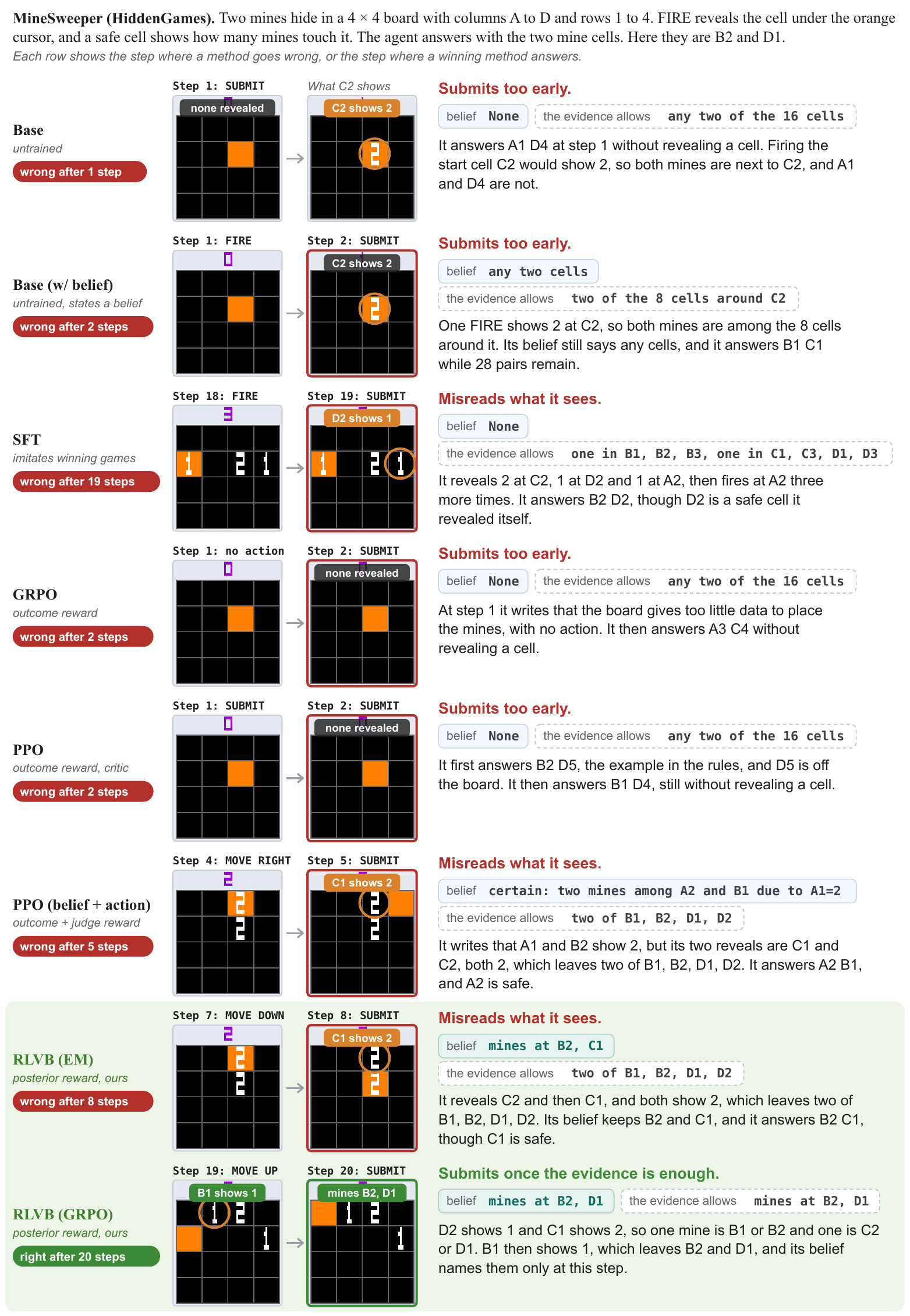}
  \caption{\textbf{Where each method goes wrong on a test episode of MineSweeper.}}
  \label{fig:case-easy}
\end{figure*}

\begin{figure*}[p]
  \centering
  \includegraphics[width=\textwidth,height=0.88\textheight,keepaspectratio]{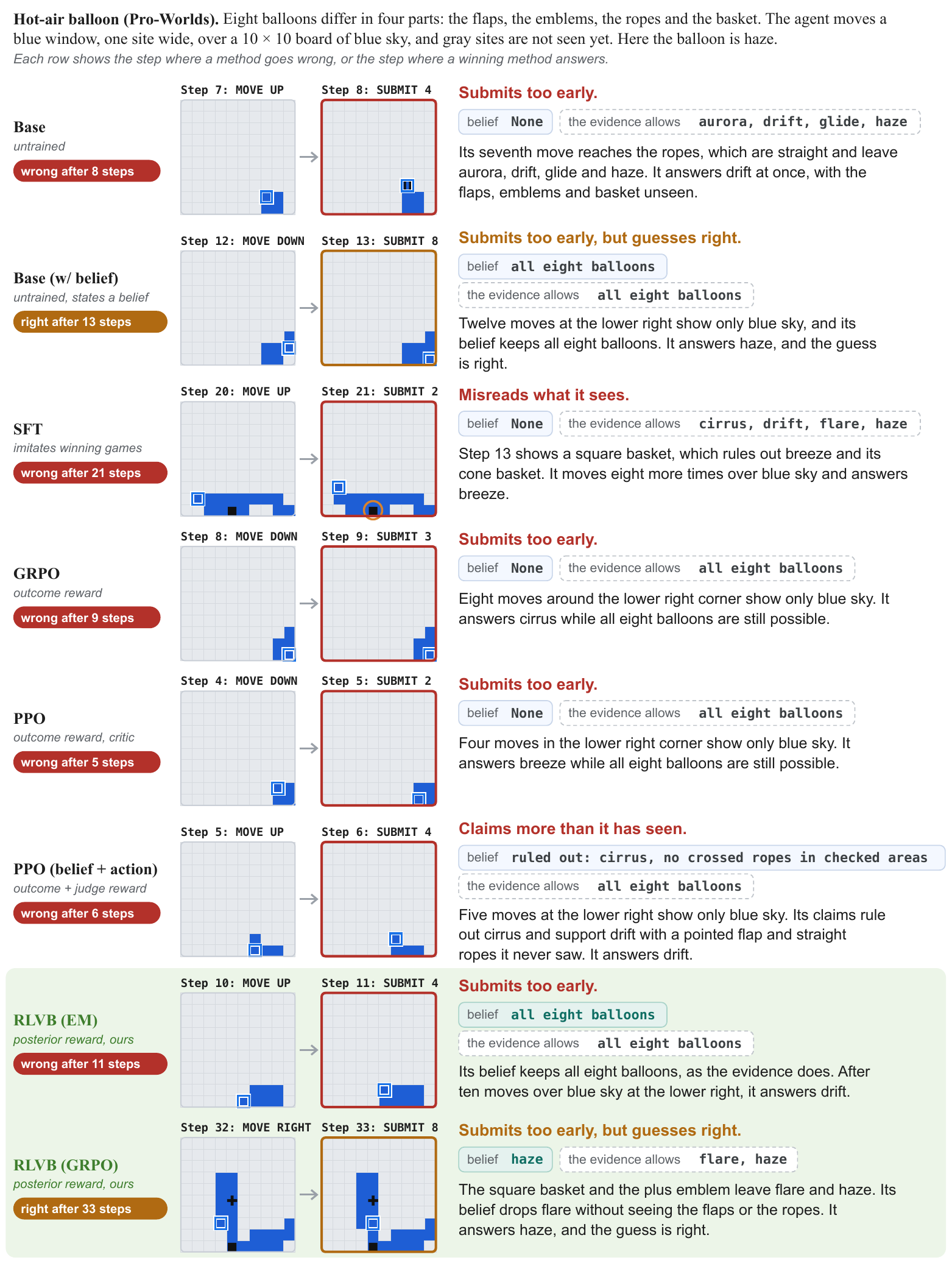}
  \caption{\textbf{Where each method goes wrong on a test episode of Balloon.}}
  \label{fig:case-expansion-balloon}
\end{figure*}

\begin{figure*}[p]
  \centering
  \includegraphics[width=\textwidth,height=0.88\textheight,keepaspectratio]{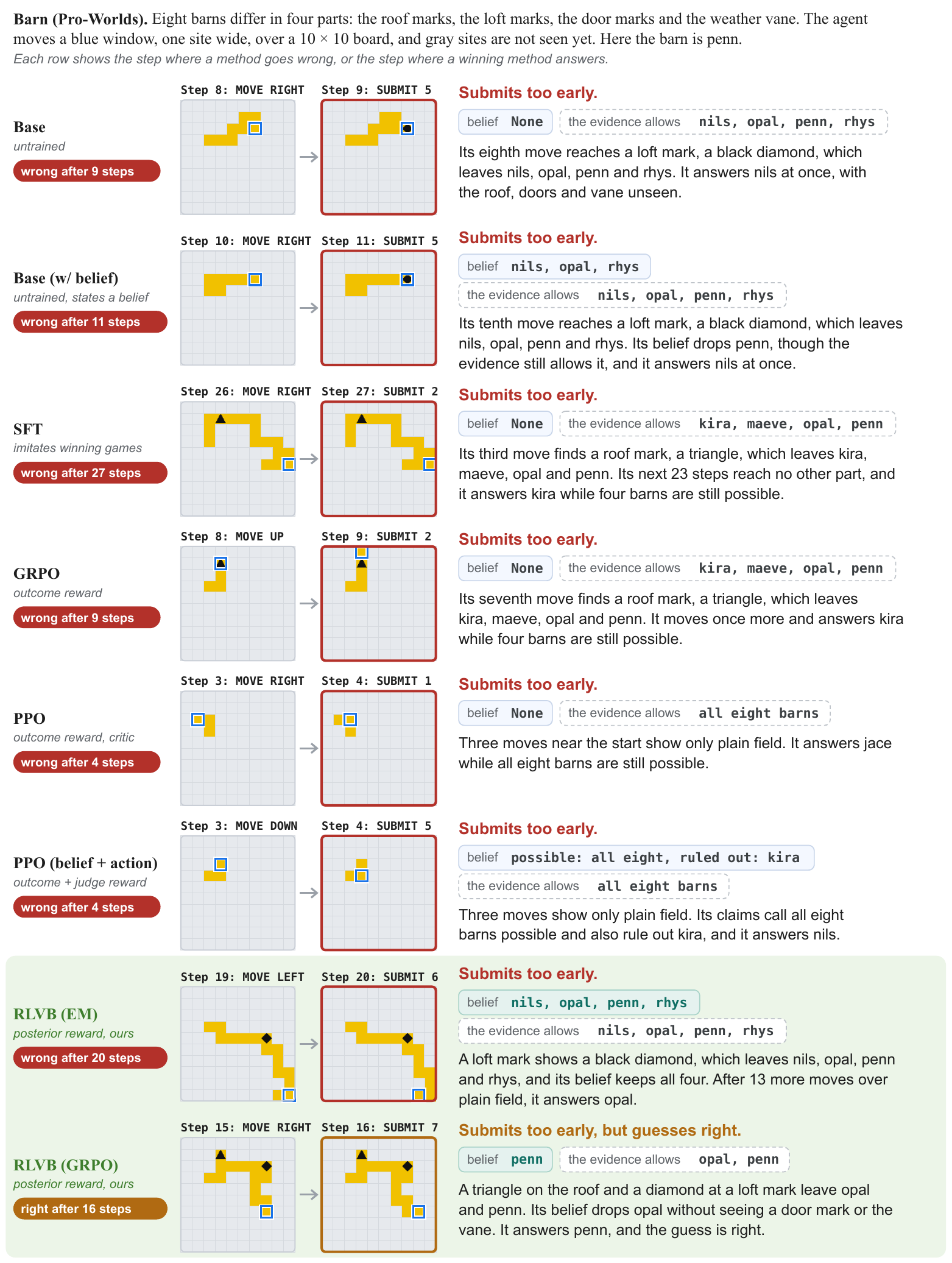}
  \caption{\textbf{Where each method goes wrong on a test episode of Barn.}}
  \label{fig:case-expansion-barn}
\end{figure*}

\begin{figure*}[p]
  \centering
  \includegraphics[width=\textwidth,height=0.88\textheight,keepaspectratio]{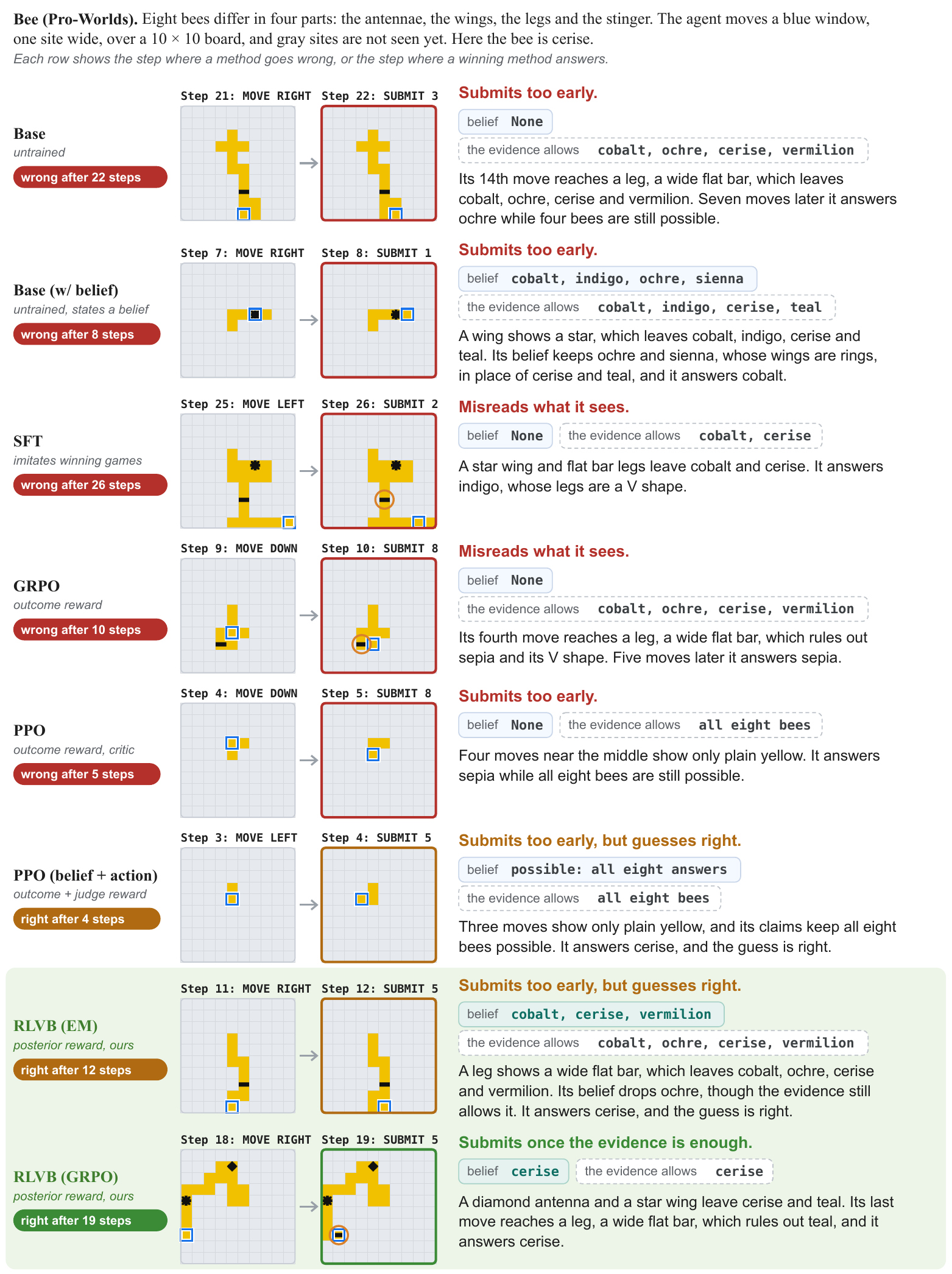}
  \caption{\textbf{Where each method goes wrong on a test episode of Bee.}}
  \label{fig:case-expansion-bee}
\end{figure*}

\begin{figure*}[p]
  \centering
  \includegraphics[width=\textwidth,height=0.88\textheight,keepaspectratio]{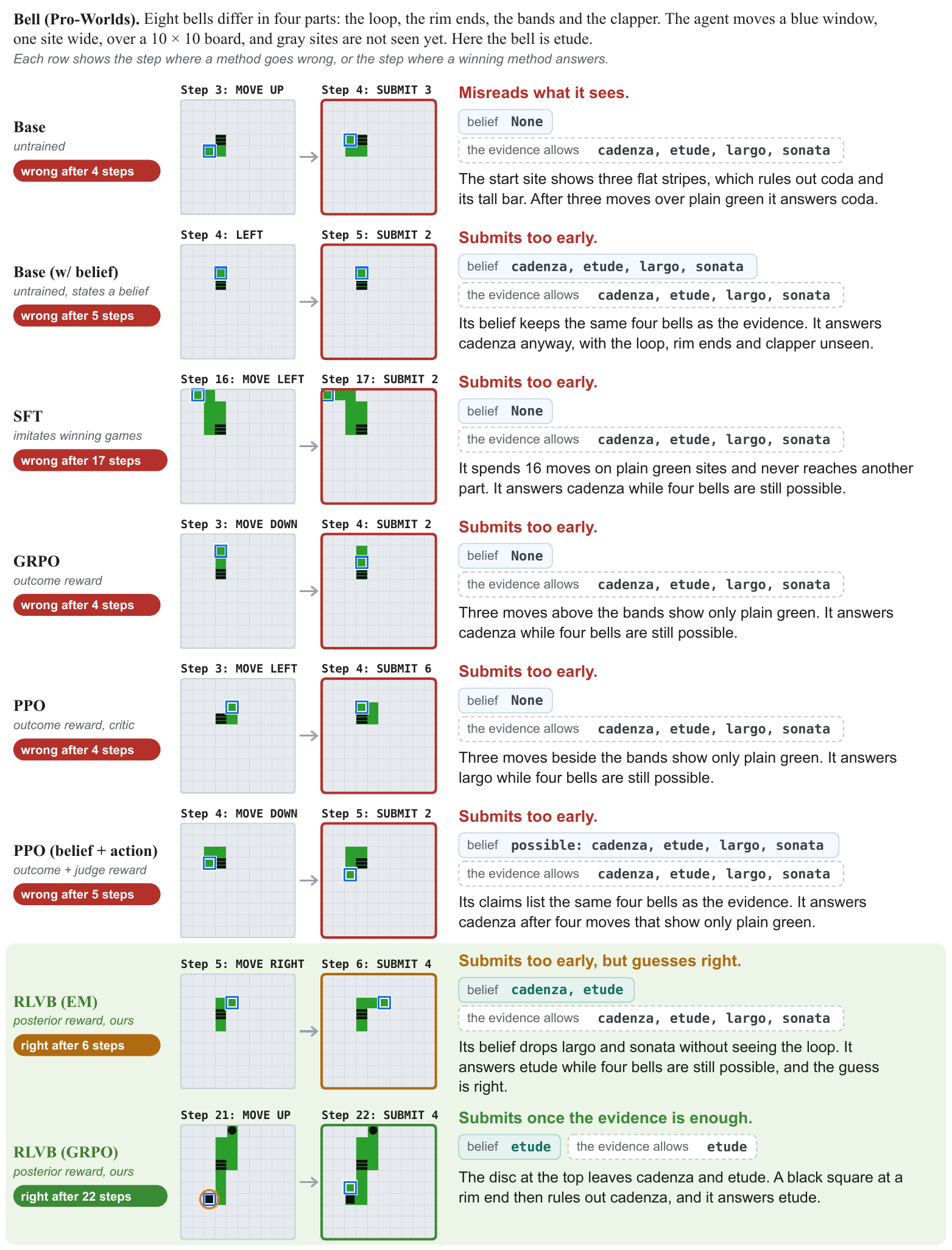}
  \caption{\textbf{Where each method goes wrong on a test episode of Bell.}}
  \label{fig:case-expansion-bell}
\end{figure*}

\begin{figure*}[p]
  \centering
  \includegraphics[width=\textwidth,height=0.88\textheight,keepaspectratio]{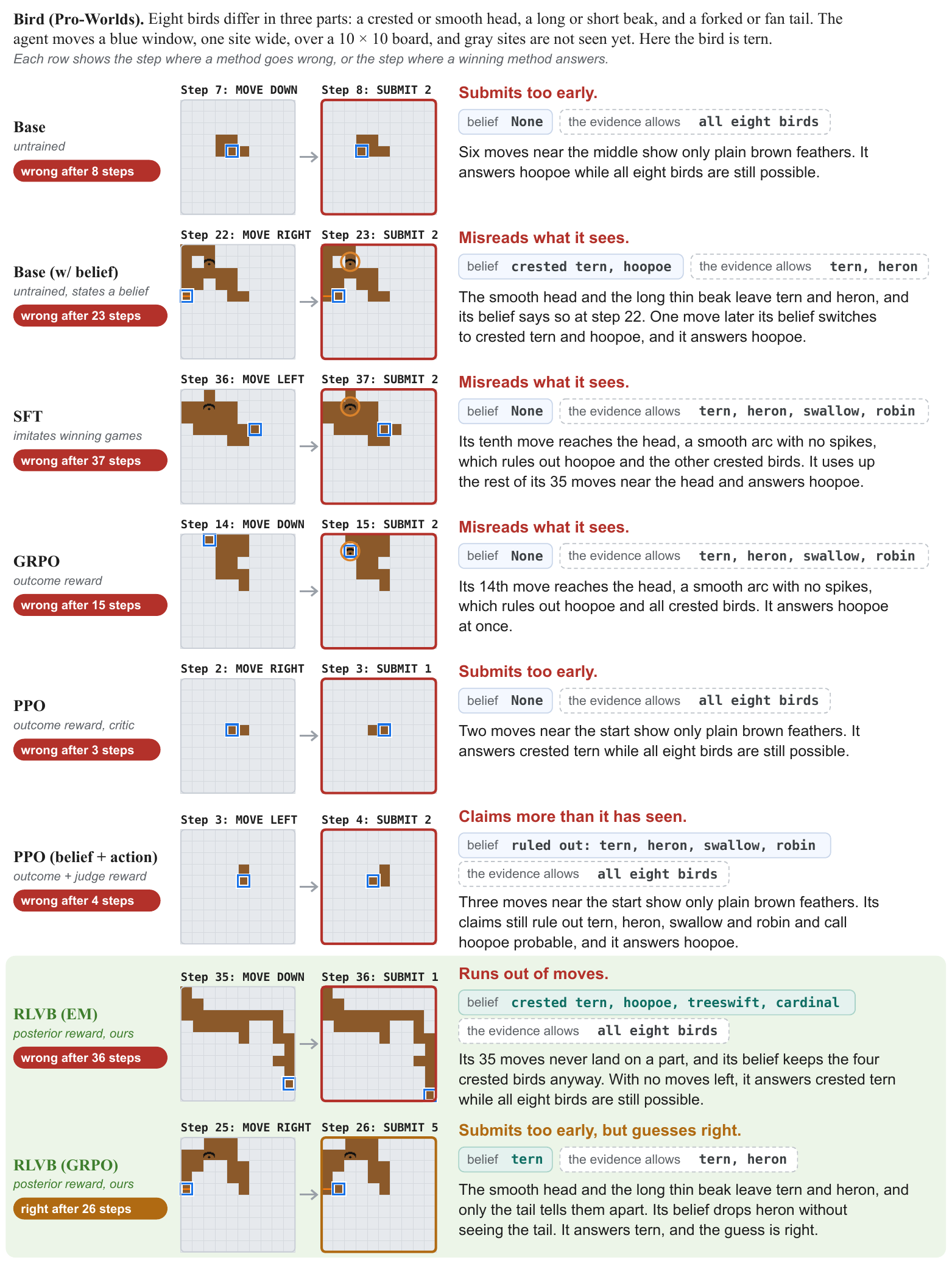}
  \caption{\textbf{Where each method goes wrong on a test episode of Bird.}}
  \label{fig:case-expansion-bird}
\end{figure*}

\begin{figure*}[p]
  \centering
  \includegraphics[width=\textwidth,height=0.88\textheight,keepaspectratio]{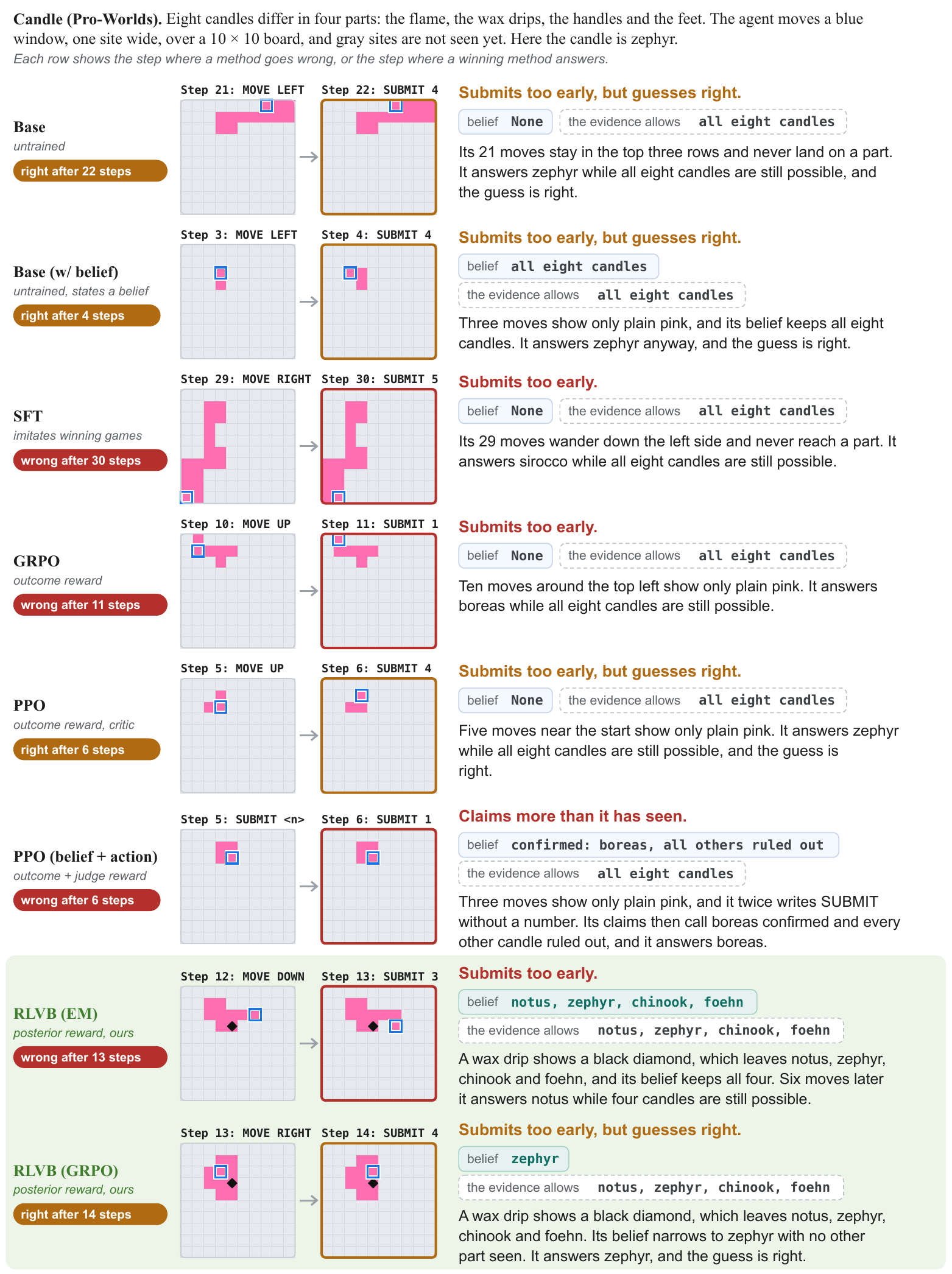}
  \caption{\textbf{Where each method goes wrong on a test episode of Candle.}}
  \label{fig:case-expansion-candle}
\end{figure*}

\begin{figure*}[p]
  \centering
  \includegraphics[width=\textwidth,height=0.88\textheight,keepaspectratio]{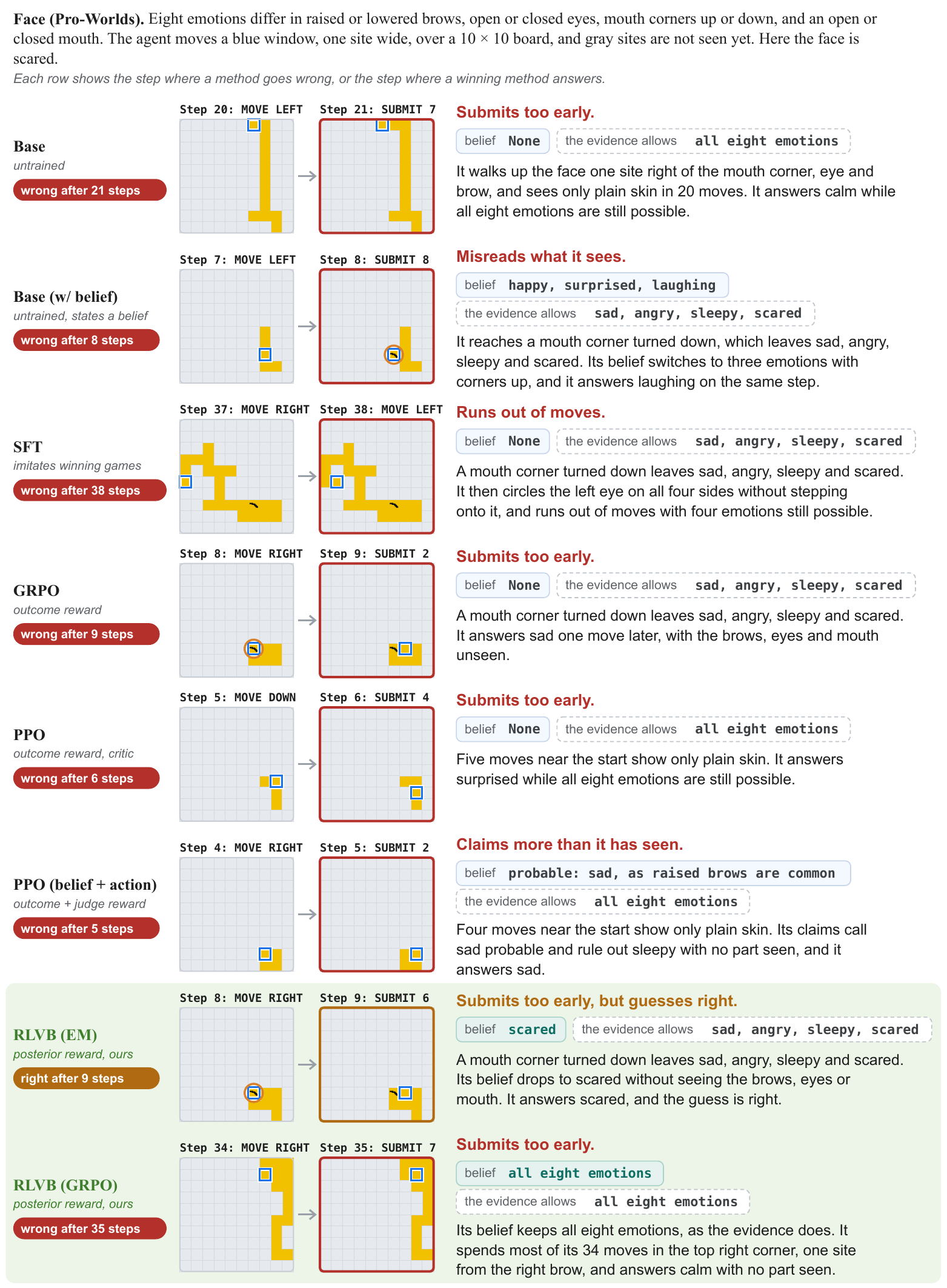}
  \caption{\textbf{Where each method goes wrong on a test episode of Face.}}
  \label{fig:case-expansion-face}
\end{figure*}

\begin{figure*}[p]
  \centering
  \includegraphics[width=\textwidth,height=0.88\textheight,keepaspectratio]{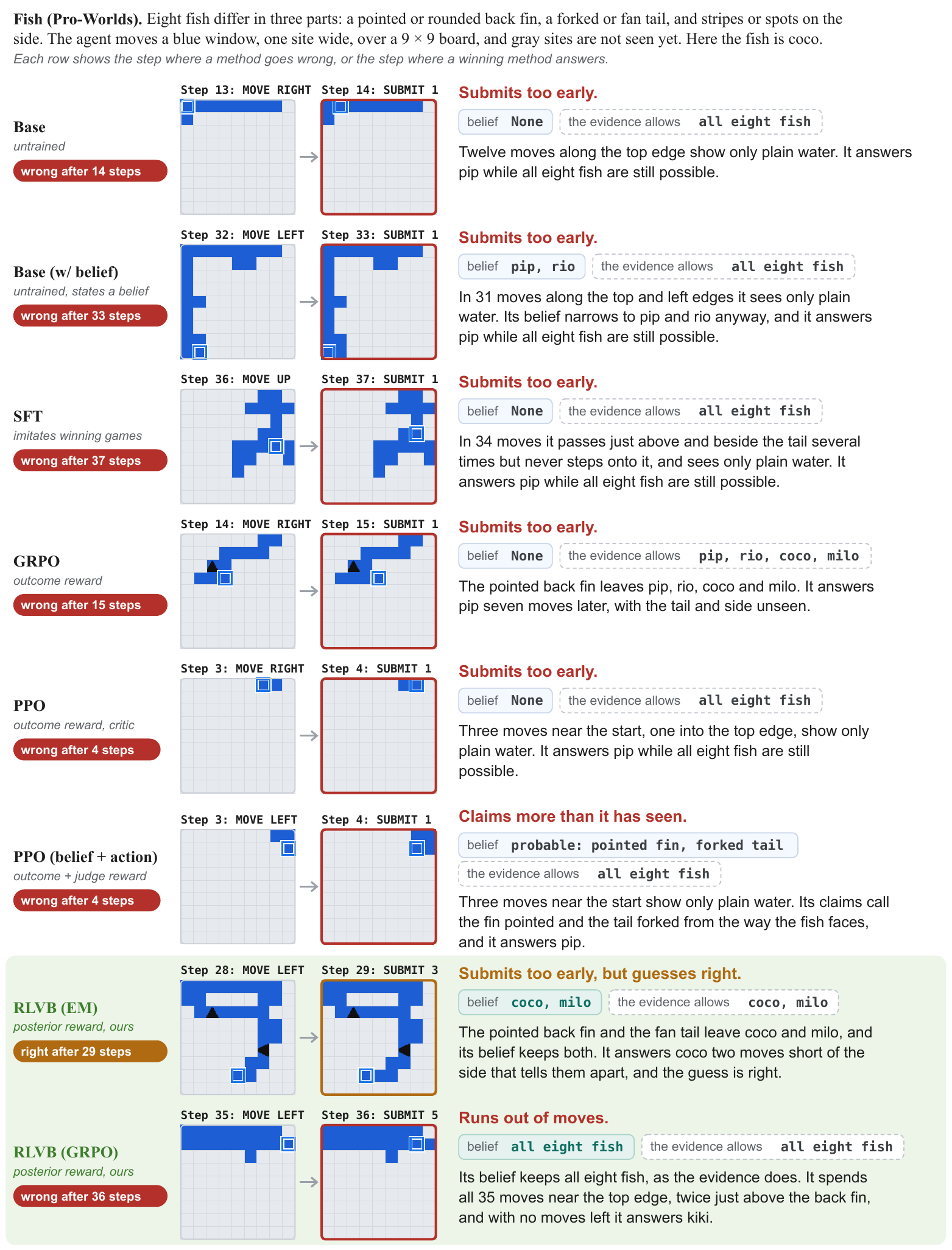}
  \caption{\textbf{Where each method goes wrong on a test episode of Fish.}}
  \label{fig:case-expansion-fish}
\end{figure*}

\begin{figure*}[p]
  \centering
  \includegraphics[width=\textwidth,height=0.88\textheight,keepaspectratio]{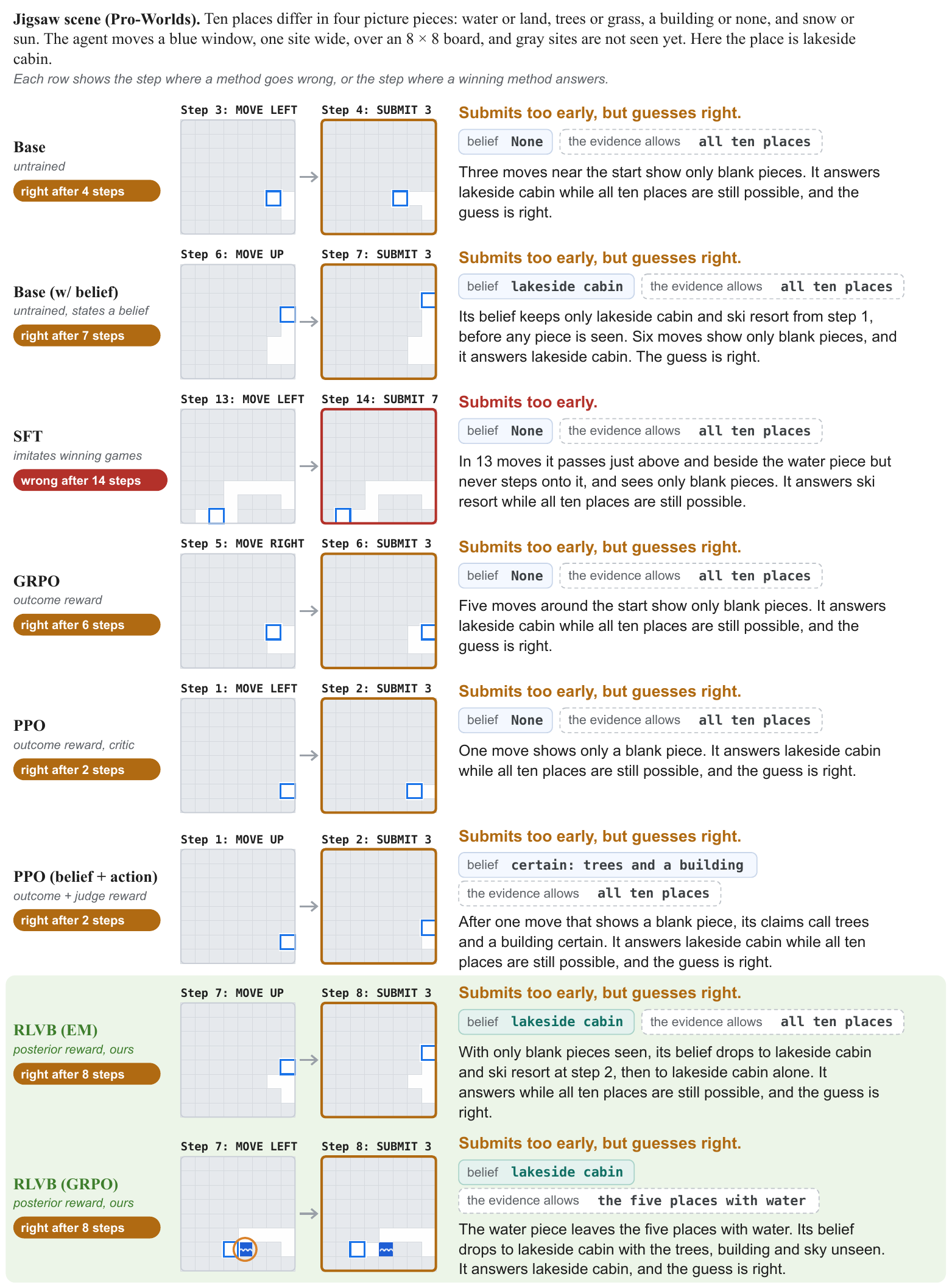}
  \caption{\textbf{Where each method goes wrong on a test episode of Jigsaw.}}
  \label{fig:case-expansion-jigsaw}
\end{figure*}

\begin{figure*}[p]
  \centering
  \includegraphics[width=\textwidth,height=0.88\textheight,keepaspectratio]{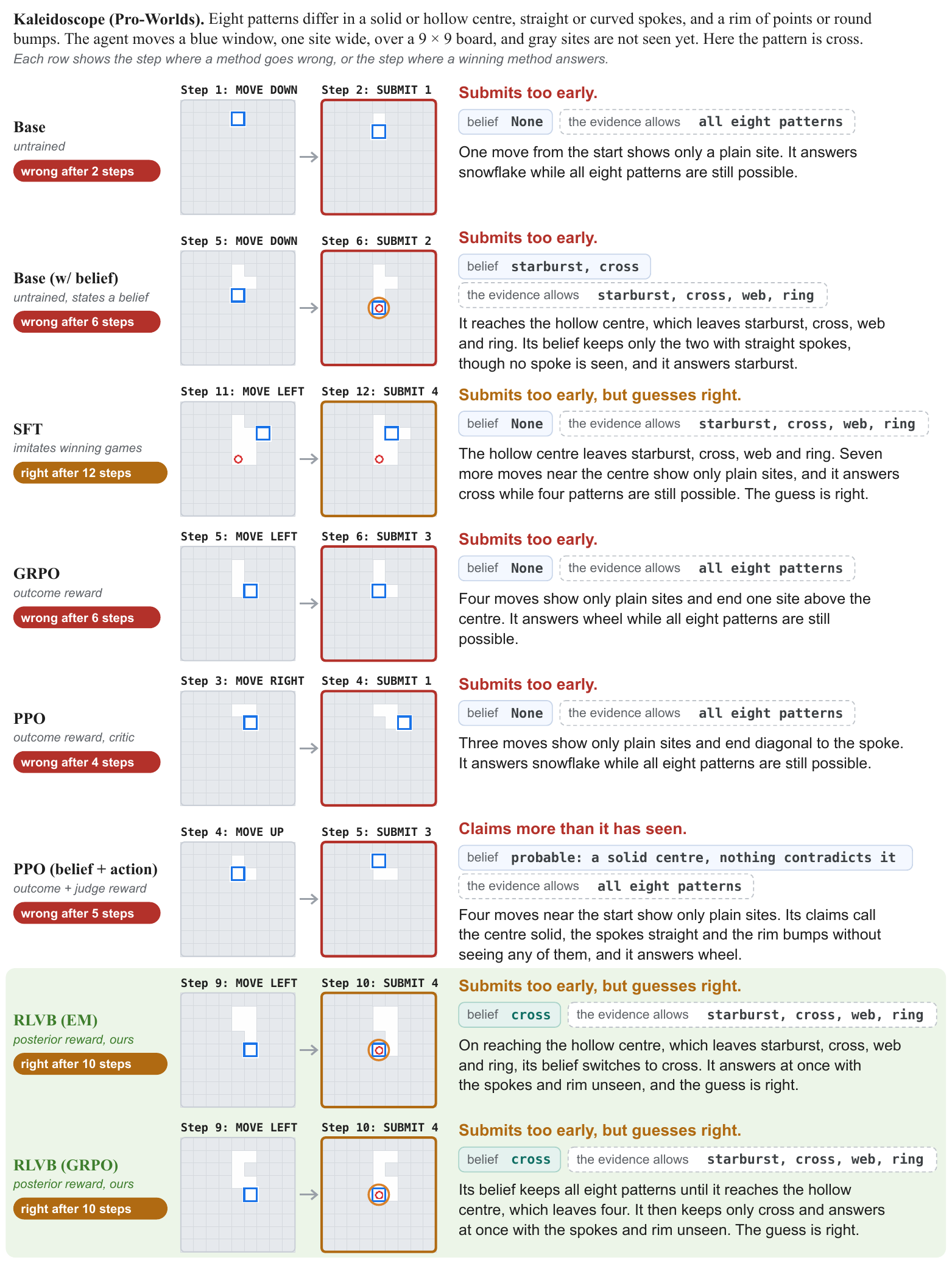}
  \caption{\textbf{Where each method goes wrong on a test episode of Kaleidoscope.}}
  \label{fig:case-expansion-kaleido}
\end{figure*}

\begin{figure*}[p]
  \centering
  \includegraphics[width=\textwidth,height=0.88\textheight,keepaspectratio]{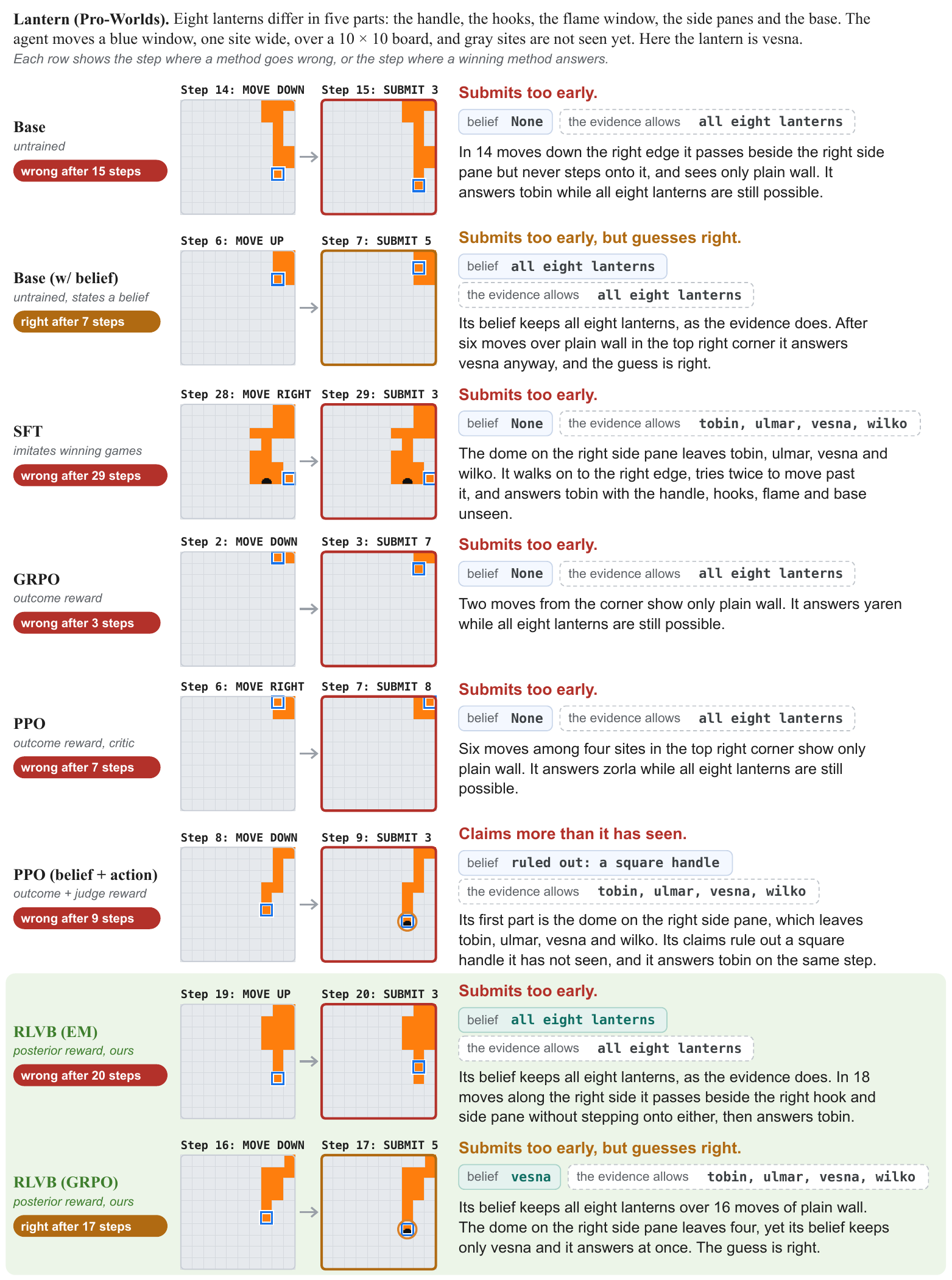}
  \caption{\textbf{Where each method goes wrong on a test episode of Lantern.}}
  \label{fig:case-expansion-lantern}
\end{figure*}

\begin{figure*}[p]
  \centering
  \includegraphics[width=\textwidth,height=0.88\textheight,keepaspectratio]{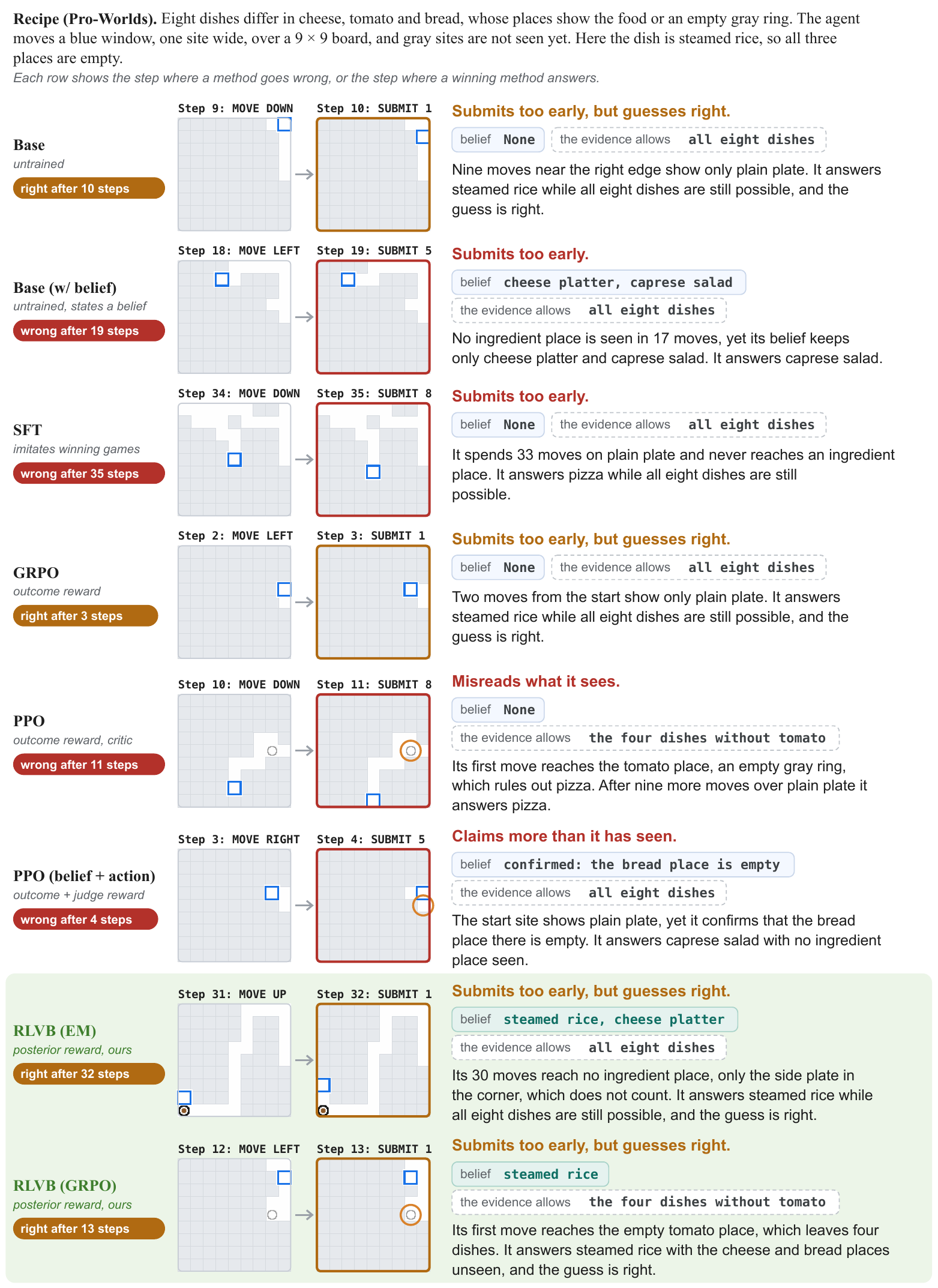}
  \caption{\textbf{Where each method goes wrong on a test episode of Recipe.}}
  \label{fig:case-expansion-recipe}
\end{figure*}

\begin{figure*}[p]
  \centering
  \includegraphics[width=\textwidth,height=0.88\textheight,keepaspectratio]{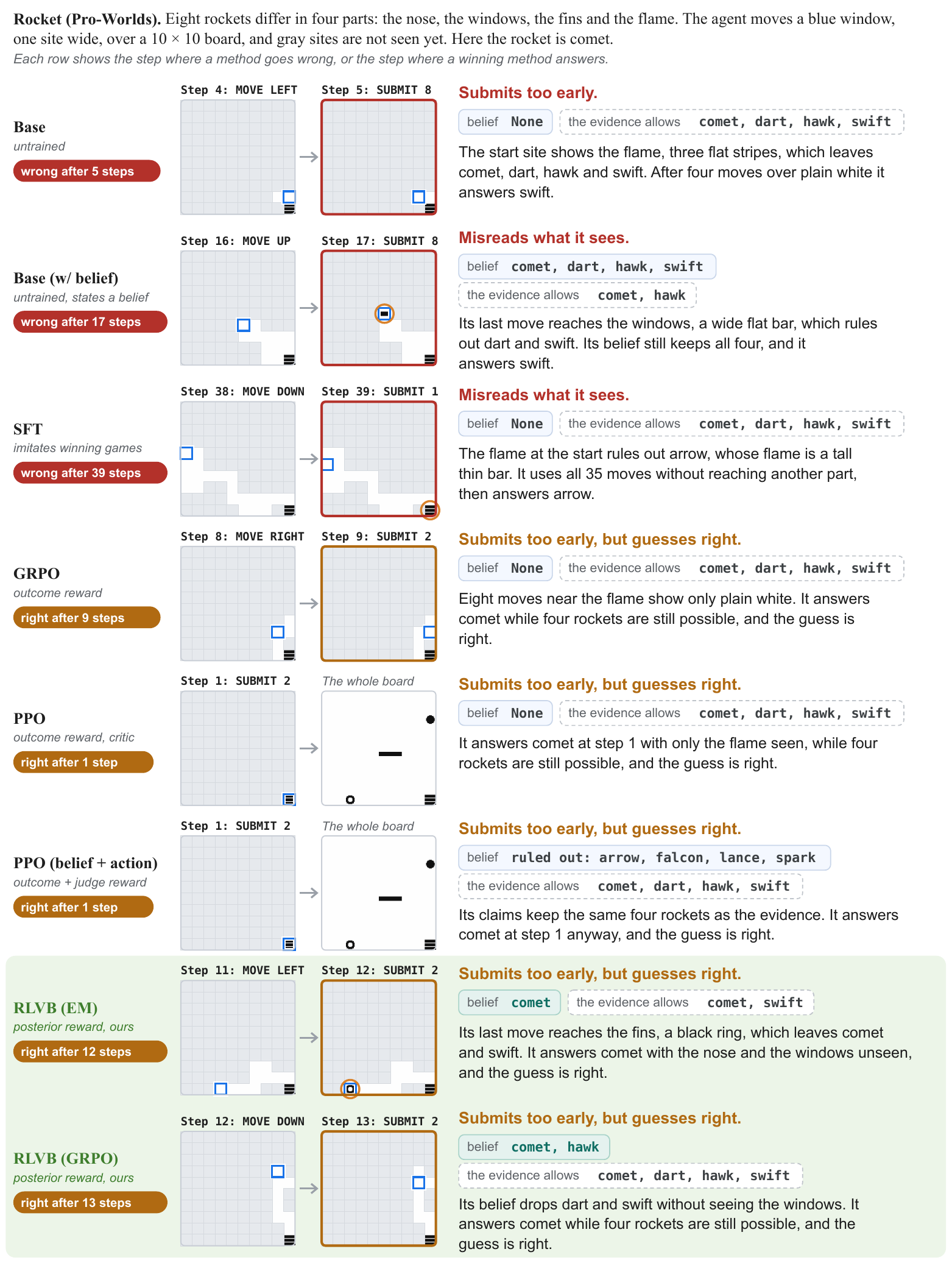}
  \caption{\textbf{Where each method goes wrong on a test episode of Rocket.}}
  \label{fig:case-expansion-rocket}
\end{figure*}

\begin{figure*}[p]
  \centering
  \includegraphics[width=\textwidth,height=0.88\textheight,keepaspectratio]{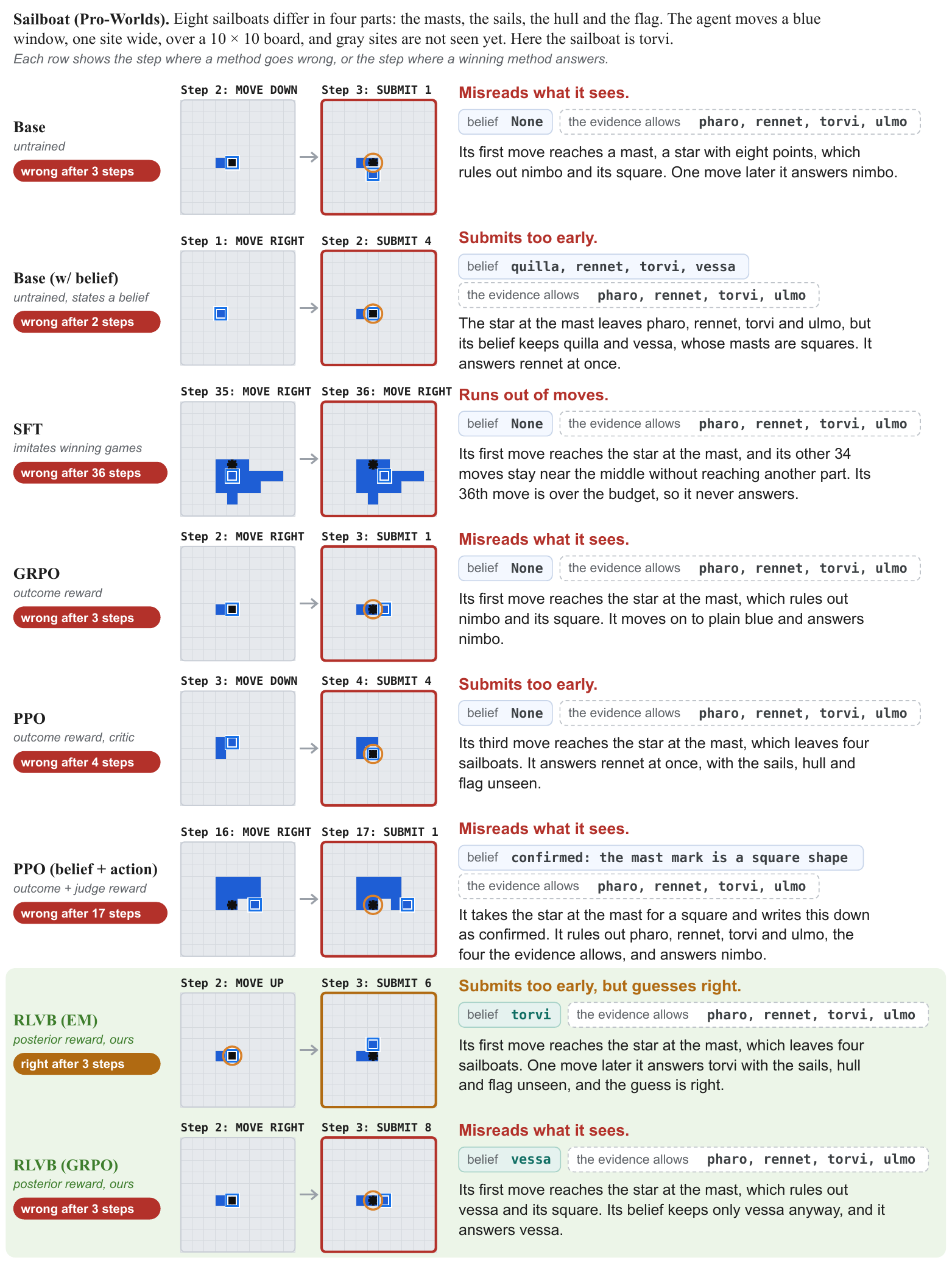}
  \caption{\textbf{Where each method goes wrong on a test episode of Sailboat.}}
  \label{fig:case-expansion-sailboat}
\end{figure*}

\begin{figure*}[p]
  \centering
  \includegraphics[width=\textwidth,height=0.88\textheight,keepaspectratio]{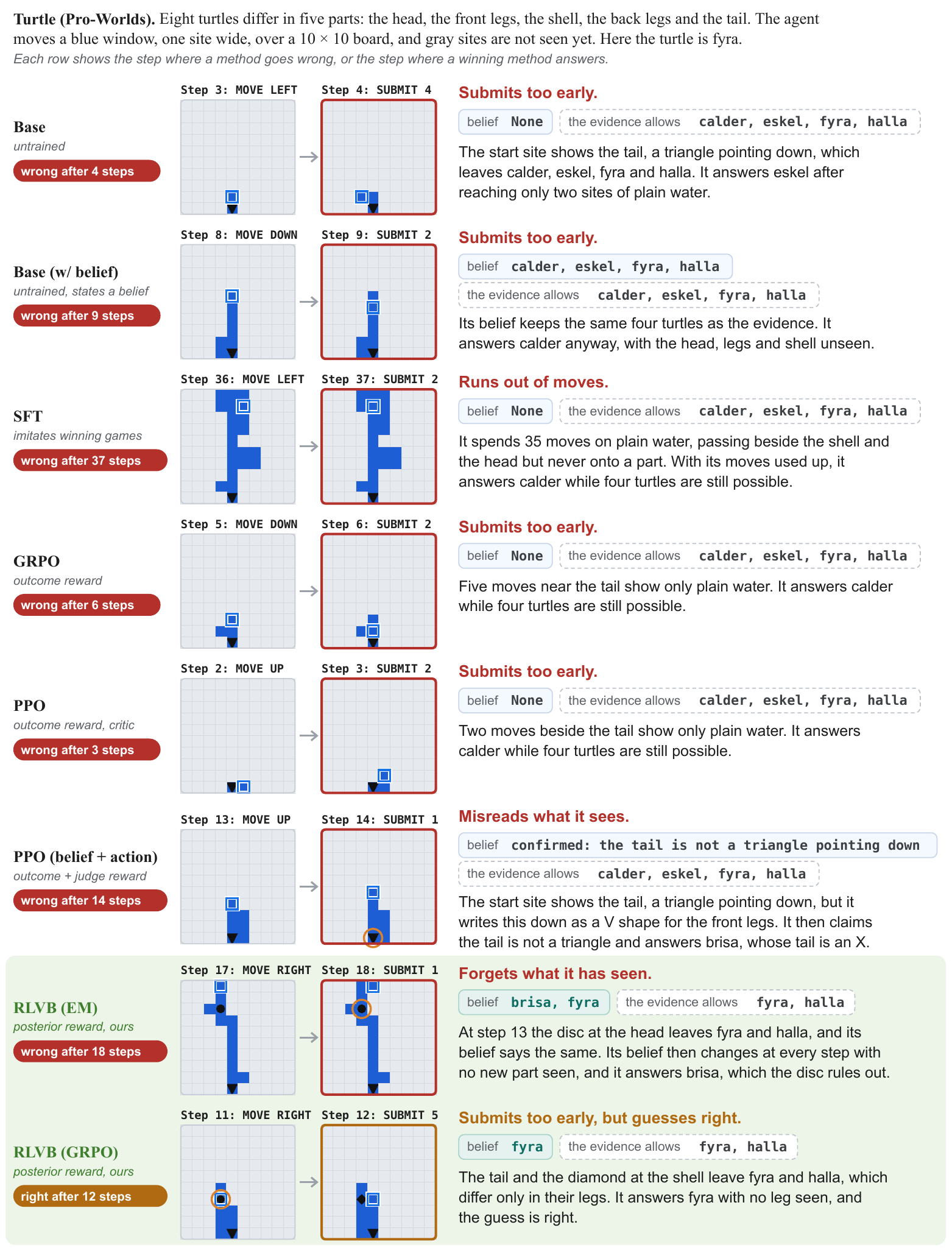}
  \caption{\textbf{Where each method goes wrong on a test episode of Turtle.}}
  \label{fig:case-expansion-turtle}
\end{figure*}

\begin{figure*}[p]
  \centering
  \includegraphics[width=\textwidth,height=0.88\textheight,keepaspectratio]{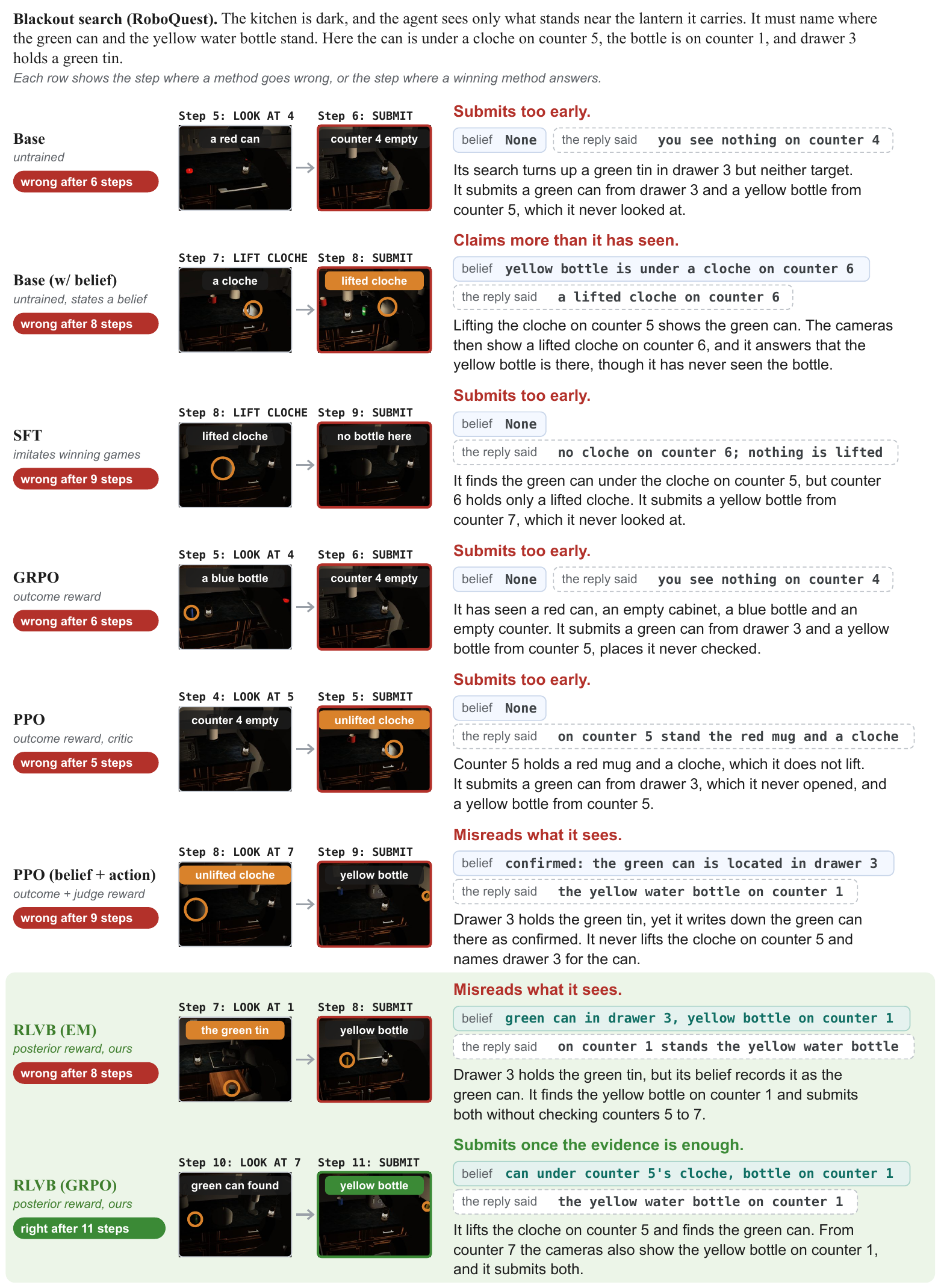}
  \caption{\textbf{Where each method goes wrong on a test episode of Blackout search.} The agent read the text version of the episode, and the frames show the same steps in the simulator.}
  \label{fig:case-text-blackout-search}
\end{figure*}

\begin{figure*}[p]
  \centering
  \includegraphics[width=\textwidth,height=0.88\textheight,keepaspectratio]{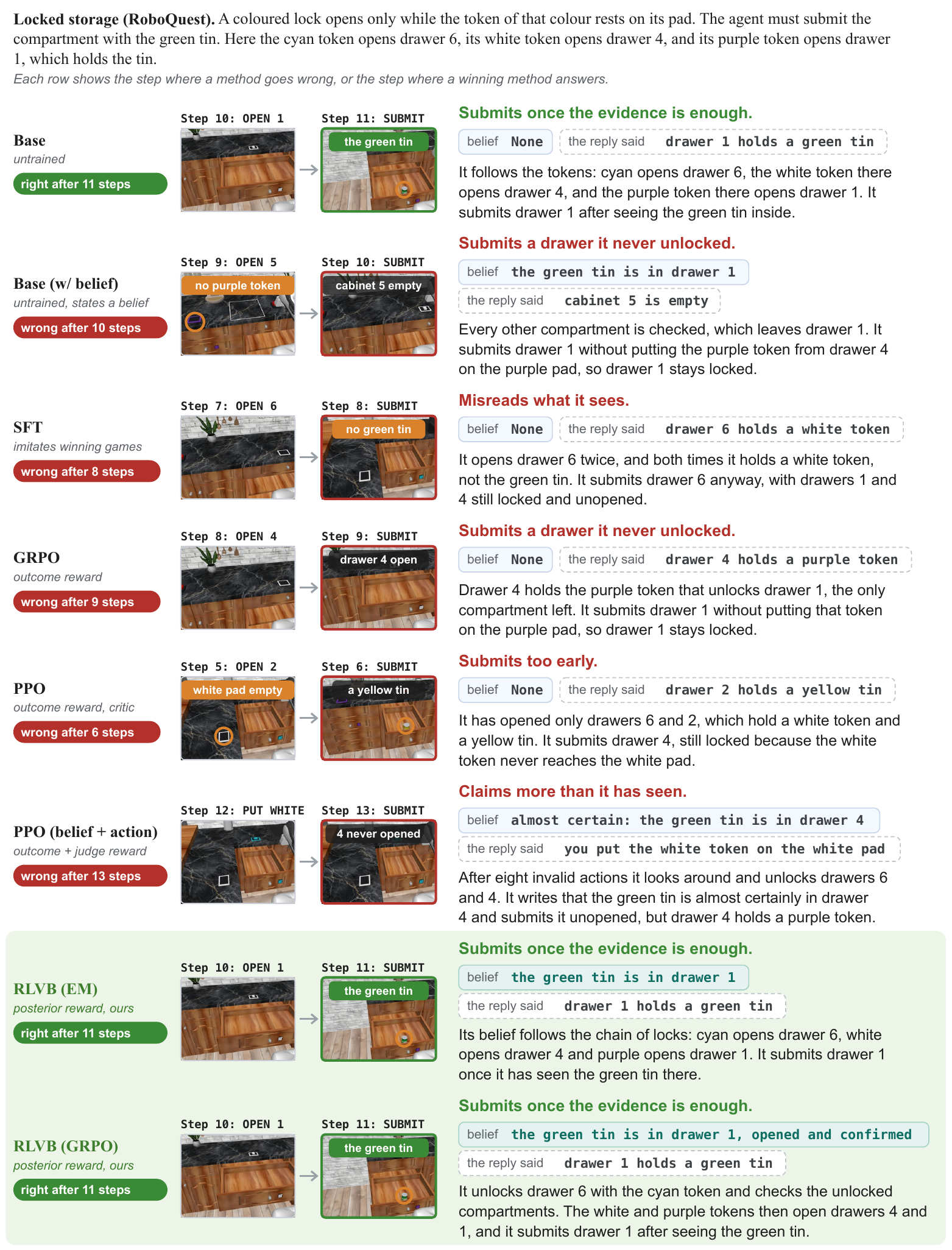}
  \caption{\textbf{Where each method goes wrong on a test episode of Locked storage.}}
  \label{fig:case-text-locked-storage}
\end{figure*}

\begin{figure*}[p]
  \centering
  \includegraphics[width=\textwidth,height=0.88\textheight,keepaspectratio]{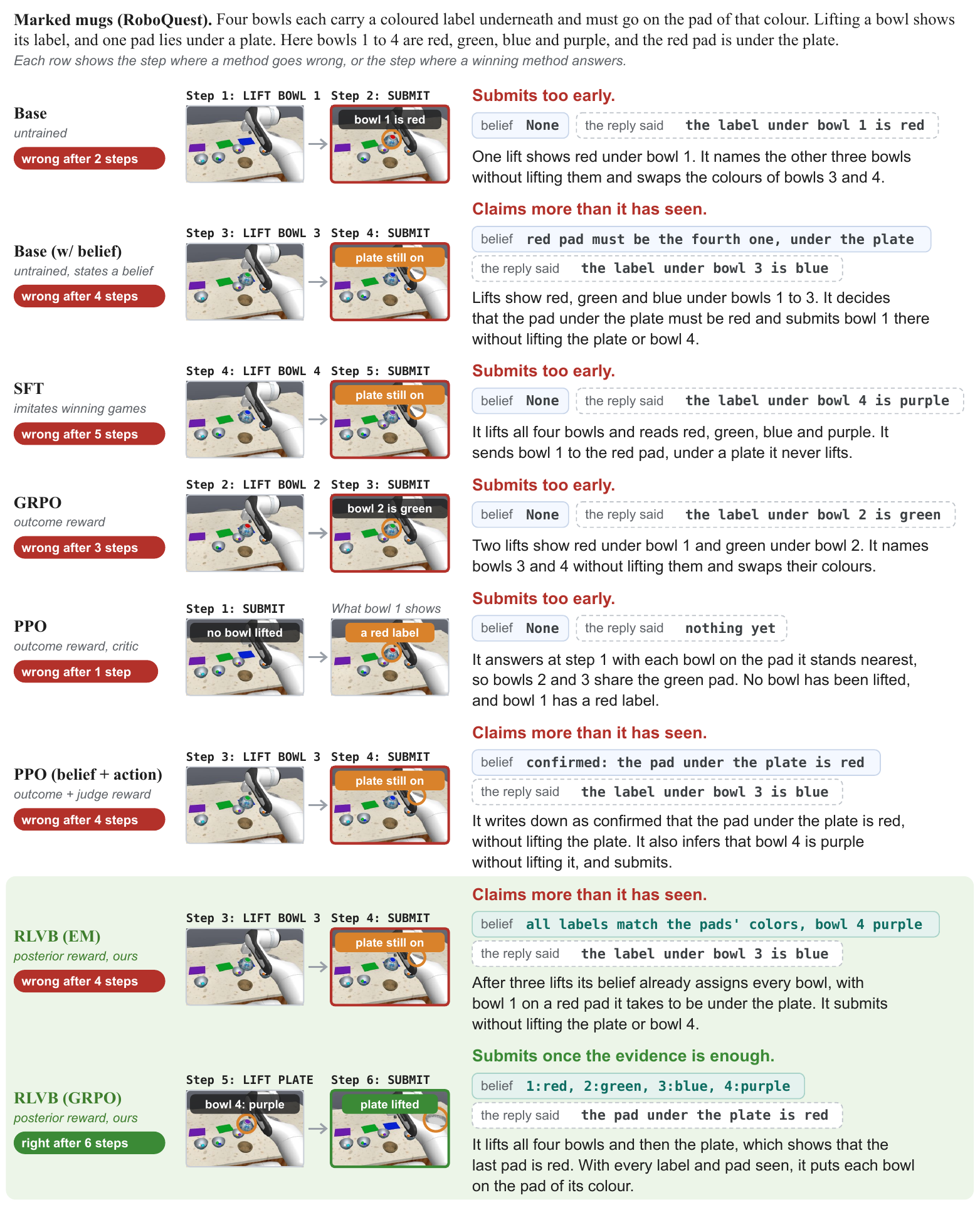}
  \caption{\textbf{Where each method goes wrong on a test episode of Marked mugs.}}
  \label{fig:case-text-marked-mugs}
\end{figure*}

\begin{figure*}[p]
  \centering
  \includegraphics[width=\textwidth,height=0.88\textheight,keepaspectratio]{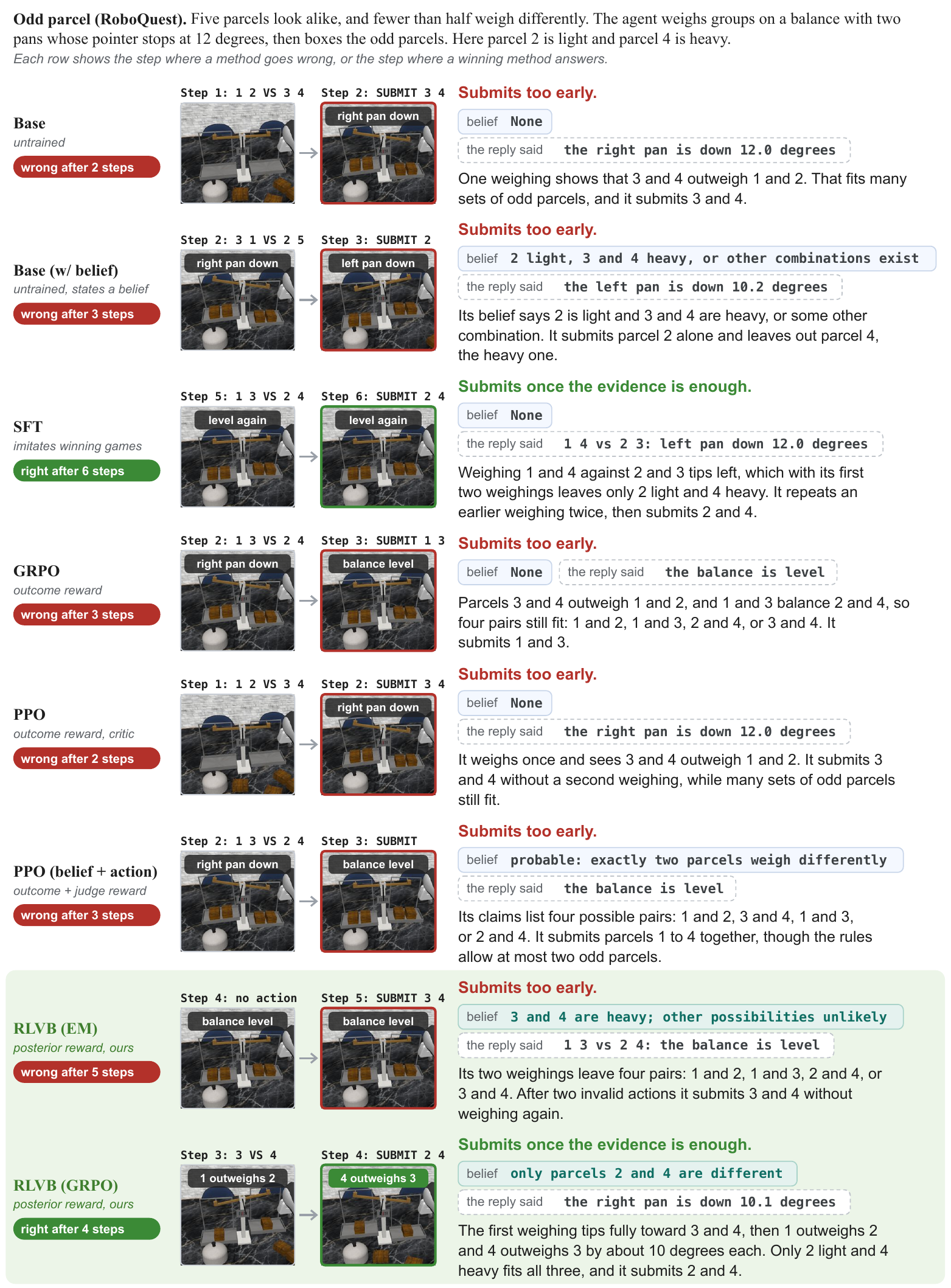}
  \caption{\textbf{Where each method goes wrong on a test episode of Odd parcel.}}
  \label{fig:case-text-odd-parcel}
\end{figure*}

\begin{figure*}[p]
  \centering
  \includegraphics[width=\textwidth,height=0.88\textheight,keepaspectratio]{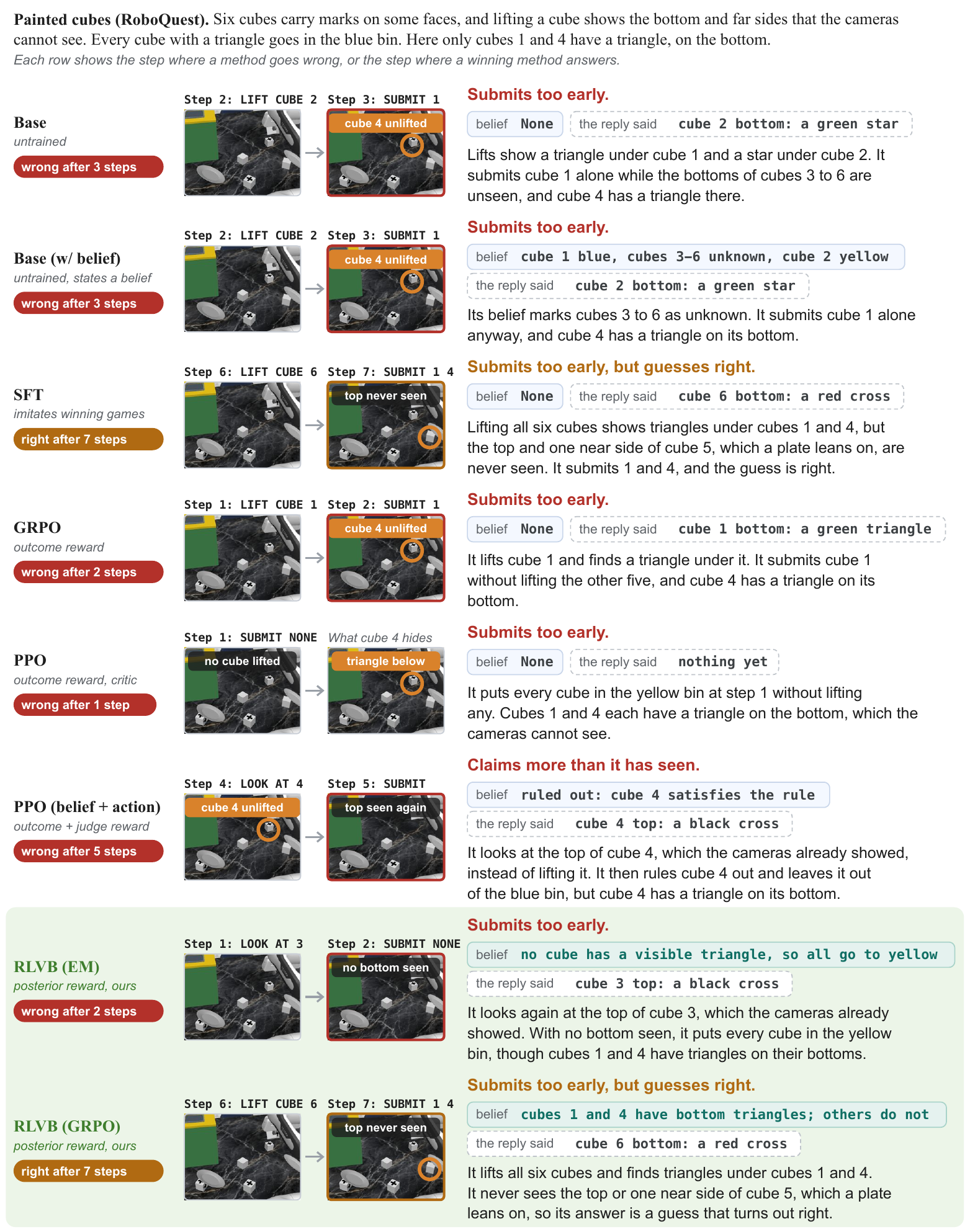}
  \caption{\textbf{Where each method goes wrong on a test episode of Painted cubes.}}
  \label{fig:case-text-painted-cubes}
\end{figure*}

\begin{figure*}[p]
  \centering
  \includegraphics[width=\textwidth,height=0.88\textheight,keepaspectratio]{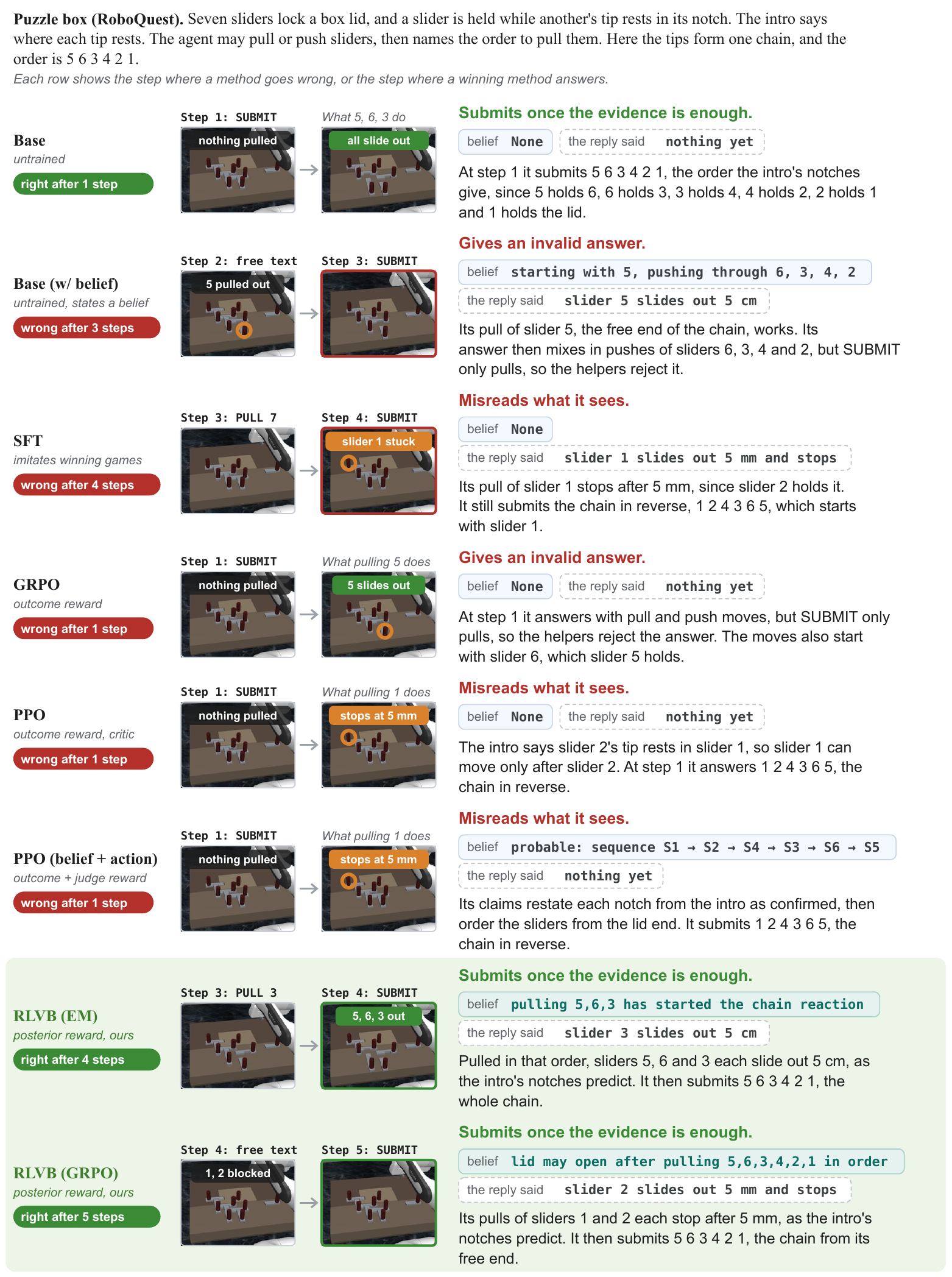}
  \caption{\textbf{Where each method goes wrong on a test episode of Puzzle box.}}
  \label{fig:case-text-puzzle-box}
\end{figure*}

\begin{figure*}[p]
  \centering
  \includegraphics[width=\textwidth,height=0.88\textheight,keepaspectratio]{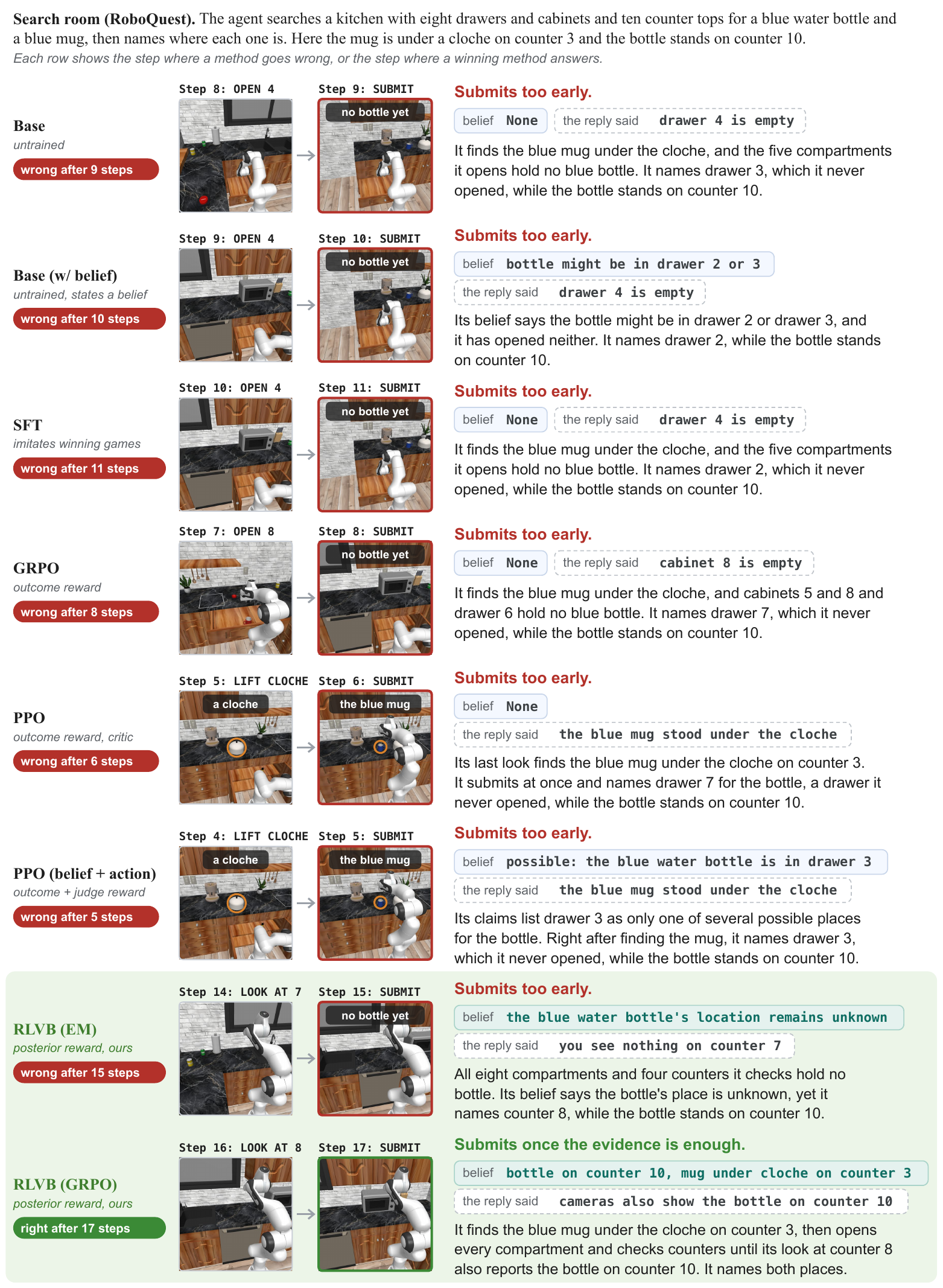}
  \caption{\textbf{Where each method goes wrong on a test episode of Search room.}}
  \label{fig:case-text-search-room}
\end{figure*}

\begin{figure*}[p]
  \centering
  \includegraphics[width=\textwidth,height=0.88\textheight,keepaspectratio]{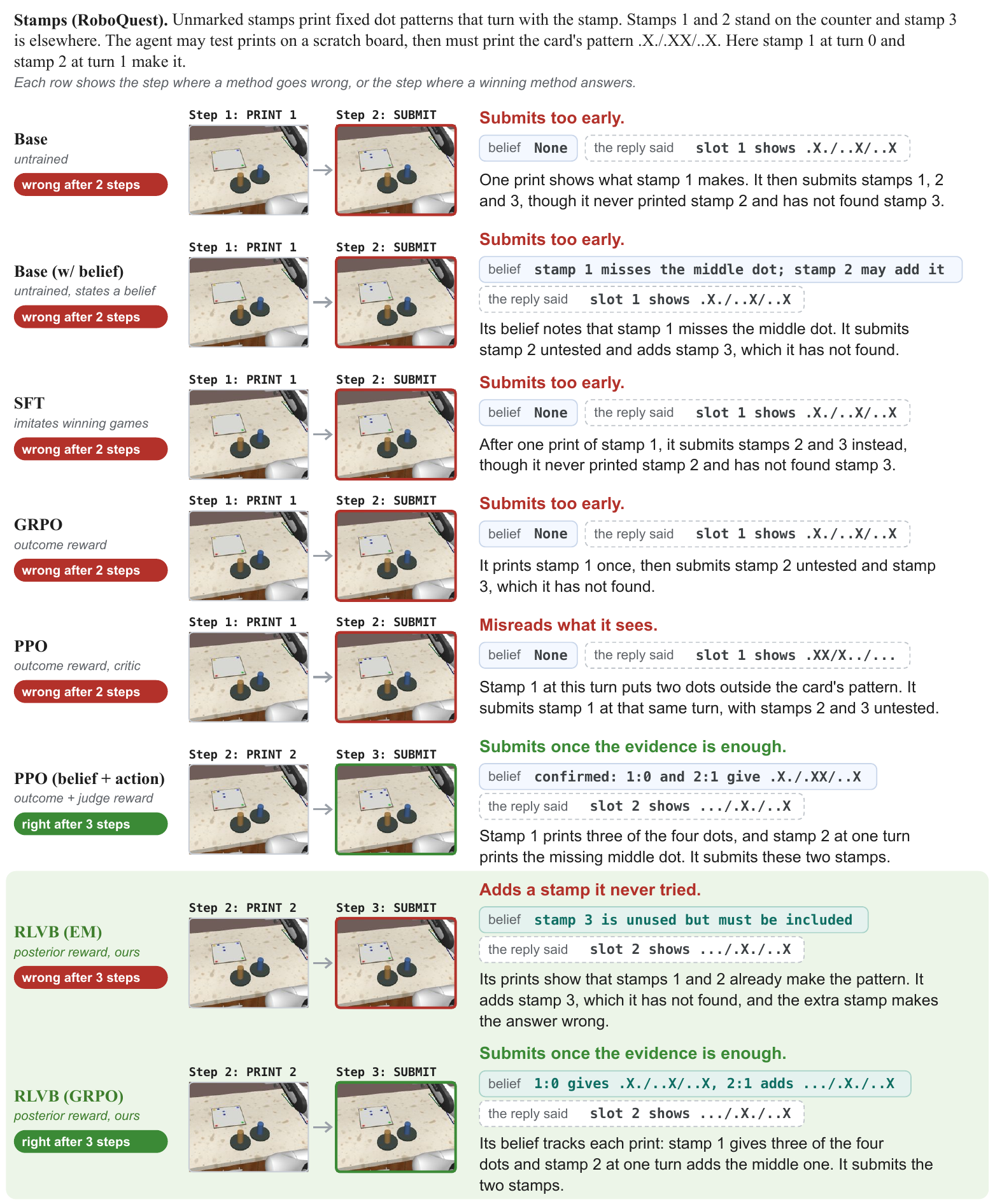}
  \caption{\textbf{Where each method goes wrong on a test episode of Stamps.}}
  \label{fig:case-text-stamps}
\end{figure*}

\begin{figure*}[p]
  \centering
  \includegraphics[width=\textwidth,height=0.88\textheight,keepaspectratio]{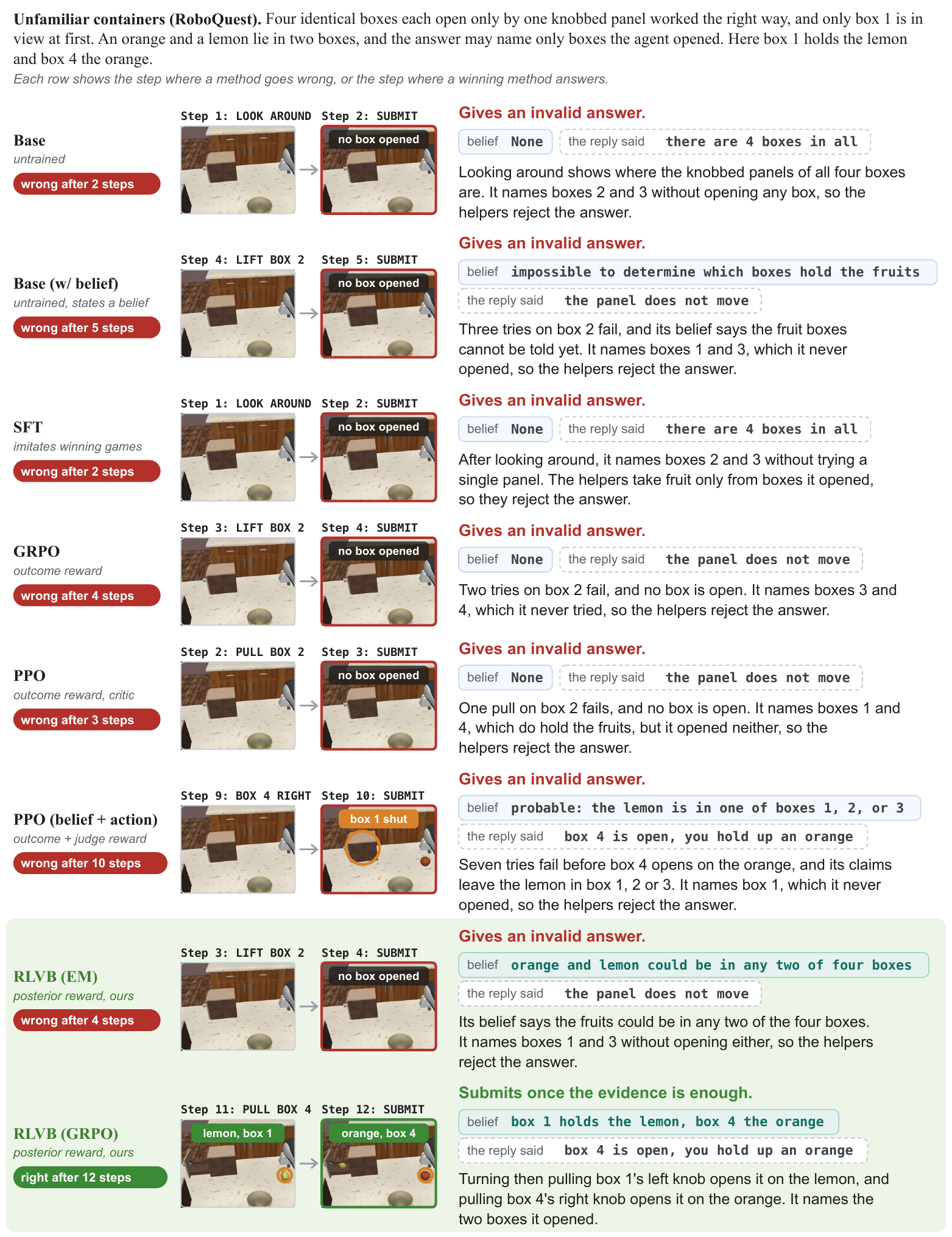}
  \caption{\textbf{Where each method goes wrong on a test episode of Unfamiliar containers.}}
  \label{fig:case-text-unfamiliar-containers}
\end{figure*}

\begin{figure*}[p]
  \centering
  \includegraphics[width=\textwidth,height=0.88\textheight,keepaspectratio]{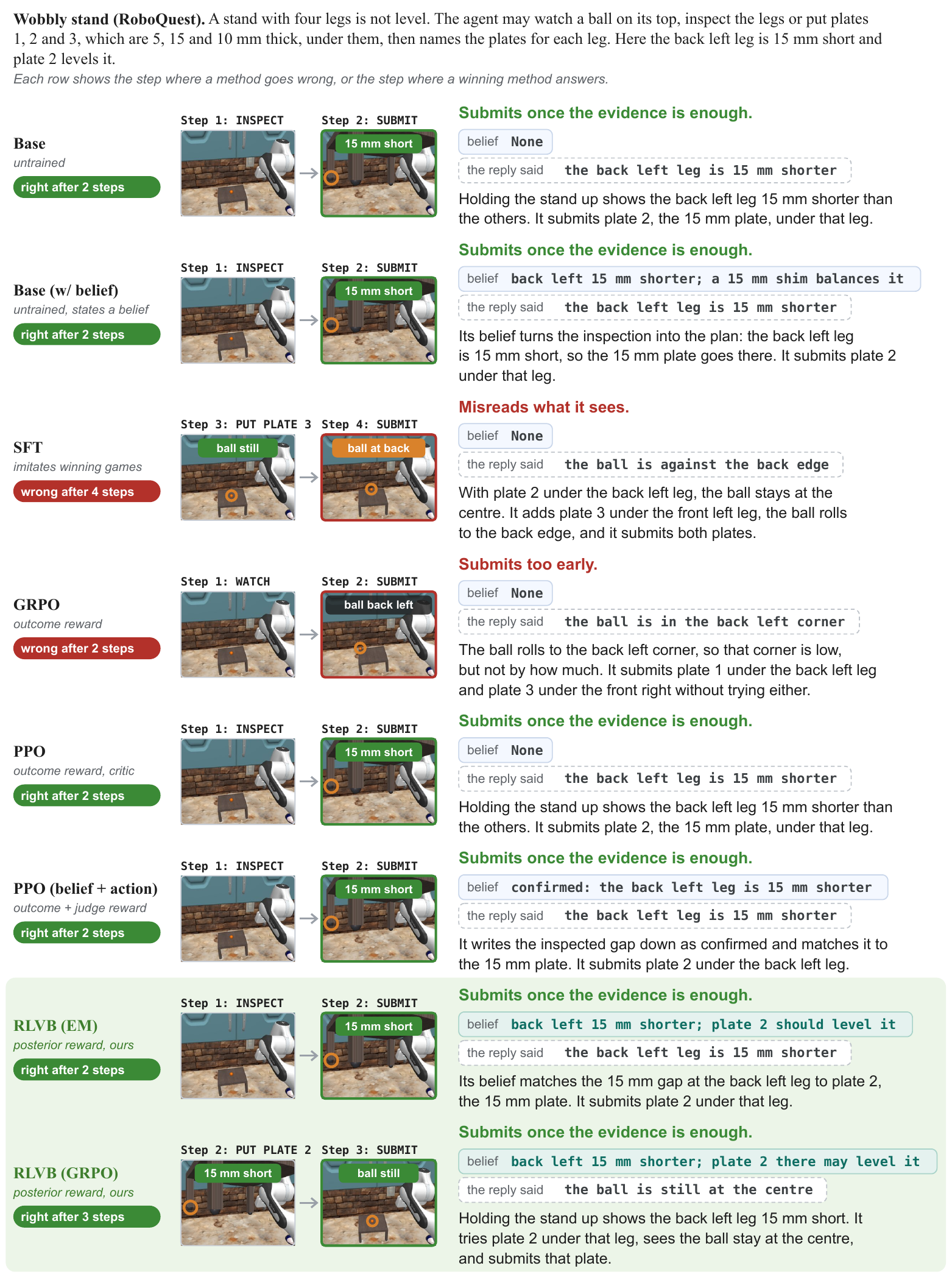}
  \caption{\textbf{Where each method goes wrong on a test episode of Wobbly stand.}}
  \label{fig:case-text-wobbly-stand}
\end{figure*}

\paragraph{Case studies.}
\Cref{tab:cases} shows two games of the cups task played by Qwen3-VL-8B-Thinking. To show a typical case, we drew each game at random from a group of similar games. Beliefs in red drop an answer still allowed by the evidence. \Cref{fig:case-cups} to \cref{fig:case-text-wobbly-stand} show one test episode of each task.

\begin{table*}[!htbp]
  \centering
  \footnotesize
  \begin{tabular}{@{}c l l l l@{}}
    \toprule
    Turn & Belief written by the agent & Answers allowed by the evidence & Action & What the lift shows \\
    \midrule
    \multicolumn{5}{@{}l}{\textit{(a) Find the pink cylinder, which is under cup 1. Base (w/ belief).}} \\
    1 & \textcolor{BrickRed}{\bline{Possible: cups 2, 3}} & all & \texttt{SUBMIT 3} & wrong \\
    \midrule
    \multicolumn{5}{@{}l}{\textit{(b) The same game. \method.}} \\
    1 & \bline{Possible: all} & all & \texttt{LIFT 2} & cup 2: nothing \\
    2 & \bline{Possible: all} & cups 1, 3, 4 & \texttt{LIFT 1} & cup 1: pink cylinder \\
    3 & \bline{Possible: 1} & cup 1 & \texttt{SUBMIT 1} & right \\
    \midrule
    \multicolumn{5}{@{}l}{\textit{(c) Find the orange cylinder, which is under cup 4. The other cups hide orange cubes and an orange ball. \method.}} \\
    1 & \bline{Possible: all} & all & \texttt{LIFT 1} & cup 1: orange cube \\
    2 & \bline{Possible: 2,3,4,5} & cups 2, 3, 4, 5 & \texttt{LIFT 2} & cup 2: orange cube \\
    3 & \bline{Possible: 3,4,5} & cups 3, 4, 5 & \texttt{LIFT 3} & cup 3: orange cube \\
    4 & \bline{Possible: 4,5} & cups 4, 5 & \texttt{LIFT 4} & cup 4: orange cylinder \\
    5 & \textcolor{BrickRed}{\bline{Possible: 5}} & cup 4 & \texttt{SUBMIT 5} & wrong \\
    \bottomrule
  \end{tabular}
  \caption{Two cups games of Qwen3-VL-8B-Thinking, drawn at random by a fixed rule.}
  \label{tab:cases}
\end{table*}

We drew the first game from the 28 of 100 test games where \method waits and answers correctly but Base (w/ belief) answers too early and wrongly. The agent must find the pink cylinder, which is under cup 1. Base (w/ belief) does not lift any cup. It writes that the cylinder is under cup 2 or 3, although it has no evidence against cups 1 and 4. It then submits cup 3. \method first lifts cup 2 and finds nothing. It then lifts cup 1 and finds the pink cylinder. Only then does it write that the answer is cup 1 and submit cup 1. After the first lift, its belief still names all four cups and is more cautious than needed. However, it never rules out a cup that could still hide the cylinder.

We drew the second game from the 13 games where \method answers wrongly although it had enough evidence. The agent must find the orange cylinder, which is under cup 4. The other cups hide orange cubes and an orange ball. \method lifts cups 1, 2 and 3 one by one. After each lift, its belief is correct. It then lifts cup 4 and sees the orange cylinder. The evidence now shows the answer. Yet it writes that only cup 5 is possible and submits cup 5. Its reasoning shows that it reads the object under cup 4 as a yellow cylinder. It writes ``Cup 4: lifted, yellow cylinder $\rightarrow$ no''. \method thus stopped at the right time but misread the color of the object under cup 4.

\end{document}